%% file: main.tex
\documentclass[10pt,twocolumn,letterpaper]{article}

\usepackage{iccv}
\usepackage{times}
\usepackage{epsfig}
\usepackage{graphicx}
\usepackage{amsmath}
\usepackage{amssymb}

\usepackage{booktabs}
\usepackage{bm}
\usepackage{makecell}
\usepackage{enumitem}
\usepackage{multirow}
\makeatletter
\@namedef{ver@everyshi.sty}{}
\makeatother
\usepackage{booktabs}
\usepackage{multirow}
\usepackage[normalem]{ulem}
\usepackage{ifsym}

\usepackage{tikz}
\usepackage{comment}

\usepackage[accsupp]{axessibility}  % Improves PDF readability for those with disabilities.

\usepackage{color}
\usepackage{contour}
\usepackage{ulem}
\usepackage{textcomp}
\usepackage{soul}

\contourlength{0.5pt}

\definecolor{todo}{rgb}{1.0, 0., 0.}

\newcommand\blfootnote[1]{%
\begingroup
\renewcommand\thefootnote{}\footnote{#1}%
\addtocounter{footnote}{-1}%
\endgroup
}

\definecolor{reff}{rgb}{1.0,0.0,0.0}
\newcommand{\reff}[1]{\textcolor{reff}{{{#1}}}}

\usepackage[pagebackref=true,breaklinks=true,letterpaper=true,colorlinks,bookmarks=false]{hyperref}

\iccvfinalcopy % *** Uncomment this line for the final submission

\def\iccvPaperID{9919} % *** Enter the ICCV Paper ID here

\ificcvfinal\fi

\def\pname{md4all}

\begin{document}

%%%%%%%%% TITLE
%\title{md4all: A Simple and Effective Solution for Depth Estimation in All Conditions\\with a Single Self-supervised Monocular Model}
%Day, Night and Rain 
%\title{md4all: Monocular Depth Estimation in All Conditions with a Single Model}
\title{Robust Monocular Depth Estimation under Challenging Conditions}
%md4all: A Simple and Effective Method for Supervised and Self-Supervised Robust Monocular Depth Estimation under All Conditions

\author{
\quad Stefano Gasperini$^{*,1,2}$
\quad Nils Morbitzer$^{*,1}$
\quad HyunJun Jung$^{1}$\\
\quad Nassir Navab$^{1}$
\quad Federico Tombari$^{1,3}$\\\\
$^1$ Technical University of Munich \quad $^2$ VisualAIs \quad $^3$ Google
}

\maketitle
\ificcvfinal\thispagestyle{empty}\fi

\blfootnote{$^{*}$ The authors contributed equally.}
\blfootnote{Contact author: Stefano Gasperini (\textit{stefano.gasperini@tum.de}).}

%For self-supervised methods, this is due to the failure of the learning assumptions, while supervised models learn artifacts exhibited by the ground truth 3D sensors.
%%%%%%%%% ABSTRACT
\begin{abstract}
While state-of-the-art monocular depth estimation approaches achieve impressive results in ideal settings, they are highly unreliable under challenging illumination and weather conditions, such as at nighttime or in the presence of rain. In this paper, we uncover these safety-critical issues and tackle them with \pname: a simple and effective solution that works reliably under both adverse and ideal conditions, as well as for different types of learning supervision. We achieve this by exploiting the efficacy of existing methods under perfect settings. Therefore, we provide valid training signals independently of what is in the input. First, we generate a set of complex samples corresponding to the normal training ones. Then, we train the model by guiding its self- or full-supervision by feeding the generated samples and computing the standard losses on the corresponding original images. Doing so enables a single model to recover information across diverse conditions without modifications at inference time. Extensive experiments on two challenging public datasets, namely nuScenes and Oxford RobotCar, demonstrate the effectiveness of our techniques, outperforming prior works by a large margin in both standard and challenging conditions. Source code and data are available at: \href{https://md4all.github.io/}{https://md4all.github.io}.
\end{abstract}
%This is a problem regardless of the type of supervision, as also LiDAR-supervised models learn artifacts which are systematic in such adverse conditions.

%%%%%%%%% BODY TEXT
\section{Introduction}
% the importance of depth estimation for AD and robotics
Estimating the depth of a scene is a fundamental task for autonomous driving and robotics navigation.
% the difficulty to obtain valid ground truth (expensive and time consuming), requiring 3D sensors (e.g., LiDAR), which are not readily available
While supervised monocular depth estimation approaches have achieved remarkable results, they rely on ground truth data which is expensive and time-consuming to produce~\cite{jung2022hammer,guizilini2020packnet}. This requires costly 3D sensors (e.g., LiDAR) and significant additional data processing~\cite{jung2022hammer,guizilini2020packnet}.

% self-supervised as a solution
To circumvent these issues, geometrical constraints on stereo pairs or monocular videos have been widely explored to learn depth estimation in a self-supervised manner~\cite{godard2019monodepth2,petrovai2022pseudolabels_depth,tosi2019stereo,garg2016unsupervised_stereo,godard2017unsupervised_stereo}.
% stereo and mono
Monocular training solutions are the most inexpensive and rely on the smallest amount of assumptions on the sensor setup, as they require only image sequences captured by a single camera.
% the basic idea of self-sup. mono
% pros and cons
% the inherent issues of it that prevent its wide applicability
%However, monocular depth estimation is an ill-posed problem as there are infinite 3D scenes that correspond to the same projection on the 2D image plane.
%Methods learn from image sequences and they estimate the camera poses across the frames.

\begin{figure}[t]
\begin{center}
\includegraphics[width=1.00\linewidth]{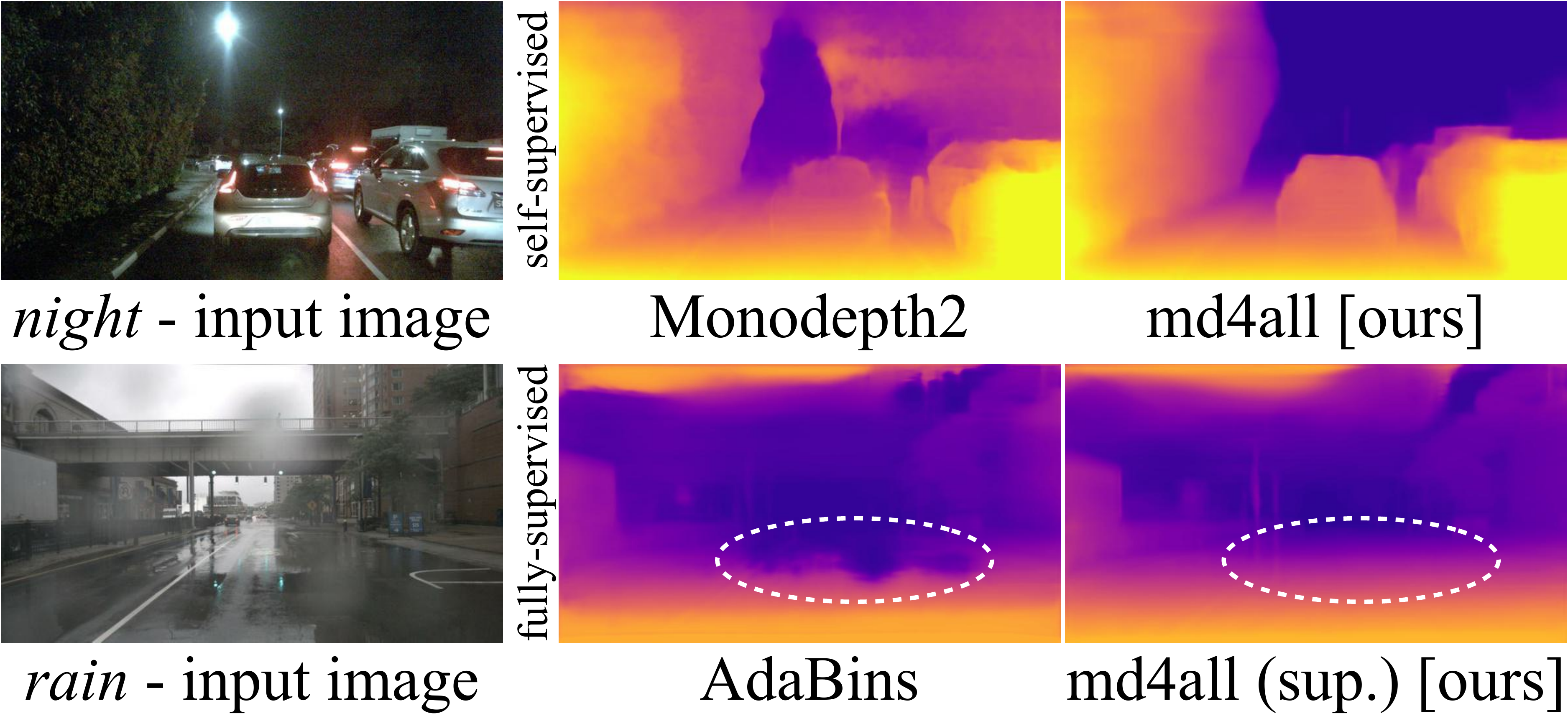}
\vspace{-0.7cm}
\end{center}
   \caption{Predictions in challenging settings~\cite{caesar2020nuscenes} for self-supervised \cite{godard2019monodepth2} and supervised~\cite{bhat2021adabins} methods. Standard approaches fail due to training assumptions or sensor artifacts. Under both supervisions, our \pname\ makes the same models robust in all conditions.}
\label{fig:teaser}
\vspace{-0.2cm}
\end{figure}

% scale ambiguity -> packnet
% dynamic objects -> R4Dyn etc
% darkness and adverse conditions
Self-supervised methods rely on photometric assumptions and pixel correspondences~\cite{godard2019monodepth2,tosi2019stereo}. State-of-the-art approaches~\cite{godard2019monodepth2,yan2021channel,song2021mlda} deliver sharp and accurate estimates in standard conditions (i.e., sunny and cloudy), but suffer from a variety of inherent issues, such as scale ambiguity and difficulties with dynamic objects. While prior works have already proposed robust methods to address these problems~\cite{guizilini2020packnet,gasperini2021r4dyn}, there is still a major issue preventing the wide applicability of self-supervised depth estimators in safety-critical settings, such as autonomous driving. Darkness and adverse weather conditions (e.g., night, rain, snow, and fog) introduce noise in the pixel correspondences. As displayed in Figure~\ref{fig:teaser}, this is detrimental to the effectiveness of such methods, thereby requiring ad hoc solutions.

\begin{figure}[t]
\begin{center}
\includegraphics[width=1.00\linewidth]{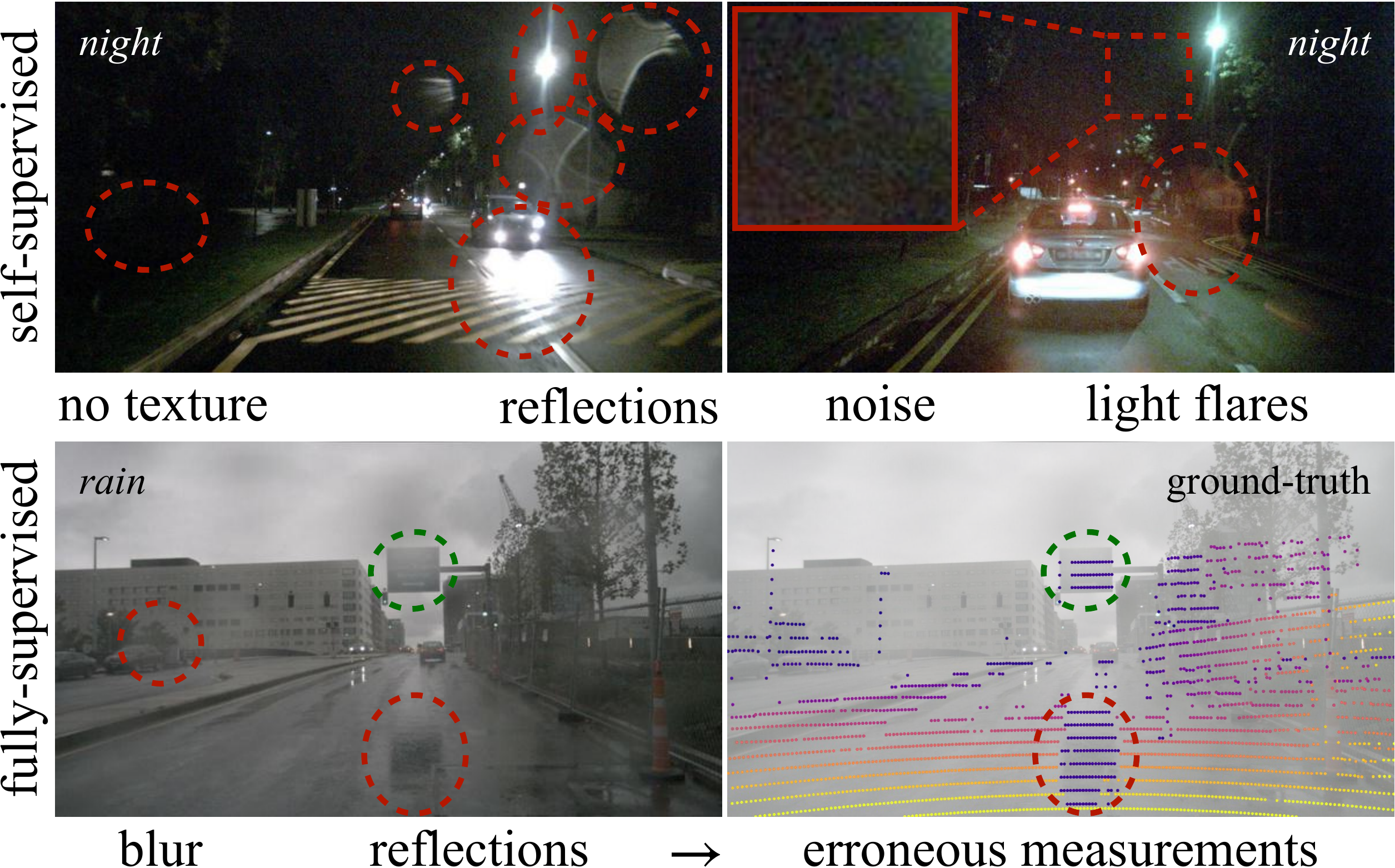}
\vspace{-0.7cm}
\end{center}
   \caption{Detrimental factors to monocular depth estimation in difficult settings from nuScenes~\cite{caesar2020nuscenes}. Self-supervised works have issues with textureless areas, reflections, and noise. Supervised ones learn artifacts from the ground truth sensor (LiDAR is shown).}
\label{fig:factors}
\vspace{-0.2cm}
\end{figure}

As shown in Figure~\ref{fig:factors}, this problem is particularly severe at nighttime due to reflections (e.g., caused by streetlights and vehicle headlights), noise, and the general inability of the embedded cameras to capture details in dark areas. This leads to wrong depth estimates, which can be dangerous in safety-critical settings.
A few pioneering works have already explored this problem, albeit with highly-complex pipelines and significant architecture changes affecting inference as well~\cite{wang2021rnw,vankadari2022sundown,liu2021allday,spencer2020defeatnet,vankadari2020unsupervised_night}, such as illumination-specific branches. Additionally, prior methods that can operate both at night- and daytime introduce a significant trade-off concerning the standard daytime performance~\cite{liu2021allday,vankadari2022sundown}, highlighting the need for a new solution.

% rain and supervised
In adverse weather conditions such as rain, monocular models are similarly fooled by reflections and decreased visibility. However, rain introduces another problem. While radars are robust in such conditions, LiDARs become unreliable, as they introduce multi-path and the so-called blooming effects (Figure~\ref{fig:factors}). In autonomous driving, since supervised depth estimation approaches learn from LiDAR data, this causes them to learn also such erroneous measurements, rendering them unreliable in rainy settings (Figure~\ref{fig:teaser}). Analogous issues occur with snow and fog. These problems are relatively unexplored, demanding new solutions.

Alarmingly, no general solution currently allows an image-based depth estimator to work reliably under all conditions.
%This problem is even more severe as it occurs regardless of the type of supervision. 
Since LiDAR can constitute a misleading training signal in adverse weather, and pixel correspondences are problematic too (e.g., at night), neither existing supervised~\cite{patil2022p3depth} nor self-supervised~\cite{godard2019monodepth2,tosi2019stereo} techniques work well in such challenging settings. A straightforward solution for the supervised case would be using synthetic data~\cite{xiong2022robust_fog,shi2023even}, as by simply not modeling the sensor issues, a simulator could produce perfect ground truth in adverse weather. However, this is not only unexplored, but it would introduce a series of problems, such as a substantial syn2real gap due to the difficulty of modeling challenging conditions realistically (requiring, e.g., domain adaptation).

% in this paper, ...
In this paper, we address these open issues with a simple and effective solution that works reliably in a variety of conditions and for multiple types of supervision. We approach this challenging problem by considering the success of existing methods in standard illumination and weather settings~\cite{godard2019monodepth2,guizilini2020semantic_depth,guizilini2020packnet,gasperini2021r4dyn}. This motivated us to find a way for them to work also under challenging scenarios, exploiting what makes them learn depth effectively in ideal conditions. Our core idea is based on training the model by providing always valid training signals as if it was sunny or cloudy, even when samples with adverse conditions are given.
We apply this general principle to both supervised and self-supervised depth estimation via a set of techniques to improve the model robustness and reduce the performance gap between standard and hard conditions.
The main contributions of this paper can be summarized as follows:
\begin{itemize}
    \item We show how estimating depth in adverse conditions (e.g., night and rain) is problematic for both self- and fully-supervised approaches, requiring new solutions.%even for fully-supervised approaches, due to the LiDAR inherent issues.
    \item We propose \pname: a simple and effective technique to make standard models robust in diverse conditions.
    \item We apply our generic method to both fully- and self-supervised monocular settings.
    \item We generate and share open-source images in adverse conditions corresponding to the sunny and cloudy samples of nuScenes~\cite{caesar2020nuscenes} and Oxford Robotcar~\cite{maddern2017oxford}.
\end{itemize}
With \pname, we substantially outperform prior solutions delivering robust estimates in a variety of conditions.

\section{Related Work}
\label{sec:related_work}
\subsection{Supervised Monocular Depth Estimation}\label{sec:rw_supervised}
The problem of estimating depth from a single color image is challenging due to the countless 3D scenarios that can produce the same 2D projection, making it an ill-posed problem. Nevertheless, significant progress has been made, thanks to the introduction of CNN-based architectures by Eigen et al.~\cite{eigen2014depth} and fully-convolutional networks with residual connections by Laina et al.~\cite{laina2016deeper} to estimate dense depth maps from monocular inputs. While many supervised methods have focused on directly regressing to depth measurements from LiDAR sensors (as in KITTI~\cite{geiger2013kitti}) or RGB-D cameras (as in NYU-Depth v2~\cite{silberman2012nyu}), DORN~\cite{fu2018dorn} tackles the task in an ordinal manner. AdaBins~\cite{bhat2021adabins} extended DORN via a linear combination of predictions across adaptive bins.
Moreover, BTS uses a multi-stage local planar guidance~\cite{lee2019bigtosmall} and P3Depth exploits coplanar pixels~\cite{patil2022p3depth}.
Others investigated the benefit of depth estimation while tackling other tasks, such as 3D object detection~\cite{huang2022monodtr}.

\textbf{Issues} While the supervision signal from 3D sensors is reliable in ideal conditions (e.g., sunny, cloudy), it severely degrades in photometrically challenging scenarios~\cite{jung2022hammer}. Outdoor, LiDAR sensors deliver erroneous measurements in adverse weather conditions, such as rain, snow and fog. As Jung et al.~demonstrated indoor \cite{jung2022hammer}, training on an inexact ground truth leads depth models to learn the sensor artifacts and deliver wrong outputs. This problem is relatively unexplored outdoors, e.g., with rain. A few works investigated depth completion in simulated settings with LiDAR and radar in input~\cite{xiong2022robust_fog} or event cameras and RGB~\cite{shi2023even}. In this paper, we explore this issue on AdaBins~\cite{bhat2021adabins} and provide a simple solution to estimate depth reliably in diverse conditions, regardless of the sensor artifacts.

\begin{figure*}[t]
\begin{center}
\includegraphics[width=1.00\textwidth]{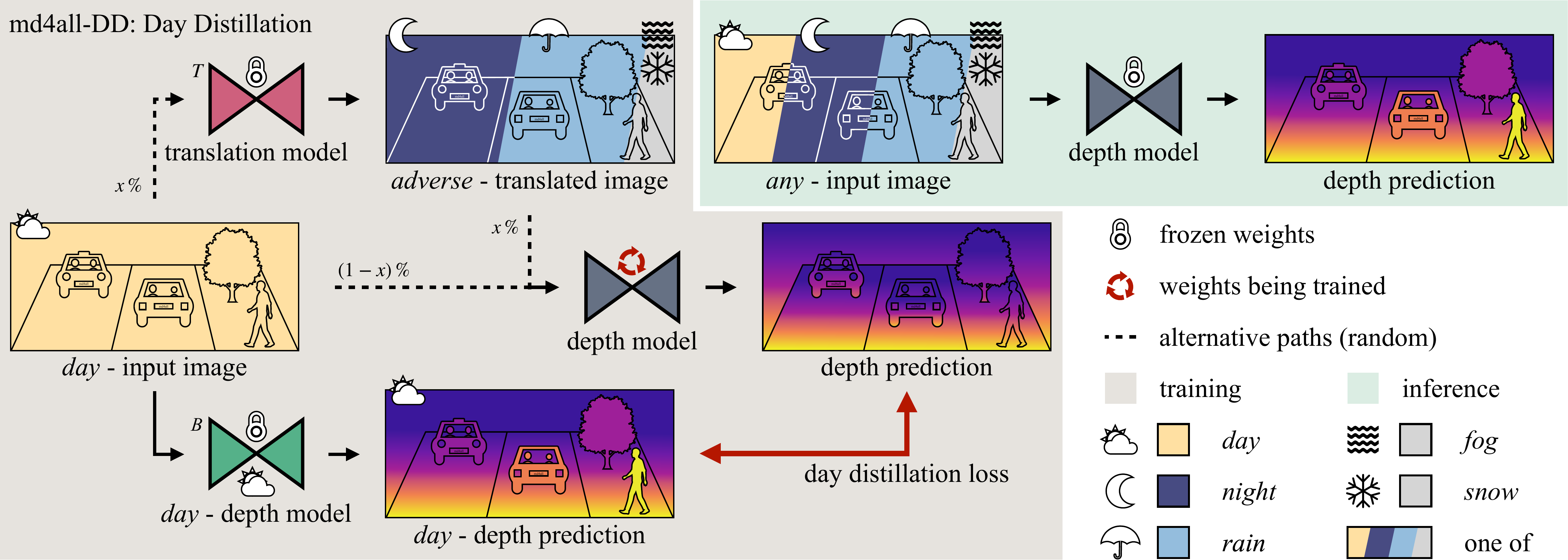}
\vspace{-0.7cm}
\end{center}
   \caption{Our \pname-DD framework. The frozen \textit{day} - depth model estimates on easy samples and provides guidance to another model fed with a mix of easy and translated inputs. Inference is done with a simple single model for both fully- and self-supervised \pname.}
\label{fig:framework-dd}
\vspace{-0.2cm}
\end{figure*}

\subsection{Self-Supervised Monocular Depth Estimation}
To bypass the need for expensive LiDAR data, self-supervised methods employ view reconstruction constraints through stereo pairs~\cite{garg2016unsupervised_stereo,godard2017unsupervised_stereo} or monocular videos~\cite{godard2019monodepth2,zhou2017unsupervised_depth}. The latter utilizes motion parallax from a moving camera in a static environment~\cite{ullman1979interpretation} and requires simultaneous depth and camera pose transformation prediction. Significant advancements have been made since Zhou et al.'s pioneering video-based approach~\cite{zhou2017unsupervised_depth}, including novel loss terms~\cite{godard2019monodepth2}, network architectures that preserve details~\cite{guizilini2020packnet}, the use of cross-task dependencies~\cite{jiao2018lookdeeper,guizilini2020semantic_depth}, pseudo labels~\cite{petrovai2022pseudolabels_depth}, vision transformers~\cite{zhao2022monovit}, uncertainty estimation~\cite{poggi2020uncertainty_depth}, and 360 degrees depth predictions~\cite{guizilini2022surround_depth}.

\subsubsection{Solutions to Inherent Issues}\label{sec:rw_self-sup_issues}
\textbf{Scale ambiguity}
%Due to the infinite number of 3D scenes corresponding to the same 2D projection, 
Video-based methods predict depth up to scale, requiring median-scaling with ground truth data at test time~\cite{godard2019monodepth2}. Guizilini et al.~\cite{guizilini2020packnet} used the readily available odometry information to achieve scale awareness via weak velocity supervision on the pose transformation.

\textbf{Dynamic scenes}
Due to the moving camera in a static world assumption~\cite{ullman1979interpretation}, video-based methods have issues with dynamic objects, e.g., cars. To address this, Monodepth2~\cite{godard2019monodepth2} uses an auto-masking loss on the static pixels, R4Dyn~\cite{gasperini2021r4dyn} adds weak radar supervision on the objects, and DRAFT~\cite{guizilini2022depth_flow} combines optical and scene flows.

\textbf{Darkness}
Low visibility is detrimental to the losses used to learn depth because noise and lack of details prevent establishing pixel correspondences across the frames.
DeFeat-Net~\cite{spencer2020defeatnet} was among the first to mitigate this, with a cross-domain dense feature representation. ADFA~\cite{vankadari2020unsupervised_night} uses a generative adversarial network (GAN) to adapt nighttime features to daytime ones.
R4Dyn~\cite{gasperini2021r4dyn} shows that radar is beneficial not only for dynamic objects but also at nighttime as a byproduct.
RNW~\cite{wang2021rnw} reduces the irregularities at nighttime via, e.g., image enhancement and a GAN-based regularizer.
ADIDS~\cite{liu2021allday} uses separate networks for day and night images, partially sharing weights.
ITDFA~\cite{zhao2022depth_night_rain} is similar to ADFA, doing feature adaptation from night to day, with images generated with a GAN.
WSGD~\cite{vankadari2022sundown} combines denoising with a lighting change decoder to predict per-pixel changes.
While these works made significant steps towards solving the problem, they either have complex pipelines with dedicated branches for day and night~\cite{liu2021allday,zhao2022depth_night_rain}, use additional sensors~\cite{gasperini2021r4dyn}, suffer from a significant trade-off on the daytime performance~\cite{vankadari2022sundown}, or are not meant to operate on multiple conditions, such as both day and night~\cite{vankadari2020unsupervised_night,wang2021rnw,zhao2022depth_night_rain}. Therefore, an effective solution without inference complications is yet to be found.

\textbf{Adverse weather}
As at nighttime, in adverse weather such as rain, fog, and snow, the limited visibility prevents establishing correct correspondences. Even fully-supervised approaches have issues in these settings~\cite{jung2022hammer}.
So far, only a handful of works have explored depth estimation with adverse weather. ITDFA~\cite{zhao2022depth_night_rain} requires an encoder for each condition and was not shown to work in both standard and adverse settings. R4Dyn~\cite{gasperini2021r4dyn} and MonoViT~\cite{zhao2022monovit} are robust methods that delivered improvements also in adverse conditions as a side effect. Thus, this problem is largely unexplored, demanding a general solution.

Unlike prior works, in this paper, we propose a simple and effective solution enabling a standard monocular model to estimate depth in diverse conditions (e.g., day, night, and rain) without any difference at inference time compared to a common encoder-decoder pipeline~\cite{godard2019monodepth2}. Additionally, ours does not degrade the output quality in standard settings.

\section{Method}
\label{sec:method}

% overview (1 paragraph)
In this paper, we enable a model to estimate depth reliably in diverse conditions (e.g., day, night, and rain). Displayed in Figures~\ref{fig:framework-dd} and~\ref{fig:framework-ad}, our techniques exploit the effectiveness of existing approaches in standard conditions (e.g., daytime in good weather) to increase their robustness in adverse settings. Towards this end, we perform day-to-adverse image translation, train on the generated adverse samples, and learn only from valid training signals from the original day inputs. This simple idea is suitable to both self-supervised (Section~\ref{sec:method_self-sup}) and supervised (Section~\ref{sec:method_supervised}) frameworks and is general to operate under various weather and illumination settings (including fog and snow).

% Darkness and adverse weather -> why are they an issue

\subsection{\pname\ - Self-Supervised}\label{sec:method_self-sup}
We build upon a scale-aware video-based monocular method (Section~\ref{sec:self-sup_framework}). 
As described in Section~\ref{sec:rw_self-sup_issues}, night and bad weather cause issues to self-supervised approaches. We address this with \pname by computing the losses only on the ideal samples corresponding to the hard ones given as input (Section~\ref{sec:method-ad}). We then take this concept even further by distilling knowledge from a frozen self-supervised model trained only on the ideal samples (Section~\ref{sec:method-dd}).

\subsubsection{Self-Supervised Baseline}\label{sec:self-sup_framework}
\begin{figure*}[t]
\begin{center}
\includegraphics[width=1.00\textwidth]{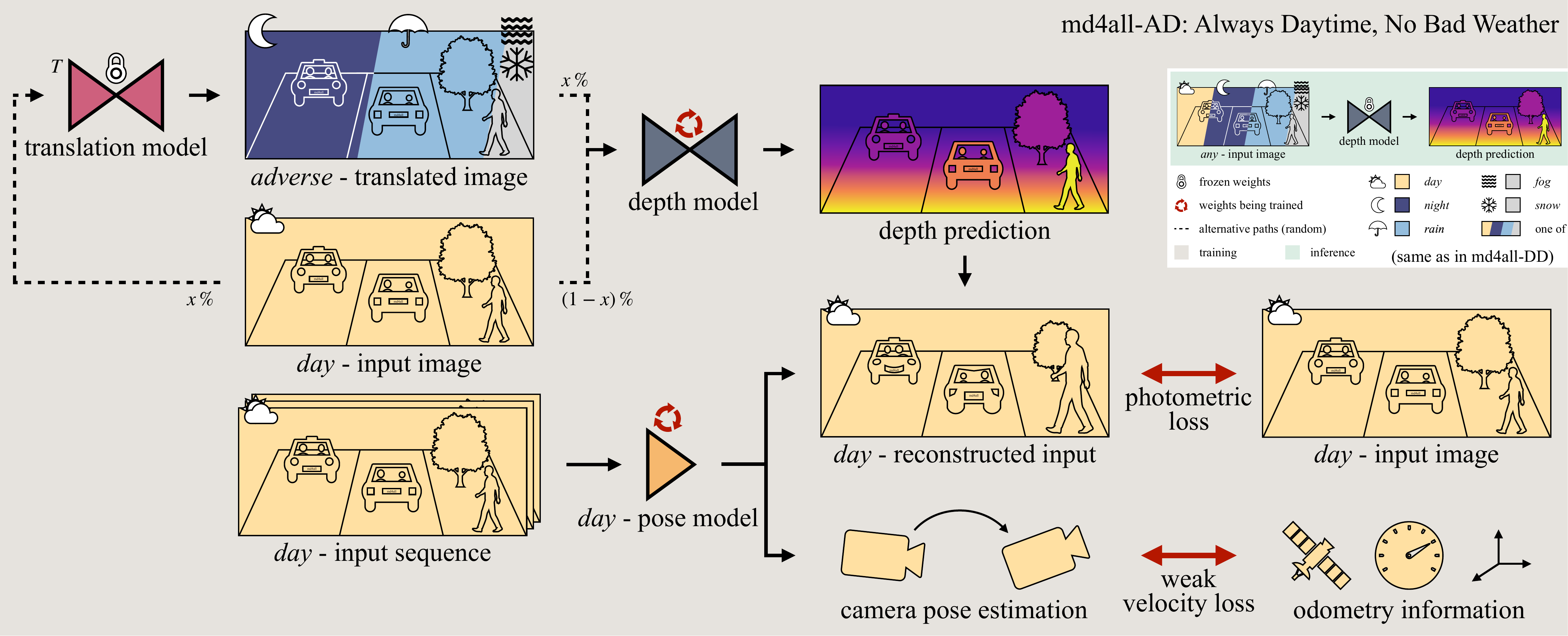}
\vspace{-0.7cm}
\end{center}
   \caption{Our self-supervised \pname-AD framework. With $x=0$, it is equivalent to the \textit{day} - depth model in Figure~\ref{fig:framework-dd} and the baseline. The depth model is trained with a mix of easy and translated samples, while the training signal is always from the easy ones.}
\label{fig:framework-ad}
\vspace{-0.2cm}
\end{figure*}

We build on a standard video-based monocular depth baseline equivalent to the framework shown in Figure~\ref{fig:framework-ad} when considering $x=0$ (i.e., no translation).
%$T_{t \to s}$ 
We predict both the depth $\hat{D}_t$ of a target frame and the pose transformations between the target $I_t$ and source frames $I_{s \in \{t-1, t+1\}}$, with which we warp the source into a reconstructed target view. As in~\cite{godard2019monodepth2,guizilini2020packnet}, a loss is computed on the appearance shift between $I_t$ and the reconstruction~\cite{zhou2017unsupervised_depth}, alongside the structural similarity~\cite{wang2004image}. Following~\cite{godard2019monodepth2}, we account for partial occlusions via the minimum reprojection error $\mathcal{L}_{p}$, and we ignore static pixels.
Another loss $\mathcal{L}_{s}$ promotes smoothness and preserves edges~\cite{godard2017unsupervised_stereo}.
$\mathcal{L}_{p}$ and $\mathcal{L}_{s}$ are calculated at all decoder scales, upsampled to the input size~\cite{godard2019monodepth2}.

So far, this is equivalent to Monodepth2~\cite{godard2019monodepth2}. Then, we add the weak velocity supervision $\mathcal{L}_{v}$ to achieve scale-awareness~\cite{guizilini2020packnet} and allow consistent predictions, beneficial when distilling knowledge between different models.

\textbf{Architecture} Unlike previous works having specialized branches~\cite{liu2021allday,zhao2022depth_night_rain}, we leave the architecture unchanged (e.g.,~\cite{godard2019monodepth2}). Instead, we act on the training process. Our approach is general and not bound to a specific architecture.

\subsubsection{\pname-AD: Always Daytime, No Bad Weather}\label{sec:method-ad}
Our \pname-AD configuration is shown in Figure~\ref{fig:framework-ad}. The core idea is learning from easy samples, even when given challenging ones (e.g., night) as if it was always daytime with good visibility (i.e., sunny or cloudy). This allows using the same established losses described in Section~\ref{sec:self-sup_framework}, which would otherwise fail with difficult inputs.

\textbf{Day-to-adverse translation} To achieve the above, we need easy samples corresponding to the challenging ones. This means having paired images ($e_i$, $h_i^c$), with $e_i \in E$ and $E$ being the set of easy samples (i.e., sunny or cloudy), $h_i^c \in H$ with $H$ the set of the difficult samples from the conditions of interest $c \in C$ (e.g., snow). While an image translation method could convert the training $H$ into easy ones, removing information is easier than adding it. Therefore, we generate $H$ from $E$ (e.g., turning $e_i$ into nighttime). Specifically, for each $e_i$ and each condition $c$ we aim to improve (e.g., night and rain), we obtain $h_i^c = T^c(e_i)$. We do this with $c$ image translation models $T^c$ trained at an earlier stage, increasing the training set size by $C\times E$.

\textbf{Training scheme} We then train depth and pose models as shown in Figure~\ref{fig:framework-ad}. During training, we feed to the depth model $m_i$, which is either $h_i^c$ (for $x\%$ of the inputs, as a random mix of $c$) or $e_i$ from the pre-existing training data. Additionally, we normalize the inputs depending on the recording time (i.e., day/night) to learn robust features agnostic of the input condition. The Appendix shows how performing this step only during training delivers similar results. Then, in the case of particularly noisy night samples (e.g., nuScenes~\cite{caesar2020nuscenes}), we augment the inputs with heavy noise. The pose model always takes the sequence $[e_{i-1},e_i,e_{i+1}]$, corresponding to $m_i$. If fed $h_i^c$, the pose network would have issues assessing the pixel correspondences.

\textbf{Learning in all conditions} Computing the losses $\mathcal{L}_{p}$ and $\mathcal{L}_{s}$ on $h_i^c$ would lead to issues because of the difficulty of establishing correspondences in adverse conditions (Section~\ref{sec:rw_self-sup_issues}). For this reason, training on $E$ and deploying on $H$ is more effective than training on both (Section~\ref{sec:quantitative}), proving the limitations of standard methods. Our solution to this challenging problem is relatively simple: as shown in the figure, we provide a reliable training signal by always calculating the losses on $E$. Specifically, they are always computed on $e_i$, even when the depth model is fed with $h_i^c$ ($x\%$). This constructed setting constitutes the ideal condition in which the losses $\mathcal{L}_{p}$ and $\mathcal{L}_{s}$ are already proven successful~\cite{godard2019monodepth2}, eliminating the source of the issues. This leads the depth model to learn to extract robust features, regardless of whether the input belongs to $E$ or $H$.

\textbf{Inference} After training depth and pose models, the latter is discarded, while our depth model is a simple encoder-decoder capable of estimating depth in multiple conditions. As shown at the top of Figure~\ref{fig:framework-dd}, since we do not apply any architectural modification, at inference time, we predict depth with the same model through the same model parameters, regardless of the input condition. While dedicated models or branches may lead to better performance, switching between them is not always trivial, e.g., at dusk or with light rain. Therefore, we opted for a single monocular model, which does not penalize inference time compared to the same model trained only on $E$.

\subsubsection{\pname-DD: Day Distillation}\label{sec:method-dd}
We take \pname-AD (Section~\ref{sec:method-ad}) to the next level by simplifying the training scheme with \pname-DD. The core idea of \pname-DD is the same as for \pname-AD: we aim to learn depth only from $E$, pretending that the conditions $C$ detrimental for the losses never occur.

Our \pname-DD framework mimics model estimates in ideal settings $E$, regardless of the difficulty of the input. As shown in Figure~\ref{fig:framework-dd}, we achieve this via knowledge distillation from a depth network $B$ (baseline) trained at an earlier stage on $E$ to a new depth model $DD$ for both easy and adverse scenarios (i.e., $E$ and $H$). The latter is fed $m_i$, i.e., the same mix of $e_i$ and $h_i^c$ as in \pname-AD (Section~\ref{sec:method-ad}), while the former is given only $e_i$. $DD$ is optimized solely through the following objective:
\begin{equation}
   \mathcal{L}_{d} = \frac{1}{N} \sum_{j=1}^N \frac{|DD(m_i)_j - B(e_i)_j|}{DD(m_i)_j}
\end{equation}
where $N$ is the number of pixels, $DD(m_i)$ is $DD$'s depth prediction on $m_i$ (i.e., an easy or hard sample), and $B(e_i)$ is $B$'s estimation on $e_i$ (i.e., an easy sample).
$DD$ learns to follow $B$ at the output level, even when fed the problematic $h_i^c$, without being affected by the detrimental factors occurring in adverse settings.
Inference is unchanged.

\input{tables/main_nuscenes.tex}

\subsection{\pname\ - Supervised}\label{sec:method_supervised}
Learning depth from a 3D sensor in adverse conditions exposes issues inherent to the sensor and the way it measures depth~\cite{jung2022hammer}. With bad weather (e.g., rain), LiDARs provide erroneous measurements (Figure~\ref{fig:factors}), so learning from their signal means copying their artifacts as well (Figure~\ref{fig:teaser}). This has been ignored so far for monocular depth.

Regardless of the input, we address the sensor issues by learning from $E$. Analogously to the self-supervised setting, we use image pairs ($e_i$, $h_i^c$) and specify our method as \pname-AD following the self-supervised definition (Section~\ref{sec:method-ad}), except for the supervision signal. Thus, we train the depth model with $m_i$ and learn from the LiDAR signal of $e_i$. 
Thus, the supervision is from artifact-free data in ideal conditions $E$, such that the models never experiences the sensor issues. As in the self-supervised setup, the inference is unchanged.
%With \pname-DD, we distill knowledge from $B$ trained at an earlier stage on $E$ (always estimating depth on $e_i$), to a new model $DD$ fed with $s_i$ (Figure~\ref{fig:framework-dd}). In both configurations, 
While \pname-DD (Section~\ref{sec:method-dd}) also applies to the supervised case, using AD is more reasonable since reliable ground truth data from $e_i$ is available.

\section{Experiments and Results}
\label{sec:exp_results}

\subsection{Experimental Setup}
\label{sec:setup}

\textbf{Datasets and metrics}
%KITTI~\cite{geiger2013kitti} does not contain any scene at night or in adverse weather, so w
We used two public driving datasets containing various illumination and weather conditions: nuScenes~\cite{caesar2020nuscenes} and Oxford RobotCar~\cite{maddern2017oxford}.
\textbf{nuScenes} is a challenging large-scale dataset with 15h of driving in Boston and Singapore, diverse scenes, and difficult conditions. We distinguished good visibility (i.e., \textit{day-clear}), \textit{night} (including \textit{night-rain}), and \textit{day-rain}. We used the official split following R4Dyn~\cite{gasperini2021r4dyn}, with 15129 training images (with synced sensors), and 6019 validation ones (of which 4449 \textit{day-clear}, 602 \textit{night}, and 1088 \textit{rain}).
\textbf{RobotCar} was collected in Oxford, UK, by traversing the same route multiple times in a year. It features a mix of \textit{day} and \textit{night} scenes. We followed the split and setup of WSGD~\cite{vankadari2022sundown}, with 16563 \textit{day} training samples and 1411 test ones (with synced sensors, of which 709 \textit{night}).
While we focused on night, rain, sun, and overcast, the Appendix shows preliminary results with fog and snow from the \textbf{DENSE} dataset~\cite{bijelic2020dense}.
We report on the standard metrics and errors up to 50m for RobotCar as in~\cite{vankadari2022sundown}, and 80m for nuScenes as in~\cite{gasperini2021r4dyn}. More results can be found in the Appendix.

\input{tables/main_robotcar.tex}

\textbf{Implementation details}
Our self-supervised models use a ResNet-18 backbone~\cite{he2016resnet} and learn from an image triplet sized 576x320 for nuScenes and 544x320 for RobotCar. The supervised model and \pname-DD are given only one keyframe. At inference time, all models take a single RGB input. We set $x=|C|/(|C|+1)~\%$, with $|C|$ being the number of the adverse conditions of interest $C$, e.g., $x=66\%$ for a model to work with \textit{rain}, \textit{night} and \textit{day}, and within $x\%$ we used equally distributed data among $C$. So, our models see an equal amount of inputs for each condition. We used the same hyperparameters as Monodepth2~\cite{godard2019monodepth2} and AdaBins~\cite{bhat2021adabins} for self- and fully-supervised models, respectively.
All models were trained on a single 24GB GPU.
%Inference of our \pname\ self-supervised takes \todo{XYZ} ms.

\textbf{Image translation}
We translated each $e_i$ image to $h_i^c$.
Diffusion models~\cite{rombach2022diffusion,saharia2022palettenet} are not suitable due to the lack of already paired images. Datasets with multiple drives on the same roads~\cite{diaz2022ithaca365,sakaridis2021acdc,maddern2017oxford} do not solve this issue due to the lack of synchronization and environmental changes.
So we opted for GANs. For each condition $c$, we used a ForkGAN model~\cite{zheng2020forkgan} $T^c$ to translate all \textit{day-clear} training samples $E$ of nuScenes, with $c \in C=\left \{ \text{\textit{night},~\textit{rain}} \right \}$. We trained ForkGAN on BDD100K~\cite{yu2020bdd100k} and fine-tuned it on the nuScenes training set.
For RobotCar, we used $T^c$ to translate all \textit{day} samples $E$ into \textit{night} ones. RobotCar contains more \textit{night} samples than nuScenes, so we trained $T^c$ directly on its training set.
We share publicly all generated $h_i^c$ images.

\textbf{Prior works and baselines} We compared ours with a variety of works~\cite{gasperini2021r4dyn,godard2019monodepth2,guizilini2020packnet,wang2021rnw,vankadari2022sundown,liu2021allday,spencer2020defeatnet,bhat2021adabins}.
We applied ours on the self-supervised Monodepth2-based baseline of Section~\ref{sec:self-sup_framework} and the fully-supervised AdaBins~\cite{bhat2021adabins}.

\subsection{Quantitative Results}
\label{sec:quantitative}

\textbf{\textit{Night} -- nuScenes}
In Table~\ref{table:main_nuscenes}, we report results for nuScenes~\cite{caesar2020nuscenes} across various settings. Night samples present strong noise levels and reflections that are detrimental for self-supervised models (Figure~\ref{fig:factors}), causing the absRel errors of most methods to double from ideal conditions (i.e., \textit{day-clear}) to \textit{night}. The difficulty of learning from night inputs is evident comparing Monodepth2~\cite{godard2019monodepth2} trained only on \textit{day-clear} (\textit{d}) against \textit{all} conditions (\textit{a}), with the latter severely underperforming. PackNet~\cite{guizilini2020packnet} improved at \textit{night} and \textit{rain}, albeit doing worse in standard settings, possibly due to its large model and the relatively small dataset. PackNet's velocity supervision also helped over Monodepth2 (md2) with our baseline.
Thanks to the extra radar signal, R4Dyn~\cite{gasperini2021r4dyn} delivered significant improvements, although at \textit{night}, only adding radar in input was beneficial over md2. md2 trained only on \textit{day-clear} data outperformed RNW's complex pipeline~\cite{wang2021rnw}. We retrained RNW on the official split (the authors reported an absRel of 0.3150 at \textit{night} on their split~\cite{wang2021rnw}).
Remarkably, despite being based on the same model as md2, at \textit{night}, our simple techniques reduced absRel by 32\% and relatively increased $\delta_1$ by 37\% (DD). Our \pname\ also outperformed the radar-based R4Dyn at \textit{night}. This is thanks to the ability of our method to extract robust features from monocular data even in the dark.

\textbf{\textit{Night} -- RobotCar}
In Table~\ref{table:main_robotcar}, we report results for RobotCar~\cite{maddern2017oxford}. Here we compare with various approaches that also target depth estimation in challenging conditions~\cite{spencer2020defeatnet,liu2021allday,wang2021rnw,vankadari2022sundown}. They all focus on \textit{night} issues, tested here. Our \pname\ outperforms them all across the board, with substantially better estimates at \textit{night} than theirs during the \textit{day}: the previous best WSGD~\cite{vankadari2022sundown}'s \textit{day} absRel error is 45\% higher than ours at \textit{night}. This is thanks to the simplicity of our approach, which does not rely on complex architectures, but makes existing models robust in adverse conditions by changing their input and training signals.

\textbf{\textit{Rain} -- nuScenes}
Rain is less problematic than darkness due to the lack of cues in the latter.
Results are shown in Table~\ref{table:main_nuscenes}, with all methods performing better with \textit{rain} than at \textit{night}.
Our self-supervised monocular \pname-DD significantly improved over Monodepth2 and the baseline, performing close to the radar-based R4Dyn~\cite{gasperini2021r4dyn}.

\begin{figure*}[t]
\begin{center}
\includegraphics[width=1.00\textwidth]{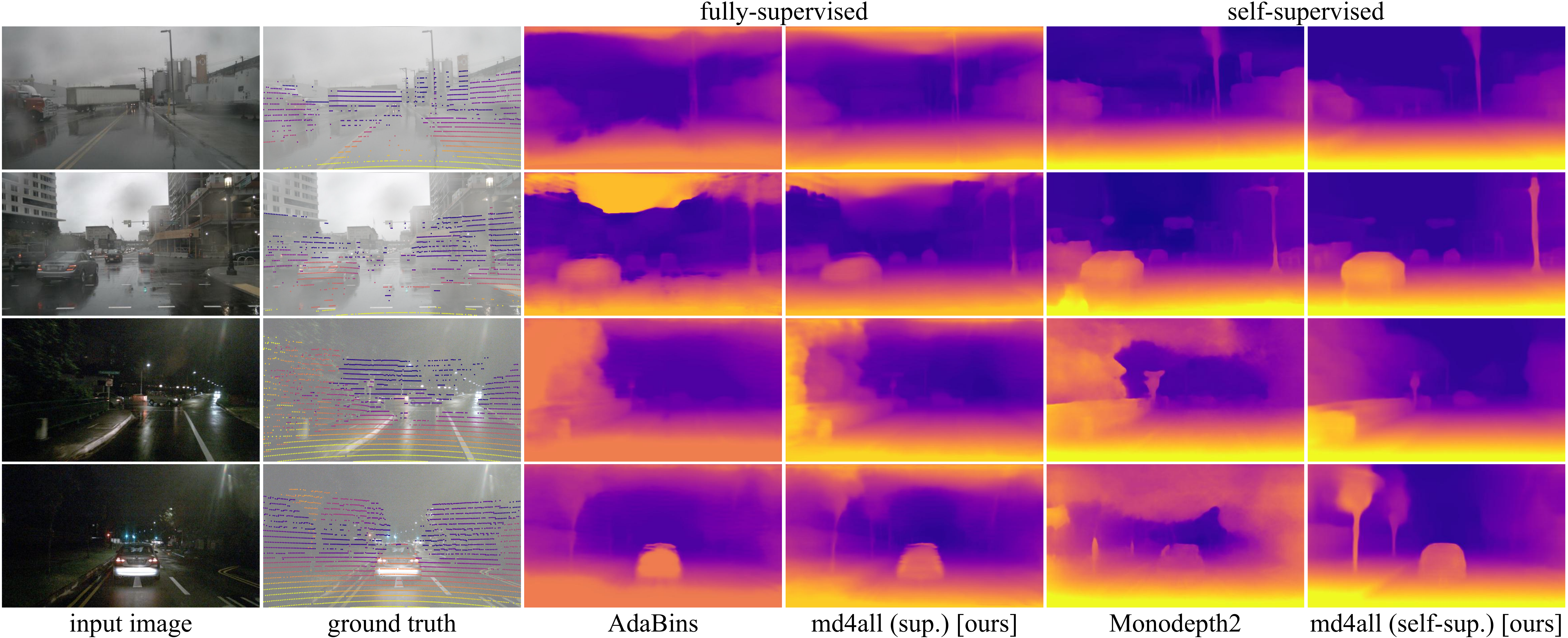}
\vspace{-0.7cm}
\end{center}
   \caption{Comparison on nuScenes~\cite{caesar2020nuscenes} between fully-sup.~AdaBins~\cite{bhat2021adabins} w/o and w/ ours, and self-sup.~Monodepth2~\cite{godard2019monodepth2} w/o and w/ ours.}
\label{fig:qualitative_nuscenes_comp}
\vspace{-0.2cm}
\end{figure*}

\textbf{Fully-supervised}
Table~\ref{table:main_nuscenes} reports also results in supervised settings. LiDAR data is reliable in the dark, so \textit{night} scenes are less of an issue. Instead, \textit{rain} inputs are particularly interesting for supervised works due to the reflection issues shown in Figures~\ref{fig:teaser} and~\ref{fig:factors}. For supervised settings, we applied our method on AdaBins~\cite{bhat2021adabins}.
It is to be considered that LiDAR artifacts may have an impact on the \textit{rain} values, such that perfect estimates would not score perfectly because the ground truth is wrong (Figure~\ref{fig:factors}). So, while we can assess the improvements of \pname\ at handling the blur caused by raindrops, we cannot correctly quantify its impact on eliminating the artifacts. Therefore, these comparisons are more meaningful when considered alongside qualitative outputs (Figure~\ref{fig:qualitative_nuscenes_comp}).
Our supervised \pname\ performed better than AdaBins both quantitatively and qualitatively, eliminating the dependency on the sensor artifacts.
Additionally, thanks to the strong regularization introduced by the translated samples, our model generalizes significantly better than the standard AdaBins, leading to vast improvements across the board, also at \textit{night}. Training on the sparse LiDAR signal of nuScenes~\cite{caesar2020nuscenes} (Figures~\ref{fig:factors} and~\ref{fig:qualitative_nuscenes_comp}) can lead to overfitting. Ours is a beneficial data augmentation technique, adding diversity to the training, as the model is shown $|C|+1$ variations of each \textit{day-clear} input.

\begin{figure}[b]
\vspace{-0.2cm}
\begin{center}
\includegraphics[width=1.00\linewidth]{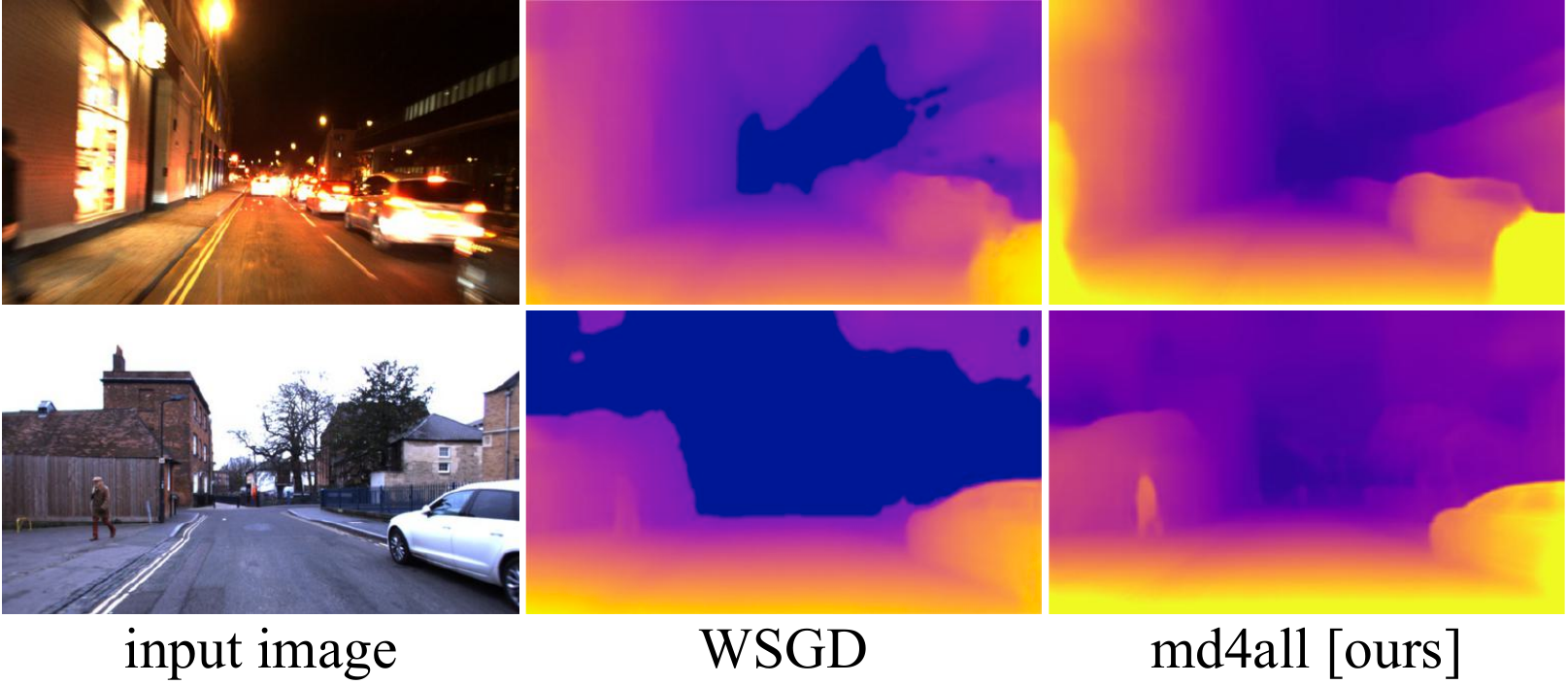}
\vspace{-0.7cm}
\end{center}
   \caption{Comparison on RobotCar~\cite{maddern2017oxford} samples between ours self-supervised and WSGD~\cite{vankadari2022sundown}. Ouputs of WSGD are from~\cite{vankadari2022sundown}.}
\label{fig:qualitative_robotcar_comp}
%\vspace{-0.2cm}
\end{figure}

\input{tables/GANs.tex}

\textbf{\textit{Day(-clear)}}
While we do not include any modification addressing standard conditions, we still see improvements over the baselines across both datasets and supervision types (Tables~\ref{table:main_nuscenes} and~\ref{table:main_robotcar}). This is due to the training mix of easy and translated samples acting as a strong data augmentation and regularization technique. Since the same weights are optimized on all conditions, they learn to extract robust features which are beneficial also with good visibility. Instead, RNW~\cite{wang2021rnw} is meant for operating only at night.
%, delivering sub-optimal estimates in other conditions.

\textbf{All conditions}
Remarkably, across all tested conditions, \pname\ significantly improves over Monodepth2 and AdaBins on which we applied it, without the need for specialized branches (Tables~\ref{table:main_nuscenes} and~\ref{table:main_robotcar}). There is no trade-off introduced when training our unique \pname\ model for multiple conditions, as the scores and errors remain equivalent or even improve compared to training only in ideal settings. This proves the effectiveness and generality of our simple ideas.
The Appendix includes preliminary results with \textit{snow} and \textit{fog} on the challenging DENSE dataset~\cite{bijelic2020dense}.
%, showing the robustness of \pname\ across a variety of adverse conditions.
%Although not tested due to the lack of data, \pname\ should be suitable also for other adverse conditions, such as \textit{snow} and \textit{fog}.

\textbf{AD and DD} %AD best in supervised, DD best in self-supervised
Our \pname\ delivers improvements both as DD and AD (Tables~\ref{table:main_nuscenes} and~\ref{table:main_robotcar}). While the two are applicable under both supervisions, available and reliable ground truth alongside the \textit{day-clear} data makes AD more suitable for supervised setups. DD works better than AD in self-supervised settings thanks to the simplified training scheme and the guidance of our strong baseline.

\textbf{Robustness against translations}
In Table~\ref{table:GANs}, we assess the impact of the quality of image translation on our method.
While the selected ForkGAN~\cite{zheng2020forkgan} translates better than CycleGAN~\cite{zhu2017cyclegan}, it does not give perfect outputs either (Appendix).
Since we use the translations to learn robust features, their imperfections even help our model's robustness by making it harder to recover information for the depth task, as the translations act as data augmentation and regularization.
The table confirms the robustness of \pname, performing similarly regardless of which GAN is used, even when degrading 10\% of the inputs via random erasing.

%\textbf{Ablation study}

\subsection{Qualitative Results}
\label{sec:qualitative}
Qualitative comparisons in Figures~\ref{fig:qualitative_nuscenes_comp} and~\ref{fig:qualitative_robotcar_comp} confirm the quantitative findings, with our \pname\ delivering improved estimates in both adverse and standard conditions. On nuScenes~\cite{caesar2020nuscenes} (Figure~\ref{fig:qualitative_nuscenes_comp}), unlike the baselines, both our models correctly identified the truck in the first rainy sample. As shown in Figure~\ref{fig:factors}, rain leads to artifacts in the LiDAR ground truth, which cause the standard fully-supervised AdaBins~\cite{bhat2021adabins} to learn them and estimate the road wrongly. Our supervised \pname\ exhibits no such artifacts as it was not trained with the problematic rainy samples but rather on our translated ones, which have reliable ground truth. Instead, self-supervised methods have issues at \textit{night}. While Monodepth2~\cite{godard2019monodepth2} could identify critical elements of the scenes (e.g., car and sign), its difficulties in extracting information in the dark are evident. Monodepth2 had fewer issues with brighter \textit{night} samples, as shown in the Appendix. Our self-supervised \pname\ delivered sharp estimates, identifying even the two trees on the left side of the bottom input, which are particularly hard to see.
For RobotCar~\cite{maddern2017oxford} (Figure~\ref{fig:qualitative_robotcar_comp}), we compared on the same samples displayed by WSGD in their paper~\cite{vankadari2022sundown}. As in Table~\ref{table:main_robotcar}, our \pname\ delivered better and sharper estimates in both conditions, correctly estimating the people's distance.
%Overall, this proves the effectiveness of our simple techniques at extracting robust features.

\textbf{Limitations} \pname\ improves in all tested conditions, but DD may propagate errors from the baseline. Thus, a stronger baseline would help. Despite the robustness against translations (Table~\ref{table:GANs}), GANs~\cite{zheng2020forkgan} could be problematic. Better translations would help eliminate the domain gap, as seen with RobotCar (Table~\ref{table:main_robotcar}). GANs require many adverse images for training. Hard-to-distinguish data distributions (e.g., light snow vs.~overcast) may create problems. \pname\ is applicable to stereo-based models too, but only given consistent translations for the stereo images. Future work may focus on eliminating the dependency on the GAN. Furthermore, \pname\ does not address the issue of dynamic objects, so flow~\cite{guizilini2022depth_flow} or weak radar supervision~\cite{gasperini2021r4dyn} may be beneficial, albeit adding complexity. The core ideas of this work can be extended to other tasks.

The \textbf{Appendix} includes a variety of extra results, e.g., experiments with \textit{snow} and \textit{fog}, and sample translations.

\section{Conclusion}
We presented the simple and effective \pname, enabling a single monocular model to estimate depth robustly in both standard and challenging conditions (e.g., night, rain). We showed \pname\ delivering significant improvements under both fully- or self-supervised settings, overcoming the detrimental factors that make adverse conditions problematic.

%\clearpage
\input{appendix/appendix_arxiv.tex}

%%%%%%%%% REFERENCES
{\small
\bibliographystyle{ieee_fullname}
\bibliography{egbib}
}

\end{document}

%% file: tables/main_nuscenes.tex
\begin{table*}
\setlength{\tabcolsep}{5.95pt}
\begin{center}
\begin{tabular}{lll|ccc|ccc|ccc}
\toprule
&&& \multicolumn{3}{c|}{\textit{day-clear} -- nuScenes} & \multicolumn{3}{c|}{\textit{night} -- nuScenes} & \multicolumn{3}{c}{\textit{day-rain} -- nuScenes} \\
Method & sup. & tr.data & absRel & RMSE & $\delta_1$   & absRel & RMSE & $\delta_1$   & absRel & RMSE & $\delta_1$  \\
\midrule

Monodepth2~\cite{godard2019monodepth2} & M$^*$ & \textit{a}: \textit{dnr} & 0.1477 & 6.771 & 85.25 & 2.3332 & 32.940 & 10.54 & 0.4114 & 9.442 & 60.58 \\
Monodepth2~\cite{godard2019monodepth2} & M$^*$ & \textit{d} & 0.1374 & 6.692 & 85.00 & 0.2828 & \dotuline{9.729} & 51.83 & 0.1727 & 7.743 & 77.57 \\
PackNet-SfM~\cite{guizilini2020packnet} & Mv & \textit{d} & 0.1567 & 7.230 & 82.64 & 0.2617 & 11.063 & 56.64 & 0.1645 & 8.288 & 77.07\\
R4Dyn w/o r in~\cite{gasperini2021r4dyn} & Mv\textbf{r} & \textit{d} & \underline{0.1296} & 6.536 & \dotuline{85.76} & 0.2731 & 12.430 & 52.85 & \dotuline{0.1465} & 7.533 & \dotuline{80.59}\\
R4Dyn~\cite{gasperini2021r4dyn} (radar) & Mv\textbf{r} & \textit{d} & \textbf{0.1259} & \textbf{6.434} & \textbf{86.97}  & \dotuline{0.2194} & 10.542 & \dotuline{62.28} & \textbf{0.1337} & \textbf{7.131} & \textbf{83.91} \\
%TTT Monodepth2 & M$^*$ & \textit{d} & 0.1374 & 6.694 & 84.99 & 0.2871 & 10.506 & 61.01 & 0.2022 & 8.554 & 74.45 \\
%md2 mean \textit{d} pred. & M$^*$ & \textit{d} \\ % TODO
RNW~\cite{wang2021rnw} & M$^*$ & \textit{dn} & 0.2872 & 9.185 & 56.21 & 0.3333 & 10.098 & 43.72 & 0.2952 & 9.341 & 57.21 \\

[ours] baseline & Mv & \textit{d} & \dotuline{0.1333} & \dotuline{6.459} & \underline{85.88} & 0.2419 & 10.922 & 58.17 & 0.1572 & \dotuline{7.453} & 79.49 \\

%[ours] \pshortname-AD, \textit{n} & Mv & \textit{dT(n)} & 0.1433 & 6.954 & 83.27 & 0.2230 & 9.004 & \dotuline{68.61} & 0.1546 & 7.915 & 78.36 \\

[ours] \pname-AD & Mv & \textit{dT(nr)} & 0.1523 & 6.853 & 83.11 & \underline{0.2187} & \underline{9.003} & \underline{68.84} & 0.1601 & 7.832 & 78.97 \\

%[ours] \pshortname-DD, \textit{n} & Mv & \textit{dT(n)} & \dotuline{0.1302} & \textbf{6.373} & 85.02 & \underline{0.1958} & \textbf{8.471} & \underline{70.12} & 0.1429 & 7.313 & 79.59\\

%[ours] \pshortname-DD, \textit{r} & Mv & \textit{dT(r)} & 0.1323 & \dotuline{6.435} & 85.17 & 0.2508 & 11.868 & 56.96 & \underline{0.1364} & \textbf{7.099} & \underline{81.38} \\

[ours] \textbf{\pname-DD} & Mv & \textit{dT(nr)} & 0.1366 & \underline{6.452} & 84.61 & \textbf{0.1921} & \textbf{8.507} & \textbf{71.07} & \underline{0.1414} & \underline{7.228} & \underline{80.98} \\

\midrule
AdaBins~\cite{bhat2021adabins} & GT & \textit{a}: \textit{dnr} & 0.1384 & 5.582 & 81.31 & 0.2296 & 7.344 & 63.95 & 0.1726 & 6.267 & 76.01 \\
%AdaBins~\cite{bhat2021adabins} & GT & \textit{d} & \underline{0.1138} & \underline{4.805} & \dotuline{87.98} & 0.3336 & 14.002 & 45.77 & \underline{0.1540} & \dotuline{6.119} & \dotuline{81.20} \\
%[ours] \pshortname-AD, \textit{r} & GT & \textit{dT(r)} & \textbf{0.1052} & \textbf{4.621} & \textbf{89.58} & \dotuline{0.2644} & \dotuline{10.749} & \dotuline{55.51} & \textbf{0.1380} & \underline{6.030} & \textbf{83.32}\\
%[ours] \pshortname-AD \textit{a} & GT & \textit{dT(nr)} & 0.1052 & 4.590 & 88.93 & 0.2513 & 8.282 & 63.63 & 0.1453 & 5.938 & 82.70 \\

[ours] \textbf{\pname-AD} & GT & \textit{dnT(r)} & \textbf{0.1206} & \textbf{4.806} & \textbf{88.03} & \textbf{0.1821} & \textbf{6.372} & \textbf{75.33} & \textbf{0.1562} & \textbf{5.903} & \textbf{82.82} \\

\bottomrule
\end{tabular}
\end{center}
\vspace{-0.2cm}
\caption{Evaluation of self- and GT-supervised methods on the nuScenes~\cite{caesar2020nuscenes} validation set. Supervisions (sup.): M: via monocular videos, $^*$: test-time median-scaling via LiDAR, v: weak velocity, r: weak radar, GT: via LiDAR data. Training data (tr.data): \textit{d}: \textit{day-clear}, \textit{T}: translated in, \textit{n}: \textit{night} (incl.~\textit{night-rain}), \textit{r}: \textit{day-rain}, \textit{a}: \textit{all}. Visual support: \textbf{1$^\text{st}$}, \underline{2$^\text{nd}$}, \dotuline{3$^\text{rd}$} best. More conditions and metrics in the Appendix.}
\label{table:main_nuscenes}
\vspace{-0.2cm}
\end{table*}

%% file: tables/main_robotcar.tex
\begin{table*}
\begin{center}
\begin{tabular}{llll|cccc|cccc}
\toprule
&&&& \multicolumn{4}{c|}{\textit{day} -- RobotCar} & \multicolumn{4}{c}{\textit{night} -- RobotCar} \\
Method & source & sup. & tr.data & absRel & sqRel & RMSE & $\delta_1$   & absRel & sqRel & RMSE & $\delta_1$   \\
\midrule

Monodepth2~\cite{godard2019monodepth2} & [ours] & M$^*$ & \textit{d} & \underline{0.1196} & \underline{0.670} & \textbf{3.164} & \dotuline{86.38} & 0.3029 & 1.724 & \underline{5.038} & 45.88 \\

DeFeatNet~\cite{spencer2020defeatnet} & \cite{vankadari2022sundown} & M$^*$ & \textit{a}: \textit{dn} & 0.2470 & 2.980 & 7.884 & 65.00 & 0.3340 & 4.589 & 8.606 & 58.60 \\
ADIDS~\cite{liu2021allday} & \cite{vankadari2022sundown} & M$^*$ & \textit{a}: \textit{dn} & 0.2390 & 2.089 & 6.743 & 61.40 & 0.2870 & 2.569 & 7.985 & 49.00 \\
RNW~\cite{wang2021rnw} & \cite{vankadari2022sundown} & M$^*$ & \textit{a}: \textit{dn} & 0.2970 & 2.608 & 7.996 & 43.10 & \dotuline{0.1850} & \dotuline{1.710} & 6.549 & \dotuline{73.30} \\
WSGD~\cite{vankadari2022sundown} & \cite{vankadari2022sundown} & M$^*$ & \textit{a}: \textit{dn} & 0.1760 & 1.603 & 6.036 & 75.00 & \underline{0.1740} & \underline{1.637} & \dotuline{6.302} & \underline{75.40} \\

[ours] baseline & [ours] & Mv & \textit{d} & \dotuline{0.1209} & \dotuline{0.723} & \dotuline{3.335} & \underline{86.61} & 0.3909 & 3.547 & 8.227 & 22.51 \\

%[ours] \pname-AD & [ours] & Mv & \textit{dT(n)} & \textbf{0.1113} & \dotuline{0.707} & \dotuline{3.248} & \textbf{88.02} & \underline{0.1223} & \underline{0.851} & \underline{3.723} & \textbf{85.77} \\

[ours] \textbf{\pname-DD} & [ours] & Mv & \textit{dT(n)} & \textbf{0.1128} & \textbf{0.648} & \underline{3.206} & \textbf{87.13} & \textbf{0.1219} & \textbf{0.784} & \textbf{3.604} & \textbf{84.86} \\

%&&&& 0.1129 & 0.640 & 3.190 & 87.02 & 0.1256 & 0.824 & 3.703 & 83.87 \\

\bottomrule
\end{tabular}
\end{center}
\vspace{-0.2cm}
\caption{Evaluation of self-supervised works on the RobotCar~\cite{maddern2017oxford} test set. Trailing 0 added to the values from~\cite{vankadari2022sundown}. Notation from Table~\ref{table:main_nuscenes}.}
\label{table:main_robotcar}
\vspace{-0.2cm}
\end{table*}

%% file: tables/GANs.tex
\begin{table}[b]
\vspace{-0.7em}
\setlength{\tabcolsep}{4.5pt}
\begin{center}
%\begin{adjustbox}{max width=\linewidth}
\begin{tabular}{l|ccc}
\toprule

Method & \textit{avg/all} & \textit{day} & \textit{night} \\

\midrule

%Monodepth2~\cite{godard2019monodepth2} & 0.2122 & 0.1188 & 0.3037\\
%md2, \textit{d} & 0.1188 86.55 & 0.3037 45.78\\

%WSGD~\cite{vankadari2022sundown} & 0.1750 & 0.1760 & 0.1740 \\

% results with \delta_1
%[ours] w/ CycleGAN~\cite{zhu2017cyclegan} & 84.95 & 87.36 & 82.59 \\
%[ours] w/ ForkGAN~\cite{zheng2020forkgan} & 85.99 & 87.13 & 84.86 \\
%[ours] w/ degraded ForkGAN & \textbf{86.31} & \textbf{87.65} & \textbf{84.99} \\

[ours] w/ CycleGAN~\cite{zhu2017cyclegan} & 0.1244 & \underline{0.1159} & 0.1328 \\

[ours] w/ ForkGAN~\cite{zheng2020forkgan} & \textbf{0.1174} & \textbf{0.1128} & \textbf{0.1219} \\

[ours] w/ degraded ForkGAN & \underline{0.1213} & \underline{0.1159} & \underline{0.1266} \\

\bottomrule
\end{tabular}
%\end{adjustbox}
\end{center}
\vspace{-0.2em}
\caption{Robustness of \pname-DD against translations from different GANs. Evaluation of absRel on the RobotCar~\cite{maddern2017oxford} test set.}
\label{table:GANs}
%\vspace{-1.5em}
\end{table}

%% file: appendix/appendix_arxiv.tex
\appendix

\section{Supplementary Material}
The source code of our method, the main trained models reported in the experiments, and the generated translated images are publicly available at \href{https://md4all.github.io/}{https://md4all.github.io}.

This appendix includes additional details and results. Sections~\ref{sec:add_method} and~\ref{sec:add_setup} include additional information on the method and the experimental setup, while Sections~\ref{sec:add_quantitative} and~\ref{sec:add_qualitative} introduce more results, quantitatively and qualitatively, respectively.

In particular, this appendix is organized as follows:
\begin{itemize}
    \item Section~\ref{sec:add_method_supervised} includes details about our supervised framework.
    \item Section~\ref{sec:add_method_baseline} adds details about our self-supervised baseline.
    \item Section~\ref{sec:add_noise_norm} describes the noise and time-dependent normalization used for our self-supervised models.
    \item Section~\ref{sec:add_setup_translations} further reports details about the experimental setup for image translation.
    \item Section~\ref{sec:add_setup_nuscenes} includes details on the setup used for the nuScenes dataset.
    \item Section~\ref{sec:add_setup_robotcar} includes details on the setup used for the RobotCar dataset.
    \item Section~\ref{sec:add_setup_prior_works} adds details about the experimental setup for prior works.
    \item Section~\ref{sec:add_res_ablation_self} reports a detailed ablation study of our method on both nuScenes and RobotCar.
    \item Section~\ref{sec:add_res_dense} adds preliminary results with snow and fog on the DENSE dataset.    
    \item Section~\ref{sec:add_res_distributions} analyzes the effect of different data distributions among the conditions during training on RobotCar.
    \item Section~\ref{sec:add_res_supervised} compares different configurations of our supervised framework on nuScenes.
    \item Section~\ref{sec:add_res_nuscenes_nightrain} looks into quantitative results with rain at nighttime and averages across the various conditions on nuScenes.
    \item Section~\ref{sec:add_res_nuscenes_test} compares methods on the test set of nuScenes.
    \item Section~\ref{sec:add_res_distances} analyzes the performance of the methods at varying distances from the ego vehicle, both on nuScenes and RobotCar.
    \item Section~\ref{sec:add_res_qual_nuscenes_self} reports qualitative results of our self-supervised method on nuScenes.
    \item Section~\ref{sec:add_res_qual_nuscenes_fully} adds qualitative results of our fully-supervised method on nuScenes.
    \item Section~\ref{sec:add_res_qual_robotcar} discusses qualitative results of our self-supervised method on RobotCar.
    \item Section~\ref{sec:add_res_qual_failures} looks into failure cases of our self- and fully-supervised methods, exemplified on nuScenes.
    \item Section~\ref{sec:add_res_qual_translations} analyzes images generated via image translations for both nuScenes and RobotCar.
    \item Section~\ref{sec:add_not_work} lists attempted and alternative approaches that did not work.
\end{itemize}

\subsection{Additional Details on the Method}\label{sec:add_method}

\subsubsection{Supervised \pname}\label{sec:add_method_supervised}
In the main paper, we mostly focused on the more complex self-supervised setting (Sections~\reff{3.1.2} and~\reff{3.1.3}), and we extended our method to the supervised setup (Section~\reff{3.2}), making it the first depth estimation work to explore and address bad weather in supervised monocular settings.

\begin{figure}[t]
\begin{center}
    \includegraphics[width=1.00\linewidth]{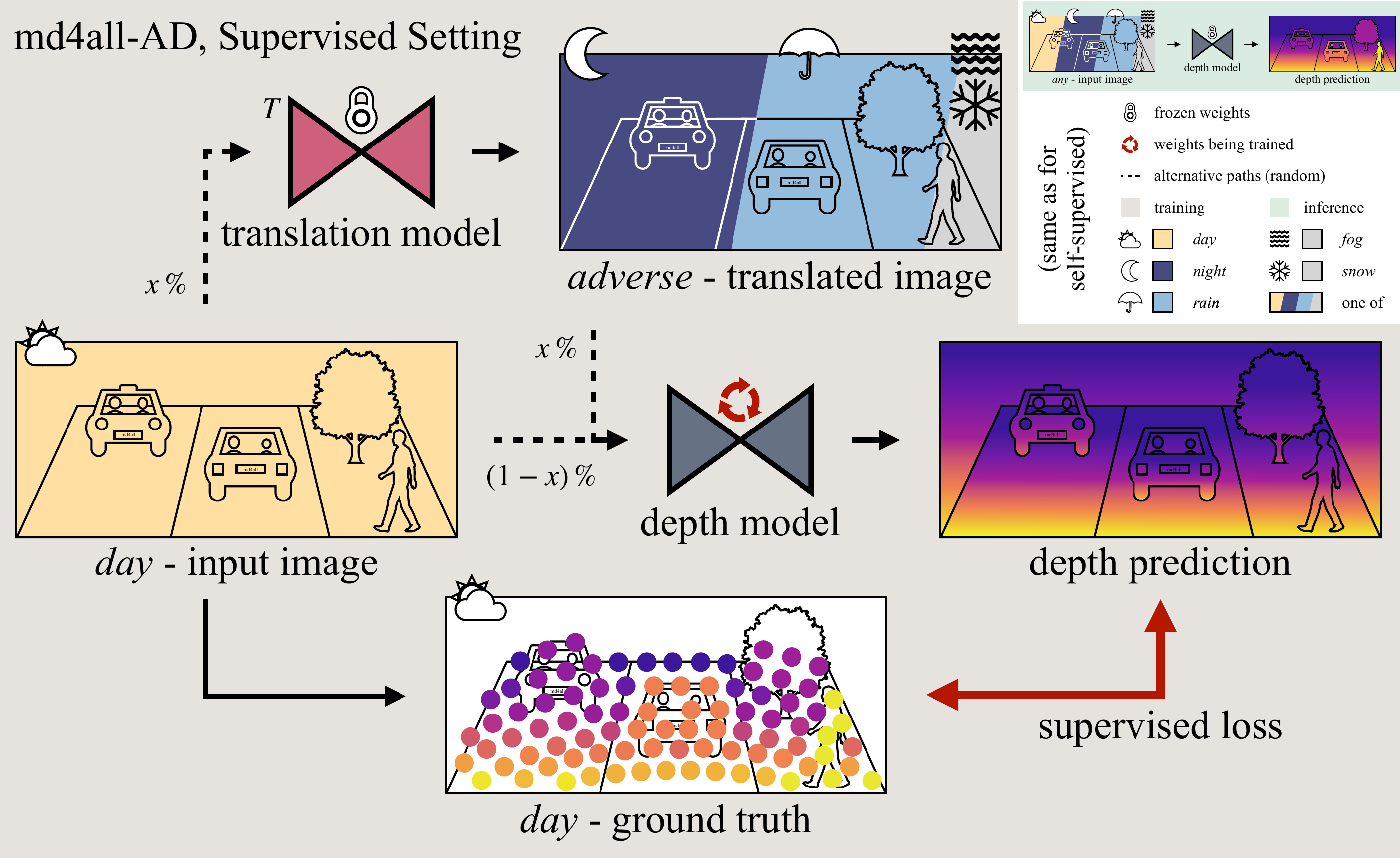}
\vspace{-0.7cm}
\end{center}
   \caption{Apart from the type of supervision, our \pname-AD supervised framework works similarly to the \pname-AD self-supervised framework described in Section~\reff{3.1.2} and Figure~\reff{4}, with Always Daytime, No Bad Weather. The depth model is trained with a mix of easy and translated samples, while its supervision is obtained from the ground truth data corresponding only to the easy samples. As in the self-supervised case, inference time is unchanged and performed with a single depth model without specialized branches (top right).}
\label{fig:framework_sup}
\vspace{-0.2cm}
\end{figure}
%\setfigurecounter{8}

Supervised models learn directly from the ground truth (e.g., LiDAR data). However, in adverse conditions (e.g., rain), the ground truth becomes unreliable (Figure~\reff{2}). Our supervised \pname\ aims to eliminate the sources of unreliability in the ground truth by providing a reliable signal in all conditions. As shown in Figure~\ref{fig:framework_sup}, we achieve this with the same principles described for the self-supervised settings: having a single depth model learn robust features agnostic of the condition in input by feeding a mix of easy and hard samples, with the ground truth always corresponding to the easy samples. Therefore, we use the same image translation model to generate adverse images corresponding to the easy ones in the training data. Then, we train the depth model with a mix of original easy and generated adverse inputs. Unlike the self-supervised settings where the training signal came from a pre-trained baseline model (\pname-DD) or the photometric losses (\pname-AD), the training signal is obtained directly from the ground truth data for supervised methods. When translating an image $e_i$ to adverse conditions $h_i^c$, we use as ground truth for $h_i^c$ the LiDAR data corresponding to $e_i$.

We associate this supervised method with our AD configuration since no distillation from a pre-trained model occurs (unlike for DD). Moreover, the depth model is trained in the same way as its baseline, i.e., via the ground truth, in an Always Daytime, no Bad Weather fashion, similarly to our self-supervised AD model (Section~\reff{3.1.2}).
Furthermore, the translated images should be used only for those conditions that render the ground truth unreliable. Therefore, for the experiments, we translated the inputs from \textit{day-clear} to \textit{day-rain}, since the ground truth is unreliable with rain (Figure~\reff{2}). Still, we used the original \textit{night} inputs since the ground truth is reliable at night (Figure~\reff{5}).

\subsubsection{Self-Supervised Baseline}\label{sec:add_method_baseline}
In this section, we further describe the loss functions used for the baseline of Section~\reff{3.1.1}. Such baseline is equivalent to Monodepth2~\cite{godard2019monodepth2} made scale-aware through weak velocity supervision from Guizilini et al.~\cite{guizilini2020packnet}.

The photometric loss is the combination of $\mathcal{L}_1$-loss and SSIM~\cite{wang2004image}, as done in~\cite{godard2017unsupervised_stereo}:
% Photometric error
\begin{equation}
\begin{gathered}
    \mathcal{L}_1 (I_t, \hat{I}_t) = \left\| I_t - \hat{I}_t\right\|_1 \\
    \mathcal{L}_{\textup{SSIM}} = 1 - \textup{SSIM}\left(I_t, \hat{I}_t\right)\\
    pe\left(I_t, \hat{I}_t\right) = \left(1-\alpha\right) \mathcal{L}_1 (I_t, \hat{I}_t) + \frac{\alpha}{2} \mathcal{L}_{\textup{SSIM}}(I_t, \hat{I}_t)
\label{eq:photometric_error}
\end{gathered}
\end{equation}
where $\alpha = 0.85$ is a weight to balance between the two terms. Furthermore, similarly to~\cite{godard2019monodepth2}, we account for partial occlusions by only considering the minimum reprojection error:
% min reproj. loss
\begin{equation}
    \mathcal{L}_{p}\left(I_t, \hat{I}_{s \to t}\right) = \min_{s} pe\left(I_t, \hat{I}_{s \to t}\right).
\label{eq:minimum_reprojection_loss}
\end{equation}

Moreover, following the so-called auto-mask from Monodepth2~\cite{godard2019monodepth2}, we automatically mask out the pixels that do not change appearance across different frames:
% automask
\begin{equation}
    M_{a} = \min_{s} \mathcal{L}_{p}\left(I_t, I_s\right) > \min_{s} \mathcal{L}_{p}\left(I_t, \hat{I}_{s \to t}\right).
\label{eq:automask}
\end{equation}

Therefore, the photometric loss is only computed in the areas where $M_a = 1$. Additionally, to encourage local smoothness and preserve sharp edges, we use the following term from~\cite{godard2017unsupervised_stereo}:
% smoothness loss
\begin{equation}
    \mathcal{L}_{s}\left(I_t, d_t^*\right) = \frac{1}{N} \sum_{p \in N} \sum_{i \in x,y} \left | \partial_i d_t^* (p) \right | e^{-\lvert \partial_i I_t \rvert}
\label{eq:smoothness}
\end{equation}
% velocity
where $\lvert \cdot \rvert$ is the absolute value computed element-wise, $\partial_x$ and $\partial_y$ are the gradients in x and y directions, and $d_t^* = d_t^*/\overline{d_t^*}$ is the inverse of the depth prediction normalized by the mean.

As already described in Section~\reff{3.1.1}, we follow~\cite{guizilini2020packnet} by using a weak velocity supervision $\mathcal{L}_{v}$ to achieve scale-awareness. This is defined as:
\begin{equation}
    \mathcal{L}_{v}\left({\hat{\mathbf{t}}_{t \to s}},\mathbf{t}_{t \to s}\right)
        = \Bigl|\lVert \hat{\mathbf{t}}_{t \to s} \rVert_2 - \lVert \mathbf{t}_{t \to s} \rVert_2\Bigr|
\label{eq:velsup}
\end{equation}
where $\hat{\mathbf{t}}_{t \to s}$ and $\mathbf{t}_{t \to s}$ are the predicted and ground truth pose translations, respectively, which can be easily obtained from the available odometry information, through the ego vehicle speed and the time interval across frames.

\subsubsection{Noise and Normalization}\label{sec:add_noise_norm}
We did not apply either of these two techniques in the supervised setting, i.e., where we applied our method on AdaBins~\cite{bhat2021adabins}, since the LiDAR ground truth provides a strong signal which already enables such robust feature extraction.

For the self-supervised models, we normalize the inputs at training time depending on the time of the day (i.e., day and night). Towards this end, we precompute the mean and variance of the pixel values across the two conditions throughout the dataset and normalize the inputs accordingly. In Tables~\ref{table:ablations_main_nuscenes} and~\ref{table:ablation_robotcar}, we show how this time-dependent normalization has a positive impact at training time, as it aligns the features in a condition-agnostic manner. Additionally, we show that this normalization can be avoided at inference time for similar results. Avoiding it ensures that the operations executed across all conditions are identical at inference time. Thus, after deployment, our method does not require any knowledge about the current weather and illumination settings, which may be hard to define and may intersect with other conditions (e.g., wet ground without rain or dusk). Nevertheless, due to the relatively small difference during inference, we used time-dependent normalization for self-supervised models unless otherwise noted.

Furthermore, in the case of camera sensors delivering significant noise levels (e.g., nuScenes~\cite{caesar2020nuscenes}), we augment the inputs of self-supervised models with heavy noise. The noise is randomly applied to 50\% of the inputs, regardless of their condition. This helps to learn more robust features. When the noise is used, we compute the losses on the samples without noise.
Specifically, we generated the noise by adding to the image a random pattern following the uniform distribution [0.005, 0.05], then clamped the pixel values to [0, 1], thereby ensuring that the input remains within a valid range.

\input{tables/ablations_main_nuscenes.tex}
%\settablecounter{5}

\subsection{Additional Details on the Experimental Setup}\label{sec:add_setup}
\subsubsection{Day-to-adverse Translation}\label{sec:add_setup_translations}
For the experiments, we focused on two adverse conditions in nuScenes~\cite{caesar2020nuscenes} (i.e., \textit{rain} and \textit{night}) and one in RobotCar~\cite{maddern2017oxford} (i.e., \textit{night}), alongside the standard conditions \textit{day-clear} / \textit{day}. Towards this end, we trained two different ForkGAN~\cite{zheng2020forkgan} models for nuScenes, one for each condition, and one for RobotCar, to enable translations from \textit{day-clear} to each challenging condition, tailored to each dataset.
For the RobotCar dataset, we trained the GAN using the 34128 daytime samples from the scene 2014-12-09-13-21-02 and the 32585 nighttime samples from 2014-12-16-18-44-24. The dataset offered enough samples to train the image translation model thanks to the high frame rate.
Instead, the nuScenes dataset only provides 6951 samples for \textit{day-rain} and 4706 for \textit{night}, which are insufficient for the GAN to learn such day-to-adverse translation.
Therefore, to learn the transition from \textit{day-clear} to \textit{day-rain}, we additionally used all \textit{day-rain} samples from the nuImages dataset~\cite{caesar2020nuscenes} resulting in a total number of 19857 \textit{day-rain} frames. We balanced this with the 19685 \textit{day-clear} images of the nuScenes training set. Since the nuImages dataset does not provide any metadata about the weather condition, we manually labeled all its samples with their respective weather condition.
Nevertheless, \textit{night} samples are insufficient in nuScenes and nuImages (14302) to train a GAN. For this reason, we first trained the \textit{day-clear} to \textit{night} translation model on BDD100K~\cite{yu2020bdd100k}, which includes 36728 \textit{day} and 27971 \textit{night} images in its training set. Then, we fine-tuned it on the available nuScenes \textit{night} samples from the training set.

\subsubsection{nuScenes}\label{sec:add_setup_nuscenes}
For the depth experiments on nuScenes~\cite{caesar2020nuscenes}, we followed the setup of R4Dyn~\cite{gasperini2021r4dyn}, using the official data splits and evaluating up to 80 meters comparing the predictions with a single LiDAR scan. As in R4Dyn, we discarded static frames (i.e., where the ego vehicle is stationary) for self-supervised models. While a single scan is highly sparse compared to the dense depth prediction, it limits the artifacts introduced by accumulating multiple scans over time for denser ground truth (e.g., due to moving objects and changing perspectives). We augmented the inputs with heavy noise for self-supervised models to mimic that in the night samples. For the supervised setting, learning from such a sparse signal means reducing the workload needed for producing the ground truth, albeit rendering it more challenging. As it is standard for supervised setups, the models do not learn the depth of the unreachable areas for the ground truth sensor (e.g., the sky for LiDAR). All qualitative images and quantitative results reported in the main paper and this supplementary material are from the validation set unless otherwise noted (e.g., test set in Table~\ref{table:test_nuscenes}).

\subsubsection{RobotCar}\label{sec:add_setup_robotcar}
For the experiments on RobotCar~\cite{maddern2017oxford}, we followed the setup of WSGD~\cite{vankadari2022sundown} using the six sequences in the 2014-12-09-13-21-02 traversal as daytime samples, and the six sequences in the 2014-12-16-18-44-24 traversal as nighttime ones. Since the peculiarity of RobotCar is that it was recorded by driving over the same route multiple times over a year, a training-test split with non-overlapping drives is required to properly assess the models' generalization capabilities. Therefore, we used the split provided by WSGD. 

As in~\cite{wang2021rnw}, we used the left images of the front stereo-camera (Bumblebee XB3), of which we removed the bottom 20\% (i.e., ego vehicle bonnet), and the ground truth data from the LMS front LiDAR sensor. We used the official toolbox to accumulate multiple LiDAR scans and project them to the input images. Towards this end, we used visual odometry, as recommended by the official documentation of the dataset, and a time margin of $\pm 4e6$ from the origin timestamp, as in~\cite{wang2021rnw}. As commonly done for self-supervised methods, we discard static frames thresholding the translation provided by the visual odometry. We did not apply heavy noise for RobotCar as the night samples did not exhibit it. Furthermore, since the RobotCar camera occasionally suffers from inconsistent illumination across neighboring frames, we discarded these too. Specifically, we removed all triplets where the keyframe's mean RGB value is $\ge0.9$, or the RGB mean value difference between two consecutive frames is $>0.05$. In addition, only the images with a corresponding LiDAR ground truth could be evaluated.

For the experiment with degraded translations via random erasing (Table~\reff{3}), we applied it randomly to 10\% of the inputs, with a patch sized randomly between 5\% and 10\% of the input dimensions, placed randomly within the image, with an aspect ratio between 0.3 and 3.3. When applying random erasing, the performance of \pname\ marginally improved in terms of $\delta_1$ by 0.32\% on \textit{all} (absRel slightly decreased as shown in Table~\reff{3}), thanks to the augmentation and regularization effect introduced by the patches. Throughout the main paper and this supplementary material, all qualitative images and quantitative results are from the test set defined by WSGD.

\subsubsection{Prior Works}\label{sec:add_setup_prior_works}
For prior works on RobotCar, we reported the values computed by Vankadari et al.~\cite{vankadari2022sundown}, who retrained RNW~\cite{wang2021rnw} on a non-overlapping split (which inherently reduced the scores), and also re-evaluated DeFeatNet~\cite{spencer2020defeatnet} and ADIDS~\cite{liu2021allday} on the same test split (again reducing the scores).
Among works focusing on depth estimation in the dark, only RNW reported its results on the more challenging nuScenes dataset. However, since the authors reported their scores on a different, custom split, we retrained their model on the official split.
For this reason, the results of RNW differ from those reported directly by Wang et al.~in their paper~\cite{wang2021rnw}. Nevertheless, the difference is relatively small as RNW reported at \textit{night}: absRel of 0.3150 (0.3333 from our experiment with RNW), sqRel of 3.793 (4.006), RMSE of 9.6408 (10.098), and $\delta_1$ of 50.81 (43.72). While this performance gap should be attributed to the different data splits used, it does not affect the comparisons since our models performed significantly better than what Wang et al.~reported in their paper~\cite{wang2021rnw}, both quantitatively and qualitatively.
Furthermore, on nuScenes, we also report the values of R4Dyn~\cite{gasperini2021r4dyn} and PackNet-SfM~\cite{guizilini2020packnet}, as provided to us by the authors of~\cite{gasperini2021r4dyn}.
Additional related works tackling adverse conditions exist (Section~\reff{2.2.1}). Still, their lack of open-source code or their use of unconventional and unclear experimental setups prevented us from directly comparing with their methods.

\input{tables/ablations_main_robotcar.tex}
%\settablecounter{6}

\subsection{Additional Quantitative Results}\label{sec:add_quantitative}

\subsubsection{Ablation Study}\label{sec:add_res_ablation_self}
In Table~\ref{table:ablations_main_nuscenes}, we report an ablation study over the main components of our method.

We started from a Monodepth2~\cite{godard2019monodepth2} trained on the entire training set of nuScenes~\cite{caesar2020nuscenes} (A0), meaning all available conditions. A0 performed poorly under adverse conditions due to the difficulty of establishing pixel correspondences across consecutive \textit{night} and \textit{rain} frames. A0 delivered scores and errors similar to those reported for Monodepth2 by prior works in their papers, such as RNW~\cite{wang2021rnw} and WSGD~\cite{vankadari2022sundown}. Furthermore, the outputs of A0 exhibit the same issues shown by RNW and WSGD in their qualitative comparisons (e.g., holes in the ground), which are not present and much improved when training Monodepth2 only on \textit{day-clear}, as reported throughout this work (A3). In particular, with A0-A3, we show how the standard Monodepth2 performs substantially better than the complex RNW overall and significantly better than WSGD in the daytime (Table~\ref{table:ablation_robotcar}).

A1 is a Monodepth2 model trained on \textit{day-clear} and \textit{night} (i.e., everything excluding \textit{day-rain}). A1 performed similarly to A0 at \textit{night}, but significantly better with \textit{rain}. Additionally, it can be seen how training on \textit{day-rain} samples negatively affected the \textit{day-clear} performance (A0) while excluding such rainy samples improved in standard conditions (A1).
Then, A2 is another Monodepth2 model, trained on \textit{day-clear} and the translated \textit{night} samples we generated with the GAN. A2 was fed a mix of \textit{day-clear} and generated \textit{night} ones with $x=15\%$, to resemble the day-night distribution of the training set (used by A1). The comparison of A1 with A2 shows mainly two aspects about the translated images (Figure~\ref{fig:translations_nuscenes}): they are not entirely realistic, and, unlike the real ones, they do not prevent establishing the pixel correspondences. If the generated samples were completely realistic (i.e., like the real \textit{night} ones from nuScenes), there would have been a much smaller difference between A1 and A2. In particular, the generated images do not fully resemble the real \textit{night} ones (Figure~\ref{fig:translations_nuscenes}), especially for the noise, which is more consistent throughout the generated frames compared to the real ones, and the darkness levels, with images that are not as black as the real \textit{night} ones of nuScenes. This lack of realism in the generated images is the reason for the performance improvement of A2 at \textit{night} compared to A1.

Similarly, WSGD~\cite{vankadari2022sundown} showed the importance of denoising \textit{night} images, with noise detrimental to the models. Since the translated images do not exhibit the same kind of noise and reflections as the real ones and are particularly unrealistic when translating a sunny sample (Figure~\ref{fig:translations_nuscenes}), A2 was able to establish pixel correspondences across the translated samples to a certain extent. Additionally, as $x$ randomizes the condition of each input independently, with A2, the translations also introduce a regularization effect as data augmentation. 
A3 was also a Monodepth2 model but trained only on the \textit{day-clear} samples. As shown in Table~\reff{1}, this improves significantly compared to training on all conditions (A0) due to the impossibility of establishing correspondences at \textit{night} for A0.

The weak velocity supervision~\cite{guizilini2020packnet} (A4) improved significantly over A3, thanks to better pose estimates. Compared to A0-A3, which need ground truth median scaling at test time, A4 is scale-aware and does not use it. With A5, we added heavy noise (consistently throughout the triplets) but computed the losses on the clean samples (i.e., without noise). This made it worse for \textit{day-clear} and \textit{day-rain}, but improved for \textit{night} compared to A4. The motivation for A5 develops from the intense noise present in the \textit{night} samples of nuScenes (Figure~\reff{2}), which may confuse the models. We did not apply this under supervised settings (i.e., our method on AdaBins). The improvement seen with adding noise while computing the loss on the inputs without noise paved the way for the concept of our AD model.
With A6, we added the translated images generated with the GAN from day to night ($x=50\%$) to the training data. This made it worse than A0 for \textit{day-clear}, but similarly to A2, it improved for \textit{night} due to the lack of realism of the generated samples, which allowed to establish pixel correspondences.
For A6 (and A2), perfectly realistic generated samples would have been detrimental to learning.

With A7, we did not feed the translated images to the pose model but only to the depth one. This guarantees reasonable pose estimates, which improve the task at hand under all three conditions. Then, with AD\textit{n}, we computed the losses only on the \textit{day-clear} $e_i$ samples, corresponding to the translated ones given as input. This significantly improved the model performance on \textit{day-clear}, reaching a level similar to A3 (i.e., only a marginal degradation on the standard conditions). It should be noted that if the translated images perfectly mimicked the real \textit{night} ones, A2, A6, and A7 would have performed relatively poorly, i.e., similarly to A0 and A1 at \textit{night}. In the case of perfect day-to-night translations, always computing the loss only on the \textit{day-clear} $e_i$ samples (as in AD\textit{n}, instead of calculating it on the translated ones, as in A2, A6, and A7) would have had a significantly positive impact at \textit{night}.

\input{tables/dense.tex}
%\settablecounter{7}

With A9, we removed the time-dependent normalization from AD\textit{n}, which was used from A6 to AD\textit{n}. This shows that this technique benefits both \textit{day-clear} and \textit{day-rain}, as it helps construct a unified representation for all conditions. With AD\textit{a}, we incorporated day-to-rain translations to AD\textit{n} alongside the day-to-night images ($x=66\%$, i.e., one-third for each condition). This delivered a similar performance to AD\textit{n} (e.g., improved for \textit{night}, and improved the RMSE, with a worse absRel for \textit{day-clear}). As for Table~\reff{1}, the LiDAR ground truth is not fully reliable for \textit{day-rain} and also significantly sparser than for \textit{day-clear} (Figure~\reff{2}). With A11, we added the day distillation loss (Equation~\reff{1}) on the translated inputs while keeping the standard losses for the \textit{day-clear} inputs. This combination improved across the board. Then, with DD, we simplified the training process by using only the day distillation loss for all inputs (including \textit{day-clear}). Thus, DD\textit{n} does this for \textit{day-clear} and \textit{night}, DD\textit{r} does it for \textit{day-clear} and \textit{day-rain}. Our day distillation provides a dense and reliable signal (from A4 inferring only on \textit{day-clear} samples), improving the errors and metrics across the board.

Finally, with A15, we show the impact of avoiding the time-dependent normalization at test time. Compared to DD\textit{a}, A15 does not apply such time-dependent normalization at inference time but only at training time. A15 obtains comparable results throughout the various settings. Instead, as shown with A9, the time-dependent normalization is helpful at training time. After training, our model has learned robust features agnostic to the condition, allowing it to perform similarly regardless of the image normalization applied at test time.
This demonstrates how our method does not need any condition-specific setups at inference time to deliver robust predictions, clearly separating our \pname\ from previous works requiring custom branches for each condition. In the supervised settings (i.e., our method applied on AdaBins), we did not perform any time-dependent normalization since the strong LiDAR supervision is enough to learn depth estimation at \textit{night}.

In Table~\ref{table:ablation_robotcar}, we report various configurations of our method on the RobotCar~\cite{maddern2017oxford} dataset. As for nuScenes (A15 in Table~\ref{table:ablations_main_nuscenes}), we show that not applying the time-dependent normalization at test-time (i.e., executing the same operations with the same setup across the different conditions) does not negatively affect the predictions, achieving comparable results. Furthermore, we show how the results change when applying the median scaling via LiDAR data at test time. This technique is used by Monodepth2~\cite{godard2019monodepth2}, WSGD~\cite{vankadari2022sundown}, and most other methods compared in this work. Our model does not need such scaling via ground truth data, thanks to its scale awareness learned via the weak velocity supervision introduced by PackNet-SfM~\cite{guizilini2020packnet}.

Nevertheless, accurate scaling can further improve the results, especially at \textit{night}. Compared to nuScenes, the RobotCar dataset provides less precise odometry information, causing difficulties for the baseline and our models to learn the correct scaling. This can be seen by the improved scores at \textit{night} when applying the median scaling via LiDAR data. With reliable scaling via the ground truth data, \pname-DD outperforms \pname-AD.

\input{tables/distributions.tex}
%\settablecounter{8}

\input{tables/fully-sup_nuscenes}
%\settablecounter{9}

\input{tables/ext_main_nuscenes.tex}
%\settablecounter{10}

\input{tables/40m_nuscenes.tex}
%\settablecounter{11}

\input{tables/ext_40m_nuscenes.tex}
%\settablecounter{12}

\input{tables/60m_nuscenes.tex}
%\settablecounter{13}

\input{tables/ext_60m_nuscenes.tex}
%\settablecounter{14}

\input{tables/test_nuscenes.tex}
%\settablecounter{15}

\textbf{AD and DD}
While the benefit of our DD configuration over AD is evident for nuScenes, the gap is not as significant for RobotCar, with the two delivering comparable results (Table~\ref{table:ablation_robotcar}). This difference can be attributed to various reasons. First of all, nuScenes is more challenging, as demonstrated by the lower scores obtained by the models across all conditions, especially at \textit{night}. Thus, the improvements of DD over AD might be reduced for RobotCar since AD already achieves solid results. The higher amount of images available on RobotCar to learn the translation task led to more realistic translations than nuScenes (Section~\ref{sec:add_res_qual_translations}). Then, the less precise odometry information of RobotCar impacted the performance of the baseline through weak velocity supervision. Therefore, the baseline possibly learned wrong poses. This is not the case on nuScenes (Table~\ref{table:ablations_main_nuscenes}), where the baseline (A4) improved significantly over Monodepth2 (A3). This did not happen for RobotCar. Since our \pname-DD learns to mimic the baseline via knowledge distillation, our model is directly affected by the weaker baseline in RobotCar, delivering similar results to AD. Instead, in nuScenes the gap between AD and DD is substantial throughout the conditions.

As shown with Monodepth2~\cite{godard2019monodepth2} and AdaBins~\cite{bhat2021adabins}, our method is widely flexible and applicable to different architectures and types of supervision. While being out of the scope of this work, our approach can be seamlessly applied to other self-supervised or supervised frameworks, such as PackNet-SfM~\cite{guizilini2020packnet}, since we do not alter the model architecture, but only its training scheme. In particular, to apply the proposed \pname\ to an existing depth estimation method, no structural changes are needed, as it is sufficient to feed to the model the translated images $x\%$ of the time during training.

\subsubsection{DENSE Dataset: \textit{Snow} and \textit{Fog}}\label{sec:add_res_dense}
Disclaimer: First, please consider that these are only preliminary experiments and that we have not yet explored these conditions and models to the same extent as \textit{night} and \textit{rain} in the rest of this work. Nevertheless, we report them here as they provide interesting insights.

In Table~\ref{table:dense}, we show a first attempt to tackle the problem of monocular depth estimation in the presence of snow or fog with the DENSE dataset~\cite{bijelic2020dense}. While our \pname\ performed better than the standard Monodepth2~\cite{godard2019monodepth2} across the board, the improvement is relatively small compared to the other datasets and conditions explored (e.g., Tables~\reff{1} and~\reff{2}). There are multiple reasons for this, explained below.

An impactful aspect to be considered is related to the available data. The condition boundaries are somewhat blurry. Overcast \textit{day-clear} samples can be similar to light \textit{fog} or light \textit{snow}. This is problematic for the GAN used for image translation, which cannot distinguish the distributions and adequately learn the translation task.

Furthermore, the term snow is generic and includes various scenarios, such as light snow, heavy snow, blizzard, partly covered ground, fully covered ground, piles of snow, or wet ground with light snow falling. These settings differ substantially, but all belong to the same \textit{snow} condition. This high variability is problematic for the translation task. While this issue can occur similarly with \textit{night} and \textit{rain} too, it is not as severe, and the diversity is more limited.

Another significant issue is the amount of usable image data for these conditions, which is insufficient to properly learn the translation task with ForkGAN~\cite{zheng2020forkgan}. As we did for \textit{night} for nuScenes (Section~\reff{4.1}), also for DENSE, we had to supplement with extra snow images taken from another dataset: Boreas~\cite{burnett_ijrr23_boreas}. 
We trained the snow ForkGAN with 17591 \textit{day-clear} and 8443 \textit{snow} samples from DENSE, plus 25036 \textit{day-clear} and 26437 \textit{snow} samples from Boreas for the pre-training.
While supplementing with data from Boreas helped, the number of images from DENSE was relatively low compared to nuScenes and RobotCar, preventing effective translations.

Training data for the depth models was 7947 \textit{day-clear} keyframes for DENSE. These keyframes were relatively few (15129 were used for nuScenes and 17790
for RobotCar), and they were extracted from short sequences, so they did not exhibit high variability. This reduced the depth estimation performance of the models. The validation set was also small with only 289 for \textit{day-clear}, 1281 for \textit{snow}, and 543 for \textit{fog}. DENSE contains more images, but those were not usable due to various reasons, e.g., they were captured by different sensors.

Furthermore, as with \textit{rain} (Figure~\reff{2}), the LiDAR is not reliable in the presence of \textit{snow} or \textit{fog}, as it often captures snowflakes, fog particles, or is even obstructed by the snow accumulated on the sensor itself. We mitigated this problem by filtering the erroneous LiDAR points via clustering, but we could not eliminate all problematic measurements. As seen for \textit{rain} on nuScenes, in adverse conditions the LiDAR sensor is unable to collect measurements at further distances (e.g., Figure~\reff{5} \textit{rain} vs.~\textit{night} ground truth depth). Therefore, we could only evaluate a limited set of points at a closer distance. We used a single LiDAR scan as ground truth.
Additionally, among the snow data, many samples were recorded in remote areas with relatively flat surroundings. Considering the limited distance and the flat surroundings, a model overfitting on flat ground may seem erroneously adequate by obtaining good quantitative results.

Additionally, for the weak velocity supervision of our baseline, we exploited the information from the CAN bus, as provided by the authors of DENSE. We used the vehicle speed and the frame rate to compute the camera translation between the frames. However, since the vehicle speed is provided as single value for each short sequence, the camera poses could only be coarsely approximated. This likely affected the performance of the baseline, hence that of our \pname-DD too. Furthermore, we used the CAN data to discard static inputs (i.e., stationary ego vehicle) and those where the ego vehicle is turning. We filtered the latter when the steering wheel angle exceeded 20°. This filtering led to the numbers indicated above.

All these points should be considered when evaluating these preliminary results on DENSE.

First of all, regarding Table~\ref{table:dense}, it can be seen how the \textit{day-clear} results are not as good as those seen for nuScenes (Table~\reff{1}) or RobotCar (Table~\reff{2}). This could be attributed to DENSE containing more challenging data. More likely, it is due to the inability of the models to properly generalize on DENSE due to the relatively low diversity in the training data and the limited amount of training samples. Therefore, our \pname-DD learned from a weak baseline which could not correctly estimate depth in standard conditions.
Nevertheless, our \pname-DD outperformed Monodepth2 in standard settings, thanks to the regularization effect of our translations.

In the table, we report light-fog for \textit{fog} and full-coverage or currently snowing for \textit{snow}. We opted for light-fog since dense-fog exhibited too few LiDAR points for the evaluation, all at relatively close distances (easier). Instead, light-fog allowed for a more thorough assessment at further distances. For reference, all results were better with dense-fog than light-fog. For \textit{snow}, we selected those with full-coverage or weather metadata snow. This is because, among the annotated conditions, they had the most precise boundaries with other conditions.

Moreover, both models perform similarly with \textit{fog} as in ideal settings (i.e., \textit{day-clear}). While this hints that fog is not as challenging as rain or night (Tables~\reff{1} and~\reff{2}), the values are also affected by the limited distance of the ground truth used for the evaluation. Therefore, the performance may degrade significantly at further distances due to the fog preventing seeing the details, but that cannot be evaluated. Nevertheless, already at the available ground truth distances, our model outperformed Monodepth2, on which ours is based.

Despite the limited distance of the ground truth, \textit{snow} appears more challenging than \textit{fog}, causing a significant drop in performance compared to the ideal settings (i.e., \textit{day-clear}). With snow, the limited data available to learn proper translations substantially impacted our method's performance, which obtained only slightly better scores than Monodepth2.

Stronger condition boundaries (e.g., more precise annotations) and more training data would significantly improve the translations and our method's outcomes. Furthermore, depth ground truth reaching further distances without any artifacts would allow us to assess the actual performance of the models. While these factors would contribute to a more considerable gap between the proposed \pname\ and Monodetph2, the issues with the translations also highlight the limitations of our approach: the difficulty in collecting adverse data that would lead to solid results (e.g., Table~\reff{2} with RobotCar~\cite{maddern2017oxford}).

Due to the substantial limitations encountered with this data, the DENSE dataset is unsuitable for depth estimation. However, we used it to provide these preliminary results with snow and fog. New real data with artifact-free long-distance ground truth is needed to properly explore monocular depth estimation in these conditions.

\subsubsection{Different Distributions of Conditions}\label{sec:add_res_distributions}
In Table~\ref{table:distributions}, we explore the effect of different data distributions among the conditions during training. We vary this via the parameter $x$. In the rest of this work, $x$ was selected to equally distribute the inputs among the conditions. So for RobotCar 50\% for half for \textit{day} and half for \textit{night} (i.e., 50\textit{d} - 50\textit{n} in the table); for nuScenes 66\% corresponding to one third for each of \textit{day-clear}, \textit{night}, and \textit{rain}; one third each also for the DENSE dataset.

While intuitively increasing the amount of \textit{day} images could improve the performance on \textit{day}, this is not the case by randomizing via $x$ at each training sample independently. This is because with enough epochs, our model sees all images in all conditions, so the training data remains unchanged, causing only minor differences as the model might be fed more or fewer translations (Table~\ref{table:distributions}).
For \textit{day}, beyond the observed regularization effect (e.g., Table~\reff{2}), there is little room for gains as long as the baseline model $B$ to distill from remains the same.
Instead, if $x$ affected which portion of the training data is translated, it would have a more significant impact than shown in Table~\ref{table:distributions}.
In that case, seeing too many or too few translated images may impair the performance as the model does not experience enough of the ideal settings or not enough adverse conditions to tackle them properly.

\subsubsection{Supervised Configuration Comparisons}\label{sec:add_res_supervised}
Table~\ref{table:fully-sup_nuscenes} reports a comparison of different supervised configurations of AdaBins~\cite{bhat2021adabins} and our \pname-AD applied on AdaBins. Specifically, AdaBins trained only on \textit{day-clear} resulted in a significant improvement on \textit{day-clear} and \textit{day-rain} compared to the AdaBins trained in all conditions (i.e., \textit{a}). Analogously, our model trained on \textit{day-clear} and translated \textit{day-rain} samples performed better than ours trained on all but substantially worse at \textit{night}. As seen in the self-supervised case, our model trained in all conditions outperformed the baseline across the board (i.e., AdaBins trained on all), thereby not introducing any trade-off while improving in adverse conditions over the model it is based on.

\subsubsection{nuScenes \textit{Night-rain} and Average}\label{sec:add_res_nuscenes_nightrain}
In Table~\ref{table:ext_main_nuscenes}, we report results on more conditions of nuScenes~\cite{caesar2020nuscenes}, such as the most difficult \textit{night-rain} and an average over \textit{all}, alongside the sqRel errors not fitting in Table~\reff{1} (due to the limited space available). \textit{All} is not computed as an average on the various conditions but rather as an average of the performance on each sample (i.e., \textit{night} counts marginally, accounting for only 10\% of the images). Our model outperforms the baseline AdaBins across the board for the supervised case. Similarly, in the self-supervised setting, our \pname\ improved significantly over the baseline and Monodepth2~\cite{godard2019monodepth2}, second only in ideal conditions (\textit{day-clear}) to the radar-based R4Dyn~\cite{gasperini2021r4dyn}.

\input{tables/40m_robotcar.tex}
%\settablecounter{16}
\input{tables/60m_robotcar.tex}
%\settablecounter{17}
\input{tables/80m_robotcar.tex}
%\settablecounter{18}

\begin{figure*}[t]
\begin{center}
\includegraphics[width=1.00\textwidth]{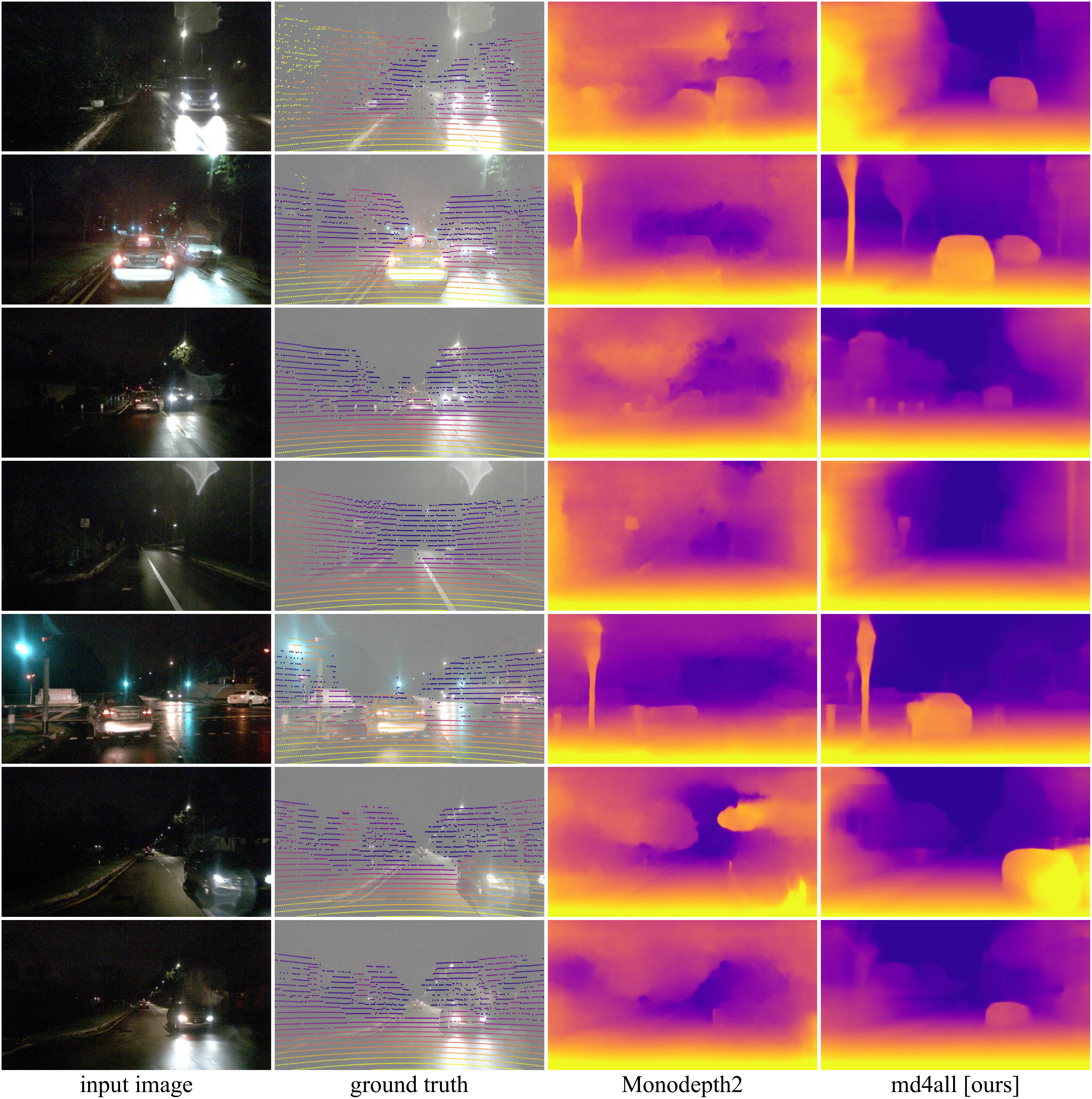}
\vspace{-0.7cm}
\end{center}
   \caption{Comparison of self-supervised models on nuScenes~\cite{caesar2020nuscenes} \textit{night} samples. The standard Monodepth2~\cite{godard2019monodepth2} is compared to our \pname-DD applied to Monodepth2. This set of samples is particularly challenging for the standard Monodepth2 due to the overall darkness and reflections.}
\label{fig:nuscenes_dark}
\vspace{-0.2cm}
\end{figure*}
%\setfigurecounter{9}

\begin{figure*}[t]
\begin{center}
\includegraphics[width=1.00\textwidth]{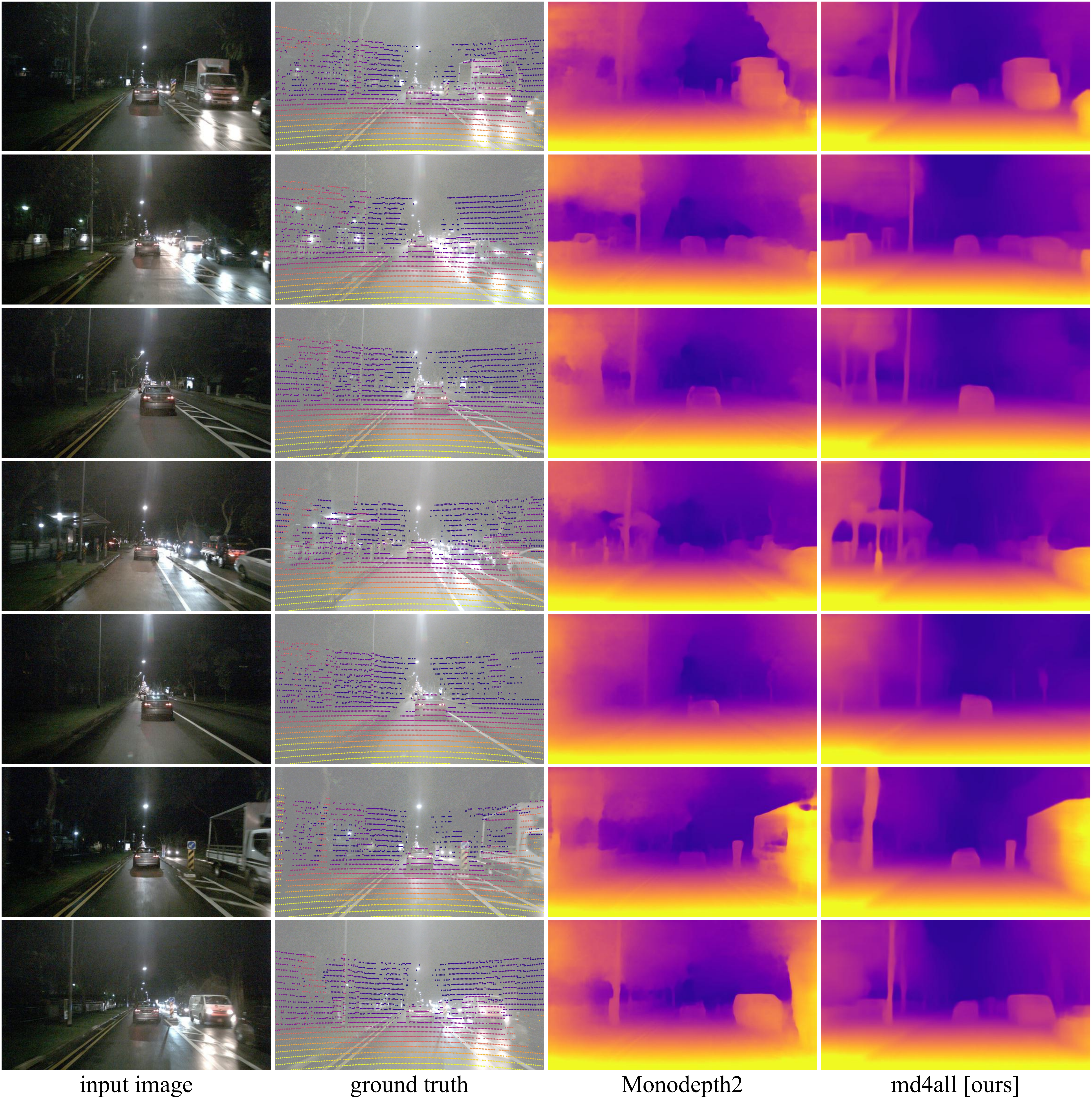}
\vspace{-0.7cm}
\end{center}
   \caption{Comparison of self-supervised models on relatively bright nuScenes~\cite{caesar2020nuscenes} \textit{night} samples. The standard Monodepth2~\cite{godard2019monodepth2} is compared to our \pname-DD applied to Monodepth2. This set of samples could be handled reasonably by the standard Monodepth2, thanks to the high brightness of the scenes.}
\label{fig:nuscenes_bright}
\vspace{-0.2cm}
\end{figure*}
%\setfigurecounter{10}

\begin{figure*}[t]
\begin{center}
\includegraphics[width=1.00\textwidth]{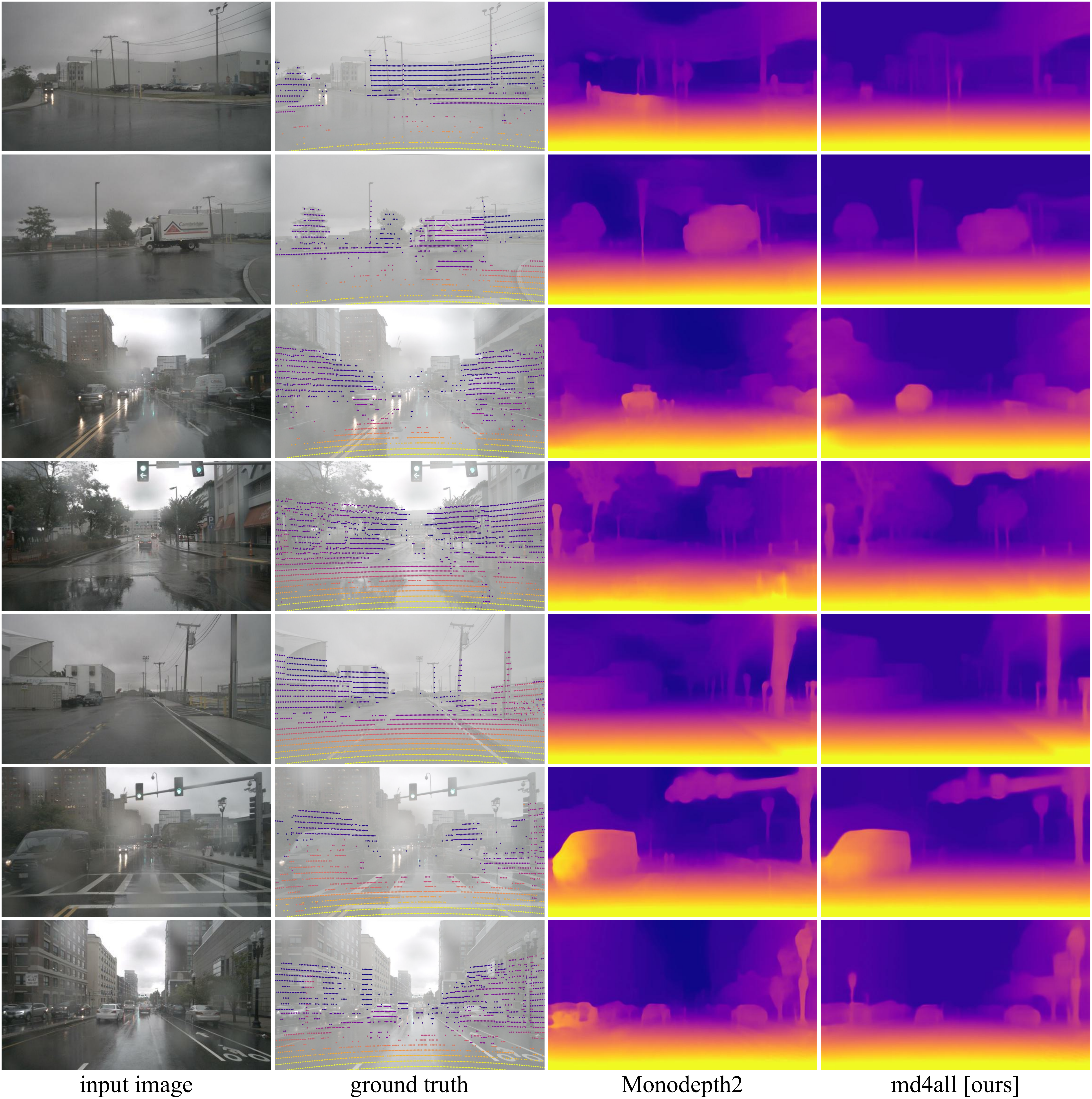}
\vspace{-0.7cm}
\end{center}
   \caption{Comparison of self-supervised models on nuScenes~\cite{caesar2020nuscenes} \textit{rain} samples. The standard Monodepth2~\cite{godard2019monodepth2} is compared to our \pname-DD applied to Monodepth2.}
\label{fig:nuscenes_rain_self_suppl}
\vspace{-0.2cm}
\end{figure*}
%\setfigurecounter{11}

\begin{figure*}[t]
\begin{center}
\includegraphics[width=1.00\textwidth]{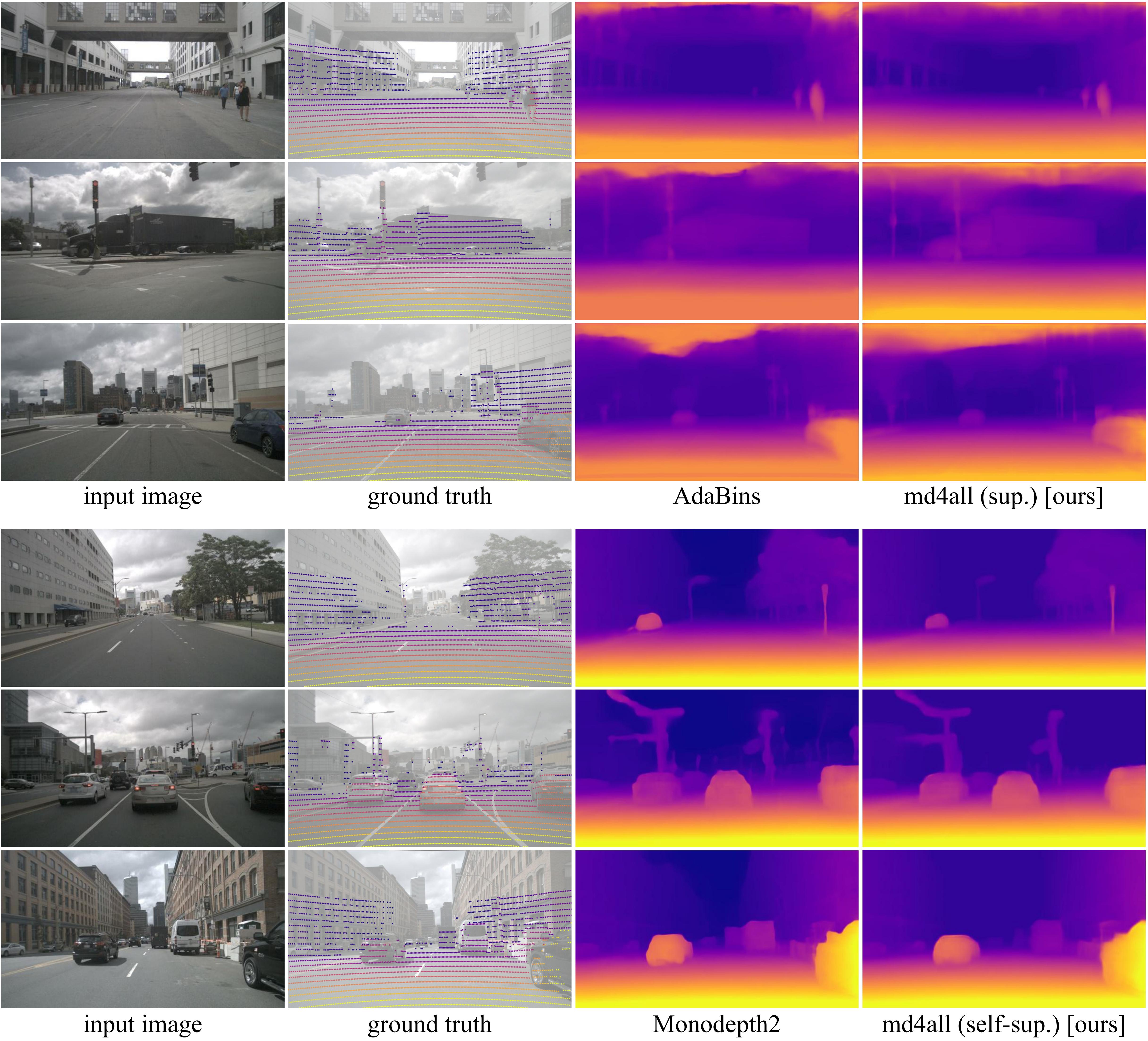}
\vspace{-0.7cm}
\end{center}
   \caption{Comparison of models on nuScenes~\cite{caesar2020nuscenes} \textit{day-clear} samples. In the upper half, the standard AdaBins~\cite{bhat2021adabins} is compared to our \pname-AD applied to AdaBins. In the lower half, the standard Monodepth2~\cite{godard2019monodepth2} is compared to our \pname-DD applied to Monodepth2.}
\label{fig:nuscenes_day_suppl}
\vspace{-0.2cm}
\end{figure*}
%\setfigurecounter{12}

\begin{figure*}[t]
\begin{center}
\includegraphics[width=1.00\textwidth]{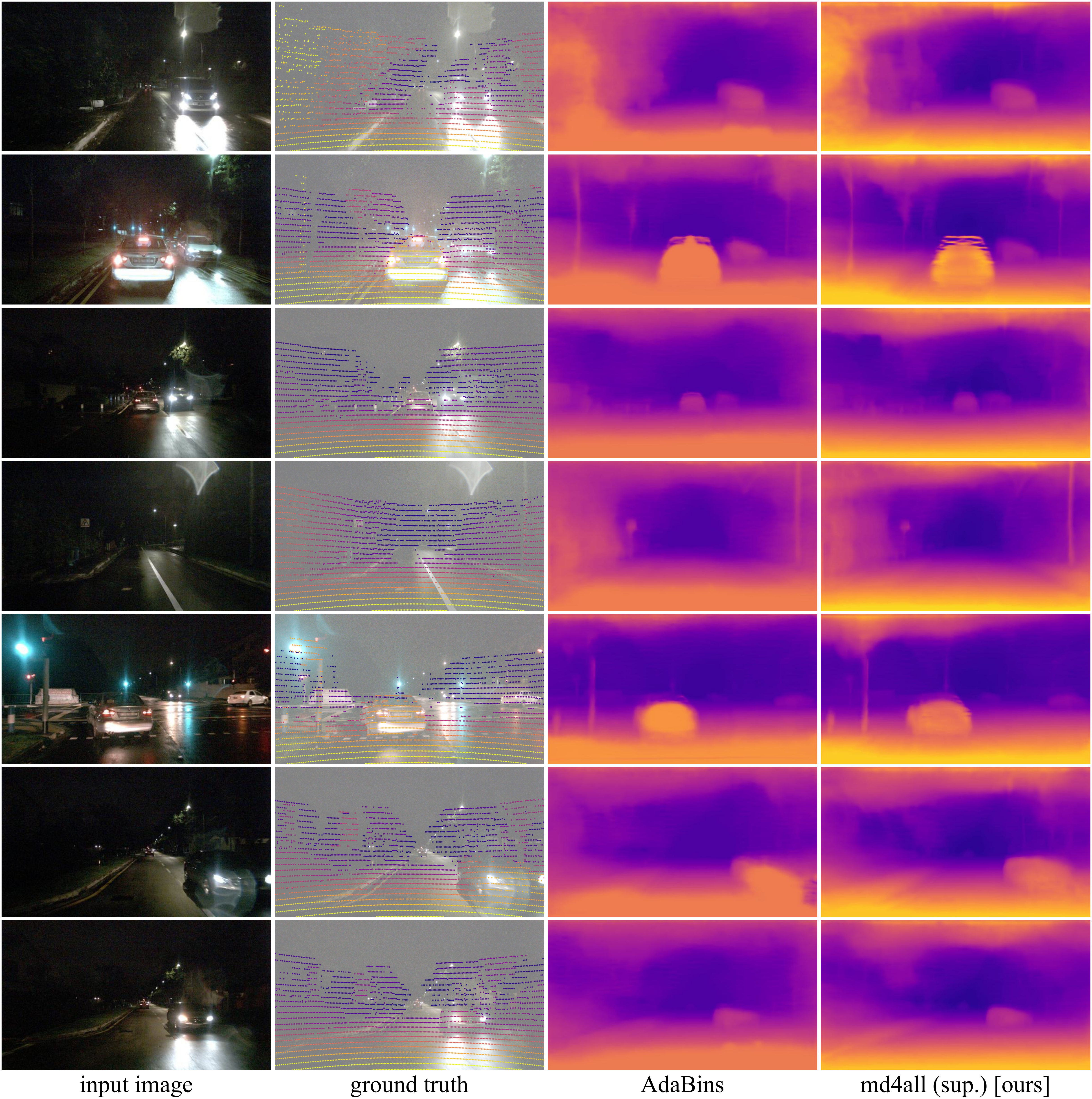}
\vspace{-0.7cm}
\end{center}
   \caption{Comparison of supervised models on nuScenes~\cite{caesar2020nuscenes} \textit{night} samples. The standard AdaBins~\cite{bhat2021adabins} is compared to our \pname-AD applied to AdaBins. This set of samples is particularly challenging due to the overall darkness and reflections.}
\label{fig:nuscenes_supervised_suppl}
\vspace{-0.2cm}
\end{figure*}
%\setfigurecounter{13}

\begin{figure*}[t]
\begin{center}
\includegraphics[width=1.00\textwidth]{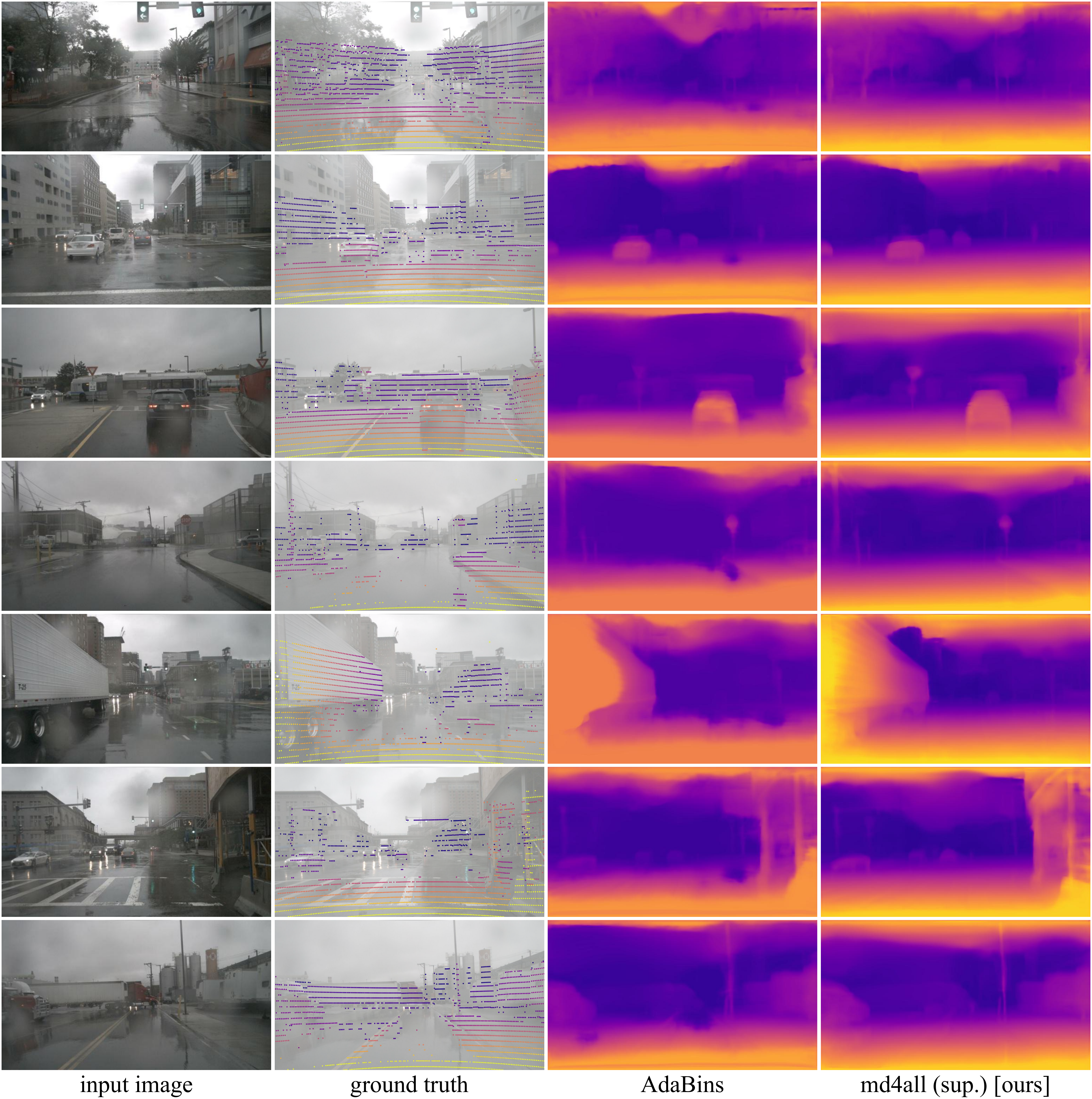}
\vspace{-0.7cm}
\end{center}
   \caption{Comparison of supervised models on nuScenes~\cite{caesar2020nuscenes} \textit{rain} samples. The standard AdaBins~\cite{bhat2021adabins} is compared to our \pname-AD applied to AdaBins.}
\label{fig:nuscenes_rain_suppl}
\vspace{-0.2cm}
\end{figure*}
%\setfigurecounter{14}

\begin{figure*}[t]
\begin{center}
\includegraphics[width=1.00\textwidth]{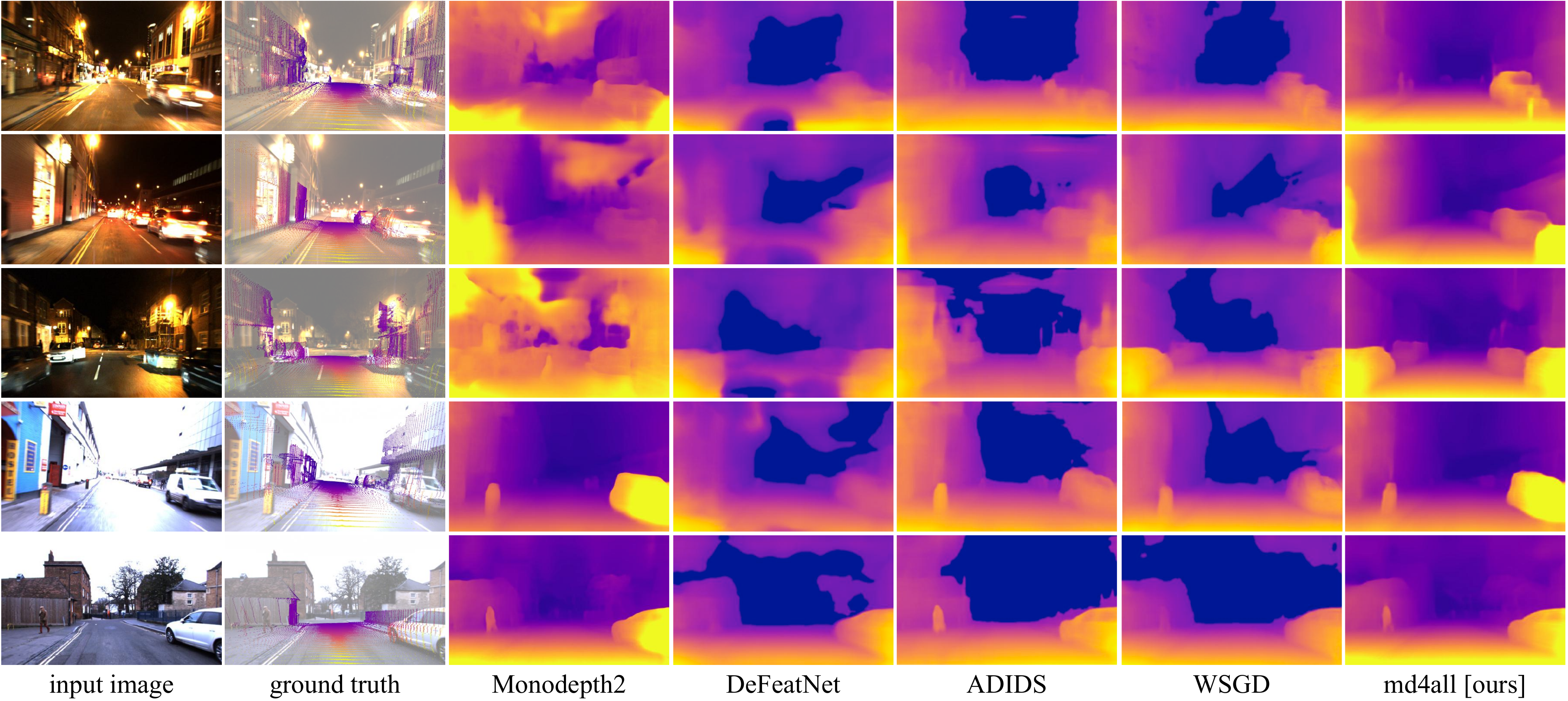}
\vspace{-0.7cm}
\end{center}
   \caption{Comparison of self-supervised models on RobotCar~\cite{maddern2017oxford} \textit{night} and \textit{day} samples. The samples are exactly the ones reported by Vankadari et al.~in WSGD~\cite{vankadari2022sundown}, from which we took directly the predictions of DeFeatNet~\cite{spencer2020defeatnet}, ADIDS~\cite{liu2021allday}, and WSGD~\cite{vankadari2022sundown}. We compare these with Monodepth2~\cite{godard2019monodepth2} and our \pname-DD.}
\label{fig:robotcar_comp_all}
\vspace{-0.2cm}
\end{figure*}
%\setfigurecounter{15}

\begin{figure*}[t]
\begin{center}
\includegraphics[width=1.00\textwidth]{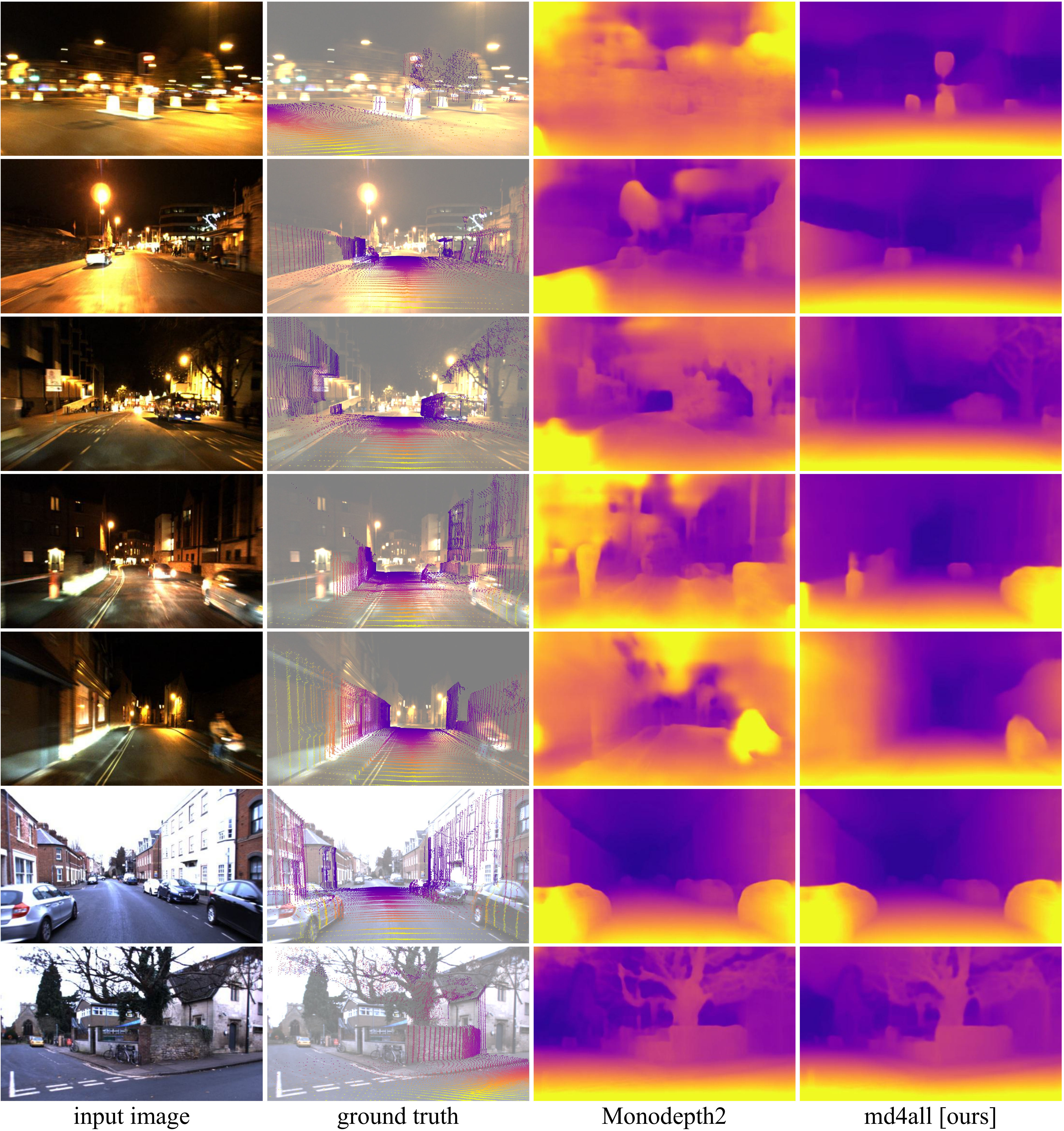}
\vspace{-0.7cm}
\end{center}
   \caption{Comparison of self-supervised models on RobotCar~\cite{maddern2017oxford} \textit{night} and \textit{day} samples. The standard Monodepth2~\cite{godard2019monodepth2} is compared to our \pname-DD applied to Monodepth2.}
\label{fig:robotcar_suppl_md2_ours}
\vspace{-0.2cm}
\end{figure*}
%\setfigurecounter{16}

\begin{figure*}[t]
\begin{center}
\includegraphics[width=1.00\textwidth]{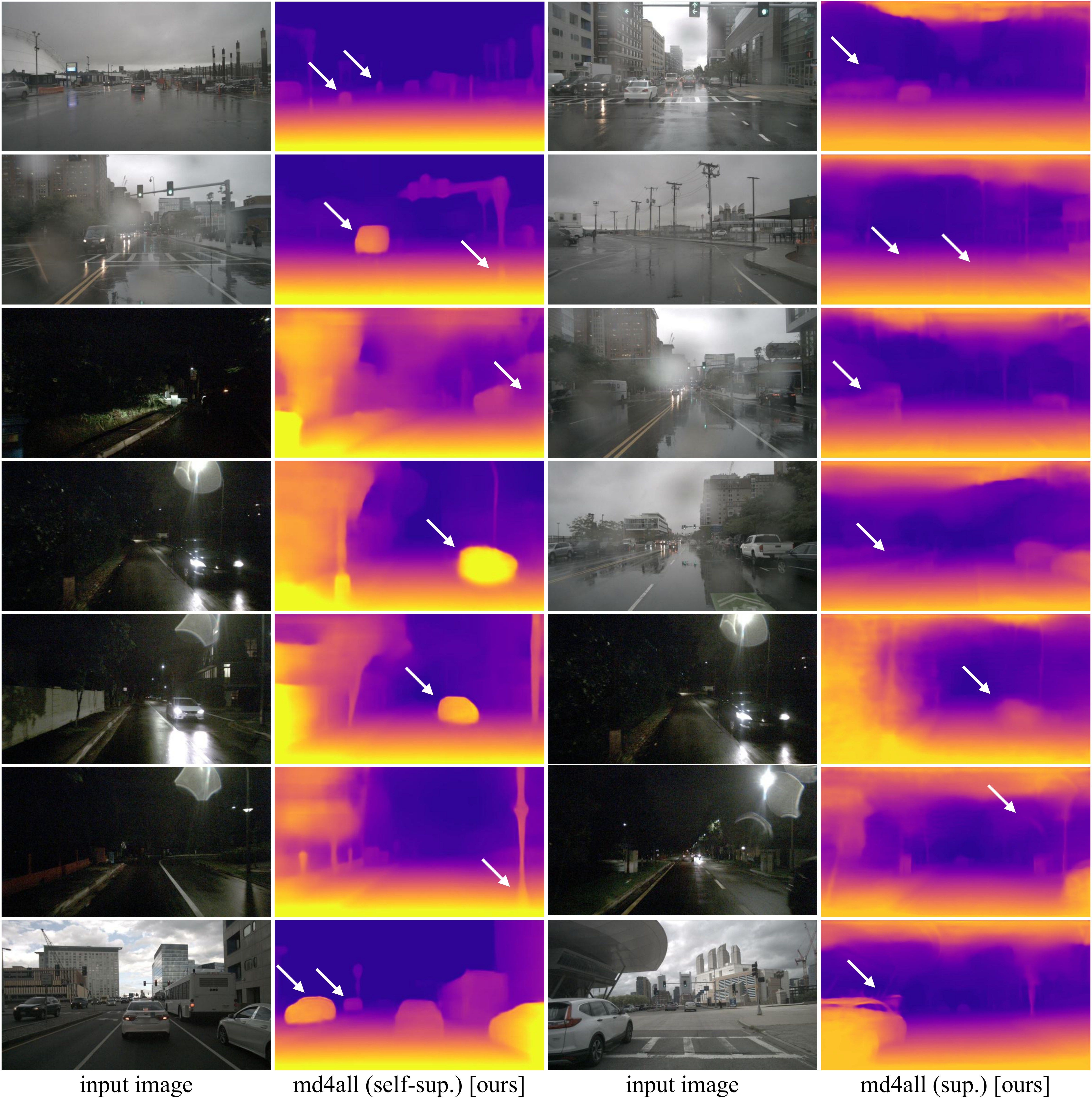}
\vspace{-0.7cm}
\end{center}
   \caption{Failure cases of our self-supervised (\pname-DD, Monodepth2-based) and supervised (\pname-AD, AdaBins-based) models on samples from nuScenes~\cite{caesar2020nuscenes}. White arrows mark issues in the predictions.}
\label{fig:failures}
\vspace{-0.2cm}
\end{figure*}
%\setfigurecounter{17}

\begin{figure*}[t]
\begin{center}
\includegraphics[width=1.00\textwidth]{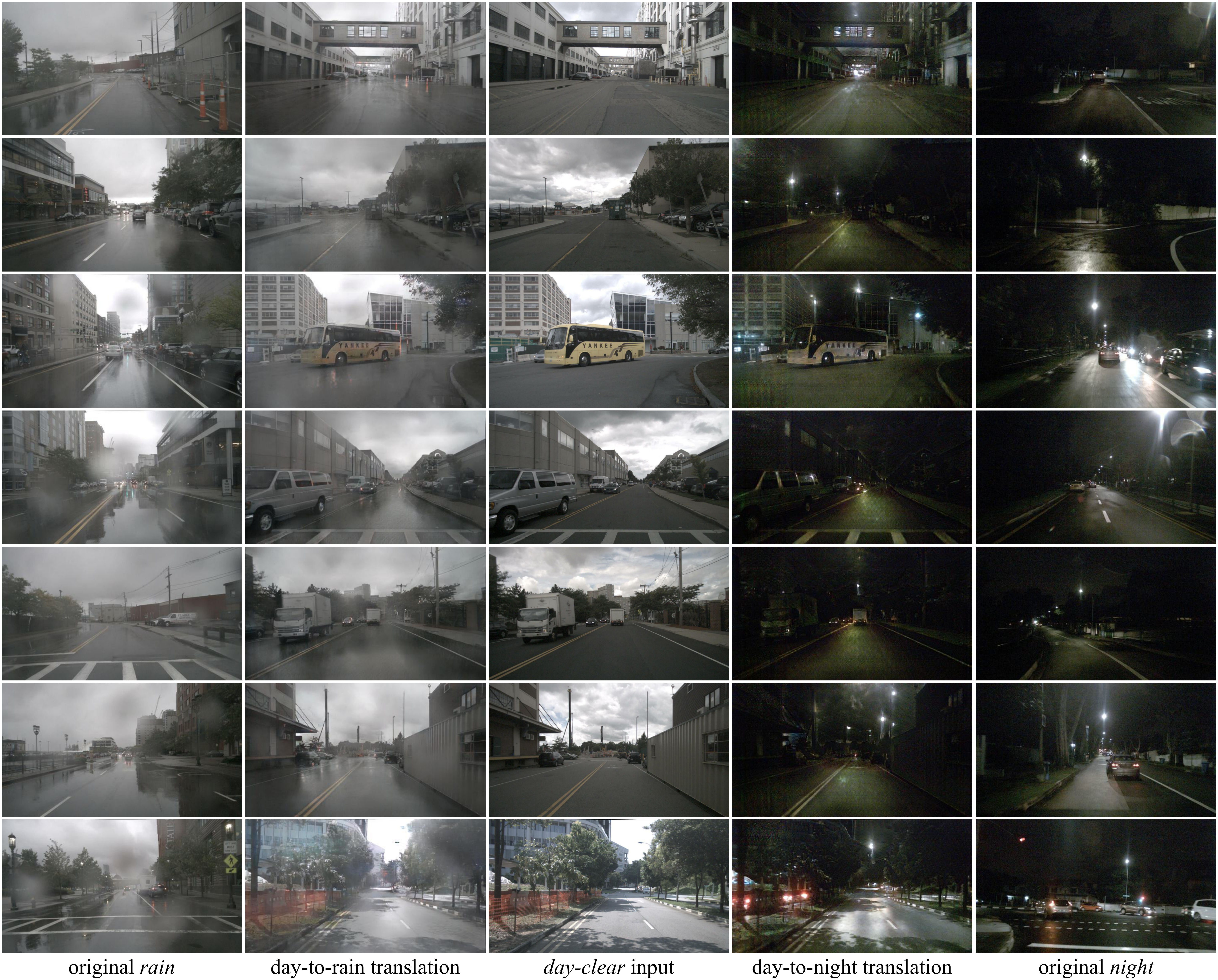}
\vspace{-0.7cm}
\end{center}
   \caption{Example of day-to-adverse image translations on nuScenes~\cite{caesar2020nuscenes}. Training samples are translated from \textit{day-clear} to both \textit{rain} and \textit{night}. For reference, in the first and last columns, we included real \textit{rain} and \textit{night} samples from the validation set. We share publicly the translated \textit{night} and \textit{rain} images for the entire \textit{day-clear} training set.}
\label{fig:translations_nuscenes}
\vspace{-0.2cm}
\end{figure*}
%\setfigurecounter{18}

\begin{figure*}[t]
\begin{center}
\includegraphics[width=1.00\textwidth]{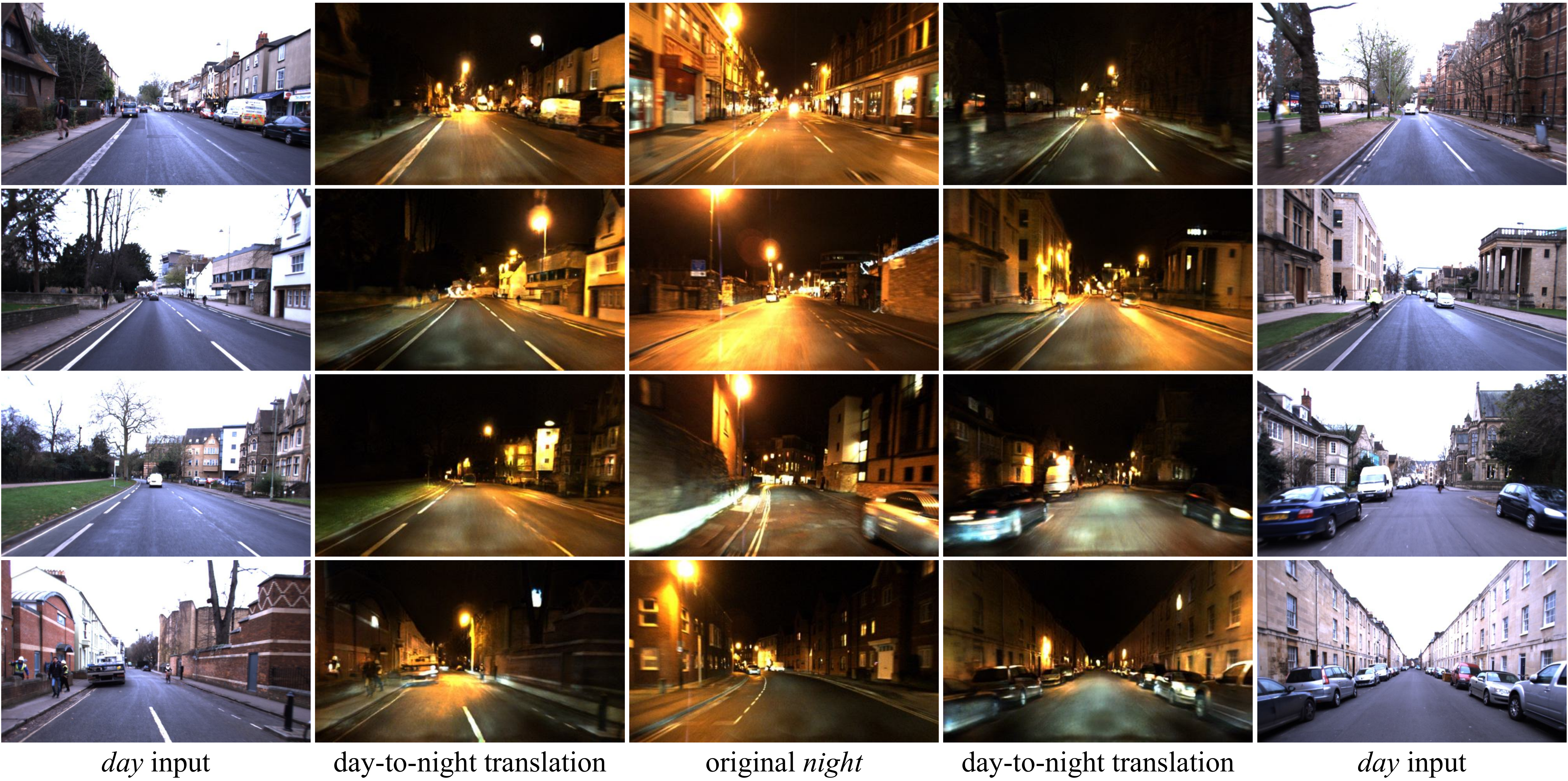}
\vspace{-0.7cm}
\end{center}
   \caption{Example of day-to-adverse image translations on RobotCar~\cite{maddern2017oxford}. Training samples are translated from \textit{day} to \textit{night}. For reference, we included real \textit{night} samples from the test set in the central column. We share publicly the translated \textit{night} images for the entire \textit{day} training set.}
\label{fig:translations_robotcar}
\vspace{-0.2cm}
\end{figure*}
%\setfigurecounter{19}

\subsubsection{nuScenes Test Set}\label{sec:add_res_nuscenes_test}
Table~\ref{table:test_nuscenes} reports errors and metrics on the official test set of nuScenes~\cite{caesar2020nuscenes}. Since no metadata about the weather and illumination conditions is available for the test set, the values are to be considered for \textit{all} conditions combined. As for the validation set, results were computed with the available ground truth data (i.e., LiDAR). The test set of nuScenes is more accessible than its validation set, with the models achieving better performance in the former across all three distance ranges. For these reasons, we focused on the validation set, and we report the test set results here for completeness. Nevertheless, our supervised and self-supervised models performed better than the models they are based on (i.e., AdaBins and Monodepth2, respectively) across the various conditions and depths.

\subsubsection{Evaluation over Different Distances}\label{sec:add_res_distances}
For completeness, we also include the results for self-supervised models computed up to different distance ranges, namely up to 40, 60, and 80 meters. This complements the values reported in the main paper (up to 80 meters for nuScenes, and up to 50 for RobotCar) and should ease comparisons with future works. These results are reported across the following tables:
\begin{itemize}
    \item Tables~\ref{table:40m_nuscenes}, \ref{table:ext_40m_nuscenes}, and~\ref{table:test_nuscenes} for nuScenes up to 40 meters.
    \item Tables~\ref{table:60m_nuscenes}, \ref{table:ext_60m_nuscenes}, and~\ref{table:test_nuscenes} for nuScenes up to 60 meters.
    \item Tables~\reff{1}, \ref{table:fully-sup_nuscenes}, \ref{table:ext_main_nuscenes}, and~\ref{table:test_nuscenes} for nuScenes up to 80 meters.
    \item Table~\ref{table:40m_robotcar} for RobotCar up to 40 meters.
    \item Table~\reff{2} for RobotCar up to 50 meters.
    \item Table~\ref{table:60m_robotcar} for RobotCar up to 60 meters.
    \item Table~\ref{table:80m_robotcar} for RobotCar up to 80 meters.
\end{itemize}
The scores improve at lower distances as the problem becomes more effortless. Changes are limited due to the sparsity of the LiDAR at further distances, especially for RobotCar. While evaluating using more LiDAR frames would have allowed us to evaluate at greater distances, we avoided it as it would have introduced artifacts in the ground truth for dynamic objects. Nevertheless, our models consistently improved across the various distances and conditions tested.
As described in Section~\reff{4.1}, the results we reported are to be considered up to 80 meters for nuScenes, and up to 50 meters for RobotCar, unless otherwise noted.

\subsection{Additional Qualitative Results}\label{sec:add_qualitative}
For all qualitative results reported in this work (also Figures~\reff{1}, \reff{5}, and \reff{6} in the main paper), the predictions of our self-supervised model were all performed by our \pname-DD trained on all conditions (i.e., \textit{dT(nr)} for nuScenes and \textit{dT(n)} for RobotCar) and based on Monodepth2, while the supervised ones were all performed by our \pname-AD trained on \textit{dnT(r)} and based on AdaBins. All qualitative images of the standard Monodepth2 were produced by the model trained only on \textit{day-clear} (\textit{day} for RobotCar). In contrast, the ones for the standard AdaBins were made by training it on all conditions, meaning that the best scoring models of each type produced them.

The following sections introduce new qualitative results on the various settings and conditions, such as nuScenes with self-supervised models (Section~\ref{sec:add_res_qual_nuscenes_self}), nuScenes with supervised methods (Section~\ref{sec:add_res_qual_nuscenes_fully}), RobotCar with self-supervised approaches (Section~\ref{sec:add_res_qual_robotcar}), failure cases (Section~\ref{sec:add_res_qual_failures}), and translated images (Section~\ref{sec:add_res_qual_translations}).

\subsubsection{nuScenes -- Self-Supervised}\label{sec:add_res_qual_nuscenes_self}

\textbf{\textit{Night} -- nuScenes}
In Figure~\ref{fig:nuscenes_dark}, we compare our \pname-DD to Monodepth2~\cite{godard2019monodepth2} on particularly challenging \textit{night} samples due to the extreme darkness levels, as well as the high amounts of reflections (also due to the wet ground, in \textit{night-rain} conditions). Instead, in Figure~\ref{fig:nuscenes_bright}, we compare the same models on brighter, i.e., easier samples. It can be seen that the standard Monodepth2 trained on \textit{day-clear} had marked difficulties with darker scenes (Figure~\ref{fig:nuscenes_dark}), while it delivered satisfactory results in brighter settings (Figure~\ref{fig:nuscenes_bright}). This significant difference can be attributed to the fact that Monodepth2 was trained only on \textit{day-clear} (the model trained on \textit{all} conditions performed significantly worse, especially at \textit{night}, as shown in Table~\reff{1}) and that the details are easier to grasp in brighter \textit{night} inputs compared to very dark ones. In brighter images, the road, vehicles, curb, grass, and trees on the sides can be seen relatively well thanks to the light emitted by the streetlights, allowing the standard Monodepth2 to estimate reasonable depth maps. Conversely, with darker scenes, the depth cues are highly challenging to extract due to the noise and general blackness.
Nevertheless, Monodepth2 tended to estimate the ground well also in darker settings. This can be due to the model being biased toward scenes with flat and regular roads in front of the ego vehicle and the road being the most visible part of the images. Instead, Monodepth2 had severe issues with the sky, which is not evaluated since the ground truth is unavailable. Nevertheless, the proposed \pname\ delivered good and sharp estimates in dark and bright environments.

\textbf{\textit{Rain} -- nuScenes}
In Figure~\ref{fig:nuscenes_rain_self_suppl}, we compare our \pname-DD to Monodepth2~\cite{godard2019monodepth2} on \textit{rainy} samples of nuScenes. While the standard Monodepth2 trained on \textit{day-clear} delivered reasonable estimates for the most part, it was severely affected by the reflections on the wet ground (e.g., first row). Also, the blurriness due to water drops affected Monodepth2, triggering overly smooth estimates. The figure shows also Monodepth2 having issues estimating the depth of the sky (second and fifth row). Since none of these problems occurs without rain (i.e., \textit{day-clear}, shown in Figure~\ref{fig:nuscenes_day_suppl}), they can be attributed to the challenging weather conditions. Instead, our \pname\ delivered reasonable depth estimates regardless of the adverse settings, because its supervision signal was always from ideal settings.

\textbf{\textit{Day-clear} -- nuScenes}
In the lower half of Figure~\ref{fig:nuscenes_day_suppl}, we compare the performance at daytime. The \textit{day-clear} scores of Monodepth2 (\textit{d}) and our \pname-DD \textit{a} are relatively close to one another (Table~\reff{1}). Based on the same model (i.e., Monodepth2), the depth maps of the two are similar, with ours delivering smoother surfaces (e.g., vehicles), with fewer artifacts around the objects, such as the black car on the left of the bottom input. Our extra smoothness is due to the day distillation scheme, propagating dense estimates from the baseline. Since our same model predicts significantly better depth maps in adverse conditions than Monodepth2, the figure confirms that our \pname\ does not degrade in ideal conditions. This proves the effectiveness of our techniques, as they do not introduce any performance trade-off in standard settings while enabling significantly better outcomes in challenging conditions.

\subsubsection{nuScenes -- Supervised}\label{sec:add_res_qual_nuscenes_fully}

\textbf{\textit{Night} -- nuScenes}
In Figure~\ref{fig:nuscenes_supervised_suppl}, we compare our \pname-AD to AdaBins~\cite{bhat2021adabins} on particularly challenging \textit{night} samples due to the extreme darkness levels and the high amount of reflections (same samples as in Figure~\ref{fig:nuscenes_dark}). Our \pname-AD delivered overall sharper and more accurate estimates, as can be seen for the thin structures (e.g., poles and tree trunks), as well as the boundaries of the vehicles in the bottom two rows. Although all depth maps are displayed with the same colormap, the standard AdaBins never estimated depths that triggered the use of the bright yellow color (i.e., closest distance). This problem is less pronounced during daytime (Figure~\ref{fig:nuscenes_day_suppl}). This can be attributed to the standard AdaBins suffering from overfitting on the sparse ground truth data of nuScenes. Towards this end, we had to apply early stopping to prevent severe horizontal artifacts from appearing on the outputs following the LiDAR detections. Applying our \pname\ on AdaBins introduced challenging augmentations during training, reducing overfitting and allowing it to train longer. This led to a model that estimated depth throughout a more extensive range without the horizontal artifacts. The qualitative results align with the scores gap in Table~\reff{1}. Moreover, some artifacts are noticeable for both models around the roof of the vehicles. These are due to the ground truth LiDAR data reporting further estimates in those areas (e.g., fifth row). This is caused by the relative position of the LiDAR sensor to the camera, with the former being at a higher location, thereby seeing beyond objects compared to the camera's perspective.

\textbf{\textit{Rain} -- nuScenes}
In Figure~\ref{fig:nuscenes_rain_suppl}, we compare our \pname-AD to AdaBins~\cite{bhat2021adabins} on \textit{rain} samples of nuScenes. As the standard AdaBins learned from the erroneous measurements of the LiDAR ground truth (e.g., Figure~\reff{2}), it systematically estimated artifacts on the ground, resulting in holes in its depth maps. The erroneous LiDAR measurements can also be seen from the ground truth reported for each image. Such wrong measurements are caused systematically by each highly reflective object (e.g., traffic sign) reflected on the wet ground. Therefore, training on such a wrong signal causes the standard AdaBins to replicate the artifacts in its output (e.g., in front of the stop sign in the fourth row). Nevertheless, thanks to the reliable training signal from ideal conditions, our \pname-AD delivered good estimates without such artifacts. 

\textbf{\textit{Day-clear} -- nuScenes}
In the upper half of Figure~\ref{fig:nuscenes_day_suppl}, we compare the performance at daytime. The scores between AdaBins and our \pname-AD trained on all conditions favor ours (Table~\reff{1}). This can be seen from the depth maps, with ours delivering smoother estimates for the road and sharper details throughout. Ours separated the truck in the second frame from the background. As discussed above, this can be attributed to the strong tendency of the standard AdaBins to overfit. Instead, by feeding a mix of original and translated images, our method acts as data augmentation, mitigating the problem. Nevertheless, the limitations of such a sparse ground truth signal are evident, with smooth edges due to its sparsity and see-through effects caused by the displacement between the LiDAR and the camera. For these reasons, self-supervised outputs look better overall, as the models are also less prone to overfitting.

\subsubsection{RobotCar -- Self-Supervised}\label{sec:add_res_qual_robotcar}

\textbf{Comparison with other \textit{night} methods}
In Figure~\ref{fig:robotcar_comp_all}, we compare our \pname-DD with various other methods targeting depth estimation in \textit{night} conditions, namely DeFeatNet~\cite{spencer2020defeatnet}, ADIDS~\cite{liu2021allday}, and WSGD~\cite{vankadari2022sundown}, as well as the standard Monodepth2~\cite{godard2019monodepth2} designed for ideal conditions. The samples shown are exactly the ones displayed by WSGD~\cite{vankadari2022sundown} in their paper, from which we took the outputs of~\cite{spencer2020defeatnet,liu2021allday,vankadari2022sundown}. However, unlike theirs~\cite{vankadari2022sundown}, for ours and Monodepth2, we do not manually threshold the maximum depth, showing the entire depth estimation, including further distances. This is possibly the reason for theirs being artificially dark in the background.
Remarkably, as seen already in Table~\reff{2} and Figure~\reff{6}, our \pname-DD delivered more accurate and sharper estimates in both conditions, and especially at \textit{night}, thanks to its robust feature extraction suitable for both \textit{day} and \textit{night} inputs.

\textbf{\textit{Night} -- RobotCar}
Figure~\ref{fig:robotcar_suppl_md2_ours} shows additional outputs of the standard Monodepth2~\cite{godard2019monodepth2} compared to our \pname-DD applied on Monodepth2. Although the \textit{night} scenes are not as dark and not as noisy as those from nuScenes~\cite{caesar2020nuscenes} (e.g., Figure~\ref{fig:nuscenes_dark}), Monodepth2 had major issues estimating the depth of the image. Darkness and streetlights were detrimental for the standard Monodepth2 (e.g., in the second row). Moreover, compared to nuScenes, the images from RobotCar are often blurry at night, especially those from turns (first row). Nevertheless, the proposed \pname\ estimated reasonable depth maps regardless of these issues in the input. Due to the textureless pure white sky in the \textit{day} samples of RobotCar (last two rows), the models had issues capturing its depth during training. The same occurred for the top of buildings, which are often too bright and indistinguishable from the sky in the images (e.g., fourth input in Figure~\ref{fig:robotcar_comp_all}). This problem is inherent to the data itself and causes the sky to always be predicted to be relatively close (brighter color). This did not happen for nuScenes.

\textbf{\textit{Day} -- RobotCar}
Figure~\ref{fig:robotcar_suppl_md2_ours} also shows additional predictions during daytime (bottom rows). As seen in Table~\reff{2}, while performing significantly better at \textit{night}, the performance of the proposed method does not degrade during the \textit{day} compared to the standard Monodepth2~\cite{godard2019monodepth2}. Due to the knowledge distillation from the baseline to our \pname-DD model, surfaces result smoother, but edges remain sharp. The bottom input is particularly challenging as it features a turn. Both models correctly estimated the scene's depth, with Monodepth2 delivering more details on the tree in the foreground but less on the tree in the background (left). As discussed for the \textit{night} scenes, the textureless white sky of RobotCar prevents the models from learning its depth correctly, which causes erroneous estimations at test time. Instead, nuScenes~\cite{caesar2020nuscenes} includes different sky conditions (e.g., sunny and cloudy), allowing the models to learn its depth.

\subsubsection{Failure Cases}\label{sec:add_res_qual_failures}
While our techniques bring significant improvements across various conditions, there is still room for improvement. Figure~\ref{fig:failures} reports failure cases of our models. Inherited from Monodepth2~\cite{godard2019monodepth2}, our self-supervised model has issues with dynamic objects, especially oncoming traffic, whose distance is wrongly estimated due to the violations of the moving camera in a static world assumption~\cite{gasperini2021r4dyn}. This occurs in all three conditions. Furthermore, the model is occasionally fooled by reflections on the ground or misleading shadows which look like objects (first and sixth row). Additionally, highly dark scenes are also challenging (third row) due to the lack of information across nearly black pixels.
These issues could be mitigated by integrating the data from the cost-effective radar, as in R4Dyn~\cite{gasperini2021r4dyn}, which is robust against adverse weather and challenging illumination conditions. Our method is not bound to a specific architecture or pipeline, so it could be applied to R4Dyn directly.
Another drawback our \pname\ inherited from Monodepth2, which we did not address, is the lack of temporal consistency in the depth predictions.
Instead, for the supervised setting, we focused on eliminating the artifacts due to the erroneous ground truth measurements. While such issues are appropriately addressed, other problems persist, such as with particular reflections and the blur caused by water drops (third row). Radar may help here too.

\subsubsection{Day-to-Adverse Translation}\label{sec:add_res_qual_translations}
This section shows samples of the translated images we generated with ForkGAN~\cite{zheng2020forkgan}. We publicly share all translated images corresponding to the ideal settings from the training sets of nuScenes~\cite{caesar2020nuscenes} and RobotCar~\cite{maddern2017oxford}.

\textbf{nuScenes} In Figure~\ref{fig:translations_nuscenes}, we show examples of the translated images used to train our models on nuScenes. The GAN~\cite{zheng2020forkgan} added plausible reflections and lights simulating wet ground or streetlights. While these additions are not always realistic, they bring challenging variations to the training set, resembling the adverse conditions in the first and last columns. The standard \textit{day-clear} training set includes a mix of sunny and cloudy scenes. For cloudy ones, shadows are limited, with less contrast overall, making it easier for the translation model. Instead, the inputs with sunny conditions are particularly challenging for the GAN, resulting in less convincing outputs (e.g., in the last row). The GAN also learned to add water drops blurring certain areas for \textit{rain}, and intense noise for \textit{night}. The results of our models would directly benefit from improvements in the translations towards greater realism. With perfect translations, our \pname\ would further reduce the gap between ideal and challenging conditions.

\textbf{RobotCar} In Figure~\ref{fig:translations_robotcar}, we report examples of the translated images used on RobotCar. In this case, the GAN was trained entirely on RobotCar, thanks to the high amount of \textit{night} samples available. Since the weather in the \textit{day} samples tends to be always cloudy, with a consistently white sky, the GAN did not exhibit the issues seen for nuScenes with sunny inputs (Figure~\ref{fig:translations_nuscenes}). Instead, the GAN delivered highly plausible samples with a higher degree of realism than for nuScenes, enabling our model to narrow the margin between daytime and nighttime performances. As seen for nuScenes, the GAN added streetlights resembling the ones seen throughout the dataset. Furthermore, it can be seen clearly that the GAN learned the headlights of the ego vehicle, which it added to illuminate the road ahead.

For both datasets, while being somewhat plausible, the translated samples are not perfect. For nuScenes, for example, noise patterns in the \textit{night} translations are repeated similarly over the lower left corner of the images. The noise is more unstructured for the real \textit{night} images. Additionally, the GAN did not learn to turn on the lights of other cars, which are often a source of issues for the models. 

\subsection{Attempted Approaches That Did Not Work}\label{sec:add_not_work}
We explored several alternative solutions for this challenging problem, and we intend to mention them in this section to help future researchers who target the same issues. As written in Section~\reff{4.1}, we experimented with diffusion models to learn our day-to-adverse image translation. However, the lack of paired images for \textit{day} and \textit{night} made it unfeasible. Therefore, we opted for GANs, which do not need paired inputs. Furthermore, we experimented with the following methods on the \textit{night} samples of nuScenes, seeking improvements over the standard Monodepth2. Spatial and temporal attention could have allowed the model to focus on valuable information. Still, it did not bring an improvement, possibly due to the large amount of noise varying across different \textit{night} images, which could have prevented to attend on the helpful information. Additionally, we experimented with incorporating geometrical priors (e.g., from edge detection) into the losses, but they were similarly not beneficial. The simple solution presented in this work was the most effective to tackle this complex problem.

%\clearpage

%% file: tables/ablations_main_nuscenes.tex
\begin{table*}
\begin{center}
\begin{tabular}{l|ll|ccc|ccc|ccc}
\toprule
&&& \multicolumn{3}{c|}{\textit{day-clear} -- nuScenes} & \multicolumn{3}{c|}{\textit{night} -- nuScenes} & \multicolumn{3}{c}{\textit{day-rain} -- nuScenes} \\
ID & Method & tr.data & absRel & RMSE & $\delta_1$   & absRel & RMSE & $\delta_1$   & absRel & RMSE & $\delta_1$  \\
\midrule

A0 & md2~\cite{godard2019monodepth2}, \textit{all} & \textit{a: dnr} & 0.1477 & 6.771 & 85.25 & 2.3332 & 32.940 & 10.54 & 0.4114 & 9.442 & 60.58   \\
A1 & md2, \textit{n} real & \textit{dn} & 0.1345 & 6.575 & \underline{85.47} & 2.4536 & 34.295 & 11.71  & 0.1753 & 7.701 & 77.13 \\
A2 & md2, \textit{n} transl.$_{15\%}$ & \textit{dT(n)} & 0.1390 & 6.670 & \dotuline{85.36} & 0.2655 & 9.892 & 54.44 & 0.1861 & 7.800 & 76.28  \\
A3 & md2, \textit{day-c} only & \textit{d} & 0.1374 & 6.692 & 85.00 & 0.2828 & 9.729 & 51.83 & 0.1727 & 7.743 & 77.57 \\
A4 & + v.-sup = b.line & \textit{d} & \dotuline{0.1333} & 6.459 & \textbf{85.88} & 0.2419 & 10.922 & 58.17 & 0.1572 & 7.453 & 79.49 \\
A5 & + noise, $\mathcal{L}$ clean & \textit{d} & 0.1428 & 6.609 & 84.43 & 0.2256 & 9.672 & 63.50 & 0.1592 & 7.619 & 78.95  \\
A6 & + all \textit{n} transl. & \textit{dT(n)}  & 0.1624 & 7.042 & 80.50 & 0.2214 & 9.092 & 67.01 & 0.1752 & 8.272 & 76.41  \\
A7 & -- pose transl. & \textit{dT(n)}  & 0.1597 & 7.143 & 81.37 & 0.2184 & 8.754 & 66.90 & 0.1689 & 8.210 & 77.23 \\
AD\textit{n} & + \textit{day} loss only  & \textit{dT(n)} & 0.1433 & 6.954 & 83.27 & 0.2230 & 9.001 & 68.63 & 0.1545 & 7.915 & 78.36 \\
A9 & -- time norm. & \textit{dT(n)} & 0.1554 & 6.949 & 81.66 & 0.2121 & \underline{8.502} & 67.43 & 0.1627 & 7.797 & 77.62 \\
AD\textit{a} & AD\textit{n} + \textit{r} transl.  & \textit{dT(nr)} & 0.1523 & 6.853 & 83.11 & 0.2187 & 9.003 & 68.84 & 0.1601 & 7.832 & 78.97 \\
A11 & + day distill. & \textit{dT(nr)} & 0.1387 & 6.621 & 84.11 & 0.1960 & 8.595 & 70.08 & 0.1444 & 7.355 & 80.20 \\
DD\textit{n} & AD\textit{n} + distill.  & \textit{dT(n)} & \textbf{0.1302} & \textbf{6.373} & 85.02 & \dotuline{0.1959} & \textbf{8.471} & \dotuline{70.14} & 0.1429 & 7.312 & 79.60 \\
DD\textit{r} & \textit{r} distill. & \textit{dT(r)} & \underline{0.1323} & \underline{6.437} & 85.18 & 0.2502 & 11.847 & 57.02 & \textbf{0.1364} & \textbf{7.100} & \textbf{81.37} \\
\textbf{DD\textit{a}} & AD\textit{a} + distill. & \textit{dT(nr)} & 0.1366 & 6.452 & 84.61 & \underline{0.1921} & \dotuline{8.507} & \textbf{71.07} & \dotuline{0.1414} & \underline{7.228} & \dotuline{80.98} \\

A15 & -- test time norm. & \textit{dT(nr)} & 0.1367 & \dotuline{6.449} & 84.56 & \textbf{0.1881} & 8.524 & \underline{70.65} & \underline{0.1412} & \dotuline{7.234} & \underline{80.99} \\

\bottomrule
\end{tabular}
\end{center}
\vspace{-0.2cm}
\caption{Ablation study on our self-supervised method on the nuScenes~\cite{caesar2020nuscenes} validation set. md2: Monodepth2~\cite{godard2019monodepth2}. DD\textit{a} is our self-supervised approach reported throughout this work for nuScenes: \pname-DD trained on \textit{dT(nr)}. Instead, AD\textit{a} is our self-supervised \pname-AD model. Notation reused from Table~\reff{1}.
}
\label{table:ablations_main_nuscenes}
%\vspace{-0.2cm}
\end{table*}

%% file: tables/ablations_main_robotcar.tex
\begin{table*}
\setlength{\tabcolsep}{5.9pt}
\begin{center}
\begin{tabular}{ll|cccc|cccc}
\toprule
&& \multicolumn{4}{c|}{\textit{day} -- RobotCar} & \multicolumn{4}{c}{\textit{night} -- RobotCar} \\
Method & tr.data & absRel & sqRel & RMSE & $\delta_1$   & absRel & sqRel & RMSE & $\delta_1$   \\
\midrule

Monodepth2~\cite{godard2019monodepth2} & \textit{d} & 0.1196 & \dotuline{0.670} & \textbf{3.164} & 86.38 & 0.3029 & 1.724 & 5.038 & 45.88 \\

WSGD~\cite{vankadari2022sundown} & \textit{a}: \textit{dn} & 0.1760 & 1.603 & 6.036 & 75.00 & 0.1740 & 1.637 & 6.302 & 75.40 \\

[ours] baseline & \textit{d} & 0.1209 & 0.723 & 3.335 & 86.61 & 0.3909 & 3.547 & 8.227 & 22.51 \\

[ours] \pname-AD & \textit{dT(n)} & \textbf{0.1113} & 0.707 & 3.248 & \textbf{88.02} & \underline{0.1223} & \dotuline{0.851} & \dotuline{3.723} & \textbf{85.77} \\

[ours] \textbf{\pname-DD} & \textit{dT(n)} & \underline{0.1128} & \underline{0.648} & \dotuline{3.206} & \underline{87.13} & \textbf{0.1219} & \textbf{0.784} & \textbf{3.604} & \underline{84.86} \\

[ours] \pname-DD w/o test time norm. & \textit{dT(n)} & \dotuline{0.1129} & \textbf{0.640} & \underline{3.190} & \dotuline{87.02} & \dotuline{0.1256} & \underline{0.824} & \underline{3.703} & \dotuline{83.87} \\

\midrule

[ours] \pname-AD w/ LiDAR scaling & \textit{dT(n)} & 0.1192 & 0.747 & 3.184 & 86.81 & 0.1275 & 0.834 & 3.641 & 86.15 \\

[ours] \pname-DD w/ LiDAR scaling & \textit{dT(n)} & \textbf{0.1133} & \textbf{0.642} & \textbf{3.052} & \textbf{87.45} & \textbf{0.1230} & \textbf{0.739} & \textbf{3.439} & \textbf{86.41} \\
\bottomrule
\end{tabular}
\end{center}
\vspace{-0.2cm}
\caption{Evaluation of self-supervised works on the RobotCar~\cite{maddern2017oxford} test set up to 50 meters. Different configurations of our method are compared with Monodepth2 and WSGD. LiDAR scaling indicates the use of LiDAR data at test-time to scale the predictions (as done by Monodepth2 and WSGD), which is equivalent to $^*$ in the supervision notation of Table~\reff{1} and~\reff{2}. The method highlighted in bold is reported throughout this work for RobotCar as \pname-DD trained on \textit{dT(n)}. This Table extends Table~\reff{2}.}
\label{table:ablation_robotcar}
%\vspace{-0.2cm}
\end{table*}

%% file: tables/dense.tex
\begin{table}[t]
\setlength{\tabcolsep}{3.5pt}
\begin{center}
%\begin{adjustbox}{max width=\linewidth}
\begin{tabular}{l|cc|cc|cc}
\toprule

 & \multicolumn{2}{c|}{\textit{day-clear}} & \multicolumn{2}{c|}{\textit{fog}} & \multicolumn{2}{c}{\textit{snow}} \\
Method & absRel & $\delta_1$ & absRel & $\delta_1$ & absRel & $\delta_1$ \\
%Method & \textit{day-clear} & \textit{fog} & \textit{snow} \\

\midrule

md2~\cite{godard2019monodepth2} & 0.1642 & 82.35 & 0.1698 & 81.97 & 0.1798 & 76.68 \\

[ours] & \textbf{0.1520} & \textbf{83.54} & \textbf{0.1524} & \textbf{83.36} & \textbf{0.1788} & \textbf{77.93} \\

\bottomrule
\end{tabular}
%\end{adjustbox}
\end{center}
\vspace{-0.2em}
\caption{
\uline{These are only preliminary results} (details in Section~\ref{sec:add_res_dense}). Evaluations with \textit{snow} and \textit{fog} on the DENSE dataset~\cite{bijelic2020dense}. AbsRel and $\delta_1$ are reported for each condition. Monodepth2~\cite{godard2019monodepth2} (md2) trained on \textit{day-clear} is compared with our \pname-DD trained on \textit{day-clear} plus translated images to \textit{fog} and \textit{snow} ($x=66\%$).
}
\label{table:dense}
\vspace{-0.2em}
\end{table}

%% file: tables/distributions.tex
\begin{table}[t]
%\vspace{-0.5em}
\setlength{\tabcolsep}{3.3pt}
\begin{center}
%\begin{adjustbox}{max width=\linewidth}
\begin{tabular}{l|cc|cc|cc}
\toprule

 & \multicolumn{2}{c|}{\textit{avg/all}} & \multicolumn{2}{c|}{\textit{day}} & \multicolumn{2}{c}{\textit{night}} \\
Method & absRel & $\delta_1$ & absRel & $\delta_1$ & absRel & $\delta_1$ \\

\midrule

md2~\cite{godard2019monodepth2} & 0.2122 & 65.92 & 0.1196 & 86.38 & 0.3029 & 45.88 \\
70\textit{d} - 30\textit{n} & 0.1189 & \textbf{86.39} & 0.1138 & \textbf{87.80} & 0.1239 & \textbf{85.01} \\
\textbf{50\textit{d} - 50\textit{n}} & \textbf{0.1174} & 85.99 & \textbf{0.1128} & 87.13 & \textbf{0.1219} & 84.86 \\
30\textit{d} - 70\textit{n} & 0.1221 & 85.86 & 0.1168 & 87.16 & 0.1273 & 84.59 \\

\bottomrule
\end{tabular}
%\end{adjustbox}
\end{center}
\vspace{-0.2em}
\caption{Impact of different training data distributions between \textit{day} (\textit{d}) and \textit{night} (\textit{n}) samples by varying the parameter \textit{x}. AbsRel and $\delta_1$ are reported on the test set of the RobotCar dataset~\cite{maddern2017oxford}. Different distributions are shown for \pname-DD, \textit{a}. In the rest of this work, the balanced 50\textit{d}-50\textit{n} configuration (i.e., $x=50\%$) was used.}
\label{table:distributions}
\vspace{-0.2em}
\end{table}

%% file: tables/fully-sup_nuscenes.tex
\begin{table*}
\begin{center}
\begin{tabular}{ll|ccc|ccc|ccc}
\toprule
&& \multicolumn{3}{c|}{\textit{day-clear} -- nuScenes} & \multicolumn{3}{c|}{\textit{night} -- nuScenes} & \multicolumn{3}{c}{\textit{day-rain} -- nuScenes} \\
Method & tr.data & absRel & RMSE & $\delta_1$   & absRel & RMSE & $\delta_1$   & absRel & RMSE & $\delta_1$  \\
\midrule
AdaBins~\cite{bhat2021adabins} & \textit{a}: \textit{dnr} & 0.1384 & 5.582 & 81.31 & \underline{0.2296} & \underline{7.344} & \underline{63.95} & 0.1726 & 6.267 & 76.01 \\
AdaBins~\cite{bhat2021adabins} & \textit{d} & \underline{0.1138} & \underline{4.805} & \dotuline{87.98} & 0.3336 & 14.002 & 45.77 & \underline{0.1540} & \dotuline{6.119} & \dotuline{81.20} \\

[ours] \pname-AD, \textit{rain} & \textit{dT(r)} & \textbf{0.1052} & \textbf{4.621} & \textbf{89.58} & \dotuline{0.2644} & \dotuline{10.749} & \dotuline{55.51} & \textbf{0.1380} & \underline{6.030} & \textbf{83.32}\\

%[ours] \pname-AD \textit{a} \textit{dT(nr)} & 0.1052 & 4.590 & 88.93 & 0.2513 & 8.282 & 63.63 & 0.1453 & 5.938 & 82.70 \\

[ours] \textbf{\pname-AD, \textit{all}} & \textit{dnT(r)} & \dotuline{0.1206} & \dotuline{4.806} & \underline{88.03} & \textbf{0.1821} & \textbf{6.372} & \textbf{75.33} & \dotuline{0.1562} & \textbf{5.903} & \underline{82.82} \\

\bottomrule
\end{tabular}
\end{center}
\vspace{-0.2cm}
\caption{Additional evaluation of LiDAR-supervised methods on the nuScenes~\cite{caesar2020nuscenes} validation set. This table adds the second and third lines compared to Table~\reff{1}.}
\label{table:fully-sup_nuscenes}
\vspace{-0.2cm}
\end{table*}

%% file: tables/ext_main_nuscenes.tex
\begin{table*}
\setlength{\tabcolsep}{5.6pt}
\begin{center}
\begin{tabular}{l|cccc|cccc|c|c|c}
\toprule
& \multicolumn{4}{c|}{\textit{avg/all} -- nuScenes} & \multicolumn{4}{c|}{\textit{night-rain} -- nuScenes} & \multicolumn{1}{c|}{\textit{d-clear}} & \multicolumn{1}{c|}{\textit{night}} & \multicolumn{1}{c}{\textit{d-rain}} \\
Method & absRel & sqRel & RMSE & $\delta_1$   & absRel & sqRel & RMSE & $\delta_1$  & sqRel & sqRel & sqRel \\
\midrule
md2~\cite{godard2019monodepth2}, \textit{d} &  0.1576 & 2.002 & 7.164 & 80.49 & 0.3148 & 3.001 & 9.523 & 46.72 & 1.820 & 2.879 & 2.296  \\
R4Dyn~\cite{gasperini2021r4dyn}, \textit{d} (radar) & \textbf{0.1365} & 1.830 & 6.957 & \textbf{84.01} & 0.2431 & 2.945 & 10.055 & 56.95 & \textbf{1.661} & 2.889 & 1.938  \\
RNW~\cite{wang2021rnw}, \textit{dn} & 0.2931 & 3.557 & 9.304 & 55.13 & 0.3400 & 4.783 & 10.189 & 44.68 & 3.433 & 4.066 & 3.796 \\
baseline, \textit{d} & 0.1480 & 2.032 & 7.065 & 82.08 & 0.2684 & 3.368 & 10.664 & 53.54 & 1.738 & 2.776 & 2.273 \\
\pname-AD, \textit{dT(nr)} &  0.1602 & 2.245 & 7.226 & 81.02 & 0.2470 & 3.442 & 9.153 & 65.17 & 2.141 & 2.991 & 2.259 \\
\textbf{\pname-DD}, \textit{dT(nr)} & 0.1429 & \textbf{1.828} & \textbf{6.782} & 82.67 & \textbf{0.2143} & \textbf{2.628} & \textbf{8.376} & \textbf{68.03} & 1.752 & \textbf{2.386} & \textbf{1.829} \\

\midrule
AdaBins~\cite{bhat2021adabins}, \textit{a} & 0.1604 & 1.103 & 5.868 & 78.72 & 0.2343 & 1.704 & 7.088 & 61.62 & 0.980 & 1.773 & 1.249 \\
\textbf{\pname-AD}, \textit{dnT(r)} & \textbf{0.1328} & \textbf{0.952} & \textbf{5.139} & \textbf{85.92} & \textbf{0.1967} & \textbf{1.632} & \textbf{6.423} & \textbf{71.67} & \textbf{0.821} & \textbf{1.525} & \textbf{1.199} \\

\bottomrule
\end{tabular}
\end{center}
\vspace{-0.2cm}
\caption{Evaluation of fully-supervised (based on AdaBins~\cite{bhat2021adabins}) and self-supervised methods (based on md2: Monodepth2~\cite{godard2019monodepth2}) on the nuScenes~\cite{caesar2020nuscenes} validation set. The models are the same as in Table~\reff{1}. This table complements Table~\reff{1} with an evaluation on \textit{all} conditions combined, as well as the most challenging \textit{night-rain}.}
\label{table:ext_main_nuscenes}
\vspace{-0.2cm}
\end{table*}

%% file: tables/40m_nuscenes.tex
\begin{table*}
\begin{center}
\begin{tabular}{l|cccc|cccc|cccc}
\toprule
& \multicolumn{4}{c|}{40m -- \textit{day-clear} -- nuScenes} & \multicolumn{4}{c|}{40m -- \textit{night} -- nuScenes} & \multicolumn{4}{c}{40m -- \textit{day-rain} -- nuScenes} \\
Method & absRel & sqRel & RMSE & $\delta_1$   & absRel & sqRel & RMSE & $\delta_1$   & absRel & sqRel &  RMSE & $\delta_1$  \\
\midrule

md2 & \textbf{0.1095} & \textbf{0.796} & \textbf{3.535} & 88.89 & 0.2401 & 1.640 & 5.842 & 60.40 & 0.1405 & 1.083 & 4.259 & 82.64 \\
b.line & 0.1131 & 0.932 & 3.624 & \textbf{89.46} & 0.2118 & 1.816 & 6.476 & 63.47 & 0.1333 & 1.200 & 4.397 & 83.66 \\
AD & 0.1306 & 1.074 & 3.866 & 86.95 & 0.1907 & 1.670 & 5.414 & 73.24 & 0.1329 & 1.083 & 4.332 & 83.54 \\
\textbf{DD} & 0.1173 & 0.877 & 3.592 & 88.22 & \textbf{0.1672} & \textbf{1.322} & \textbf{5.025} & \textbf{75.50} & \textbf{0.1190} & \textbf{0.927} & \textbf{4.036} & \textbf{85.37} \\

\bottomrule
\end{tabular}
\end{center}
\vspace{-0.2cm}
\caption{Evaluation up to 40 meters of self-supervised approaches on the validation set of nuScenes~\cite{caesar2020nuscenes}. md2: Monodepth2~\cite{godard2019monodepth2} trained on \textit{d}. b.line: baseline trained on \textit{d}. AD: \pname-AD trained on \textit{dT(nr)}. DD: \pname-DD trained on \textit{dT(nr)}. The models are the same as in Table~\reff{1}.}
\label{table:40m_nuscenes}
%\vspace{-0.2cm}
\end{table*}

%% file: tables/ext_40m_nuscenes.tex
\begin{table*}
\begin{center}
\begin{tabular}{l|cccc|cccc}
\toprule
& \multicolumn{4}{c|}{40m -- \textit{avg/all} -- nuScenes} & \multicolumn{4}{c}{40m -- \textit{night-rain} -- nuScenes}\\
Method & absRel & sqRel & RMSE & $\delta_1$   & absRel & sqRel & RMSE & $\delta_1$  \\
\midrule

md2 & 0.1276 & \textbf{0.926} & 3.882 & 85.04 & 0.2768 & 2.020 & 6.604 & 53.26 \\
b.line & 0.1262 & 1.063 & 4.034 & 85.93 & 0.2466 & 2.301 & 7.440 & 57.13 \\
AD & 0.1369 & 1.135 & 4.096 & 85.03 & 0.2184 & 2.064 & 6.066 & 68.96 \\
\textbf{DD} & \textbf{0.1225} & 0.930 & \textbf{3.807} & \textbf{86.49} & \textbf{0.1925} & \textbf{1.653} & \textbf{5.661} & \textbf{71.38} \\

\bottomrule
\end{tabular}
\end{center}
\vspace{-0.2cm}
\caption{Evaluation up to 40 meters of self-supervised approaches on the validation set of nuScenes~\cite{caesar2020nuscenes}. md2: Monodepth2~\cite{godard2019monodepth2} trained on \textit{d}. b.line: baseline trained on \textit{d}. AD: \pname-AD trained on \textit{dT(nr)}. DD: \pname-DD trained on \textit{dT(nr)}. The models are the same as in Table~\reff{1}. This table complements Table~\ref{table:40m_nuscenes}.}
\label{table:ext_40m_nuscenes}
\vspace{-0.2cm}
\end{table*}

%% file: tables/60m_nuscenes.tex
\begin{table*}
\begin{center}
\begin{tabular}{l|cccc|cccc|cccc}
\toprule
& \multicolumn{4}{c|}{60m -- \textit{day-clear} -- nuScenes} & \multicolumn{4}{c|}{60m -- \textit{night} -- nuScenes} & \multicolumn{4}{c}{60m -- \textit{day-rain} -- nuScenes} \\
Method & absRel & sqRel & RMSE & $\delta_1$   & absRel & sqRel & RMSE & $\delta_1$   & absRel & sqRel &  RMSE & $\delta_1$  \\
\midrule

md2 & 0.1283 & \textbf{1.387} & 5.447 & 86.15 & 0.2739 & 2.469 & 8.444 & 53.40 & 0.1623 & 1.779 & 6.312 & 79.23 \\
b.line & \textbf{0.1279} & 1.522 & 5.422 & \textbf{86.91} & 0.2348 & 2.779 & 9.502 & 59.28 & 0.1506 & 1.833 & 6.267 & 80.73 \\
AD & 0.1461 & 1.745 & 5.744 & 84.23 & 0.2113 & 2.526 & 7.789 & 69.92 & 0.1519 & 1.774 & 6.421 & 80.28 \\
\textbf{DD} & 0.1310 & 1.419 & \textbf{5.364} & 85.65 & \textbf{0.1859} & \textbf{2.029} & \textbf{7.377} & \textbf{72.10} & \textbf{0.1347} & \textbf{1.463} & \textbf{5.938} & \textbf{82.24} \\

\bottomrule
\end{tabular}
\end{center}
\vspace{-0.2cm}
\caption{Evaluation up to 60 meters of self-supervised approaches on the validation set of nuScenes~\cite{caesar2020nuscenes}. md2: Monodepth2~\cite{godard2019monodepth2} trained on \textit{d}. b.line: baseline trained on \textit{d}. AD: \pname-AD trained on \textit{dT(nr)}. DD: \pname-DD trained on \textit{dT(nr)}. The models are the same as in Table~\reff{1}.}
\label{table:60m_nuscenes}
%\vspace{-0.2cm}
\end{table*}

%% file: tables/ext_60m_nuscenes.tex
\begin{table*}
\begin{center}
\begin{tabular}{l|cccc|cccc}
\toprule
& \multicolumn{4}{c|}{60m -- \textit{avg/all} -- nuScenes} & \multicolumn{4}{c}{60m -- \textit{night-rain} -- nuScenes}\\
Method & absRel & sqRel & RMSE & $\delta_1$   & absRel & sqRel & RMSE & $\delta_1$  \\
\midrule

md2 & 0.1483 & 1.558 & 5.886 & 81.76 & 0.3086 & 2.782 & 8.790 & 47.76 \\
b.line & 0.1423 & 1.698 & 5.966 & 83.15 & 0.2648 & 3.130 & 9.858 & 54.09 \\
AD & 0.1536 & 1.827 & 6.057 & 82.16 & 0.2402 & 2.983 & 8.233 & 65.79 \\
\textbf{DD} & \textbf{0.1371} & \textbf{1.487} & \textbf{5.658} & \textbf{83.74} & \textbf{0.2101} & \textbf{2.368} & \textbf{7.661} & \textbf{68.58} \\

\bottomrule
\end{tabular}
\end{center}
\vspace{-0.2cm}
\caption{Evaluation up to 60 meters of self-supervised approaches on the validation set of nuScenes~\cite{caesar2020nuscenes}. md2: Monodepth2~\cite{godard2019monodepth2} trained on \textit{d}. b.line: baseline trained on \textit{d}. AD: \pname-AD trained on \textit{dT(nr)}. DD: \pname-DD trained on \textit{dT(nr)}. The models are the same as in Table~\reff{1}. This table complements Table~\ref{table:60m_nuscenes}.}
\label{table:ext_60m_nuscenes}
\vspace{-0.2cm}
\end{table*}

%% file: tables/test_nuscenes.tex
\begin{table*}
\begin{center}
\begin{tabular}{l|cccc|cccc|cccc}
\toprule
& \multicolumn{4}{c|}{test -- 40m -- nuScenes} & \multicolumn{4}{c|}{test -- 60m -- nuScenes} & \multicolumn{4}{c}{test -- 80m -- nuScenes} \\
Method & absRel & sqRel & RMSE & $\delta_1$   & absRel & sqRel & RMSE & $\delta_1$   & absRel & sqRel &  RMSE & $\delta_1$  \\
\midrule

md2 & 0.1162 & 0.811 & 3.701 & 87.59 & 0.1376 & 1.364 & 5.650 & 84.08 & 0.1465 & 1.755 & 6.941 & 82.69 \\
RNW & 0.2500 & 2.237 & 6.114 & 62.63 & 0.2781 & 3.420 & 9.222 & 57.50 & 0.2900 & 4.169 & 11.289 & 55.56 \\
b.line & 0.1126 & 0.840 & 3.827 & 87.17 & 0.1275 & 1.364 & 5.747 & 84.33 & 0.1332 & 1.679 & 6.938 & 83.22 \\
AD & 0.1214 & 0.881 & 3.851 & 86.99 & 0.1353 & 1.387 & 5.731 & 84.14 & 0.1409 & 1.691 & 6.915 & 83.00 \\
\textbf{DD} & \textbf{0.1090} & \textbf{0.757} & \textbf{3.606} & \textbf{88.29} & \textbf{0.1221} & \textbf{1.204} & \textbf{5.418} & \textbf{85.57} & \textbf{0.1277} & \textbf{1.503} & \textbf{6.607} & \textbf{84.46} \\

\midrule
AdaBins & 0.1434 & \textbf{0.617} & \textbf{3.233} & 83.09 & 0.1494 & 0.852 & 4.689 & 80.67 & 0.1532 & 1.055 & 5.849 & 79.62\\
\textbf{AD} sup. & \textbf{0.1182} & 0.641 & 3.279 & \textbf{89.40} & \textbf{0.1221} & \textbf{0.785} & \textbf{4.293} & \textbf{87.98} & \textbf{0.1240} & \textbf{0.887} & \textbf{5.021} & \textbf{87.33} \\
\bottomrule
\end{tabular}
\end{center}
\vspace{-0.2cm}
\caption{Evaluation up to 40, up to 60, and up to 80 meters of self-supervised approaches on the test set of nuScenes~\cite{caesar2020nuscenes}. All conditions are evaluated here (\textit{avg/all}). AD sup.: our \pname-AD trained on \textit{dnT(r)} applied on AdaBins~\cite{bhat2021adabins}. md2: Monodepth2~\cite{godard2019monodepth2} trained on \textit{d}. b.line: baseline trained on \textit{d}. AD: \pname-AD trained on \textit{dT(nr)}. DD: \pname-DD trained on \textit{dT(nr)}. Time-dependent normalization was not applied to obtain these results as nuScenes provides no condition annotations for the test set. The models are the same as in Table~\reff{1}.}
\label{table:test_nuscenes}
\vspace{-0.2cm}
\end{table*}

%% file: tables/40m_robotcar.tex
\begin{table*}
\begin{center}
\begin{tabular}{ll|cccc|cccc}
\toprule
&& \multicolumn{4}{c|}{40m -- \textit{day} -- RobotCar} & \multicolumn{4}{c}{40m -- \textit{night} -- RobotCar} \\
Method & tr.data & absRel & sqRel & RMSE & $\delta_1$   & absRel & sqRel & RMSE & $\delta_1$   \\
\midrule

Monodepth2~\cite{godard2019monodepth2}& \textit{d} & 0.1181 & \textbf{0.614} & \textbf{3.034} & 86.51 & 0.3022 & 1.702 & 4.984 & 45.97 \\

[ours] baseline& \textit{d} & 0.1198 & 0.678 & 3.229 & 86.69 & 0.3908 & 3.541 & 8.206 & 22.52 \\

[ours] \pname-AD& \textit{dT(n)} & \textbf{0.1099} & 0.650 & 3.130 & \textbf{88.10} & \textbf{0.1203} & 0.762 & 3.531 & \textbf{85.87} \\

[ours] \textbf{\pname-DD}& \textit{dT(n)} & 0.1120 & 0.618 & 3.125 & 87.18 & 0.1206 & \textbf{0.723} & \textbf{3.479} & 84.92 \\

\bottomrule
\end{tabular}
\end{center}
\vspace{-0.2cm}
\caption{Evaluation of self-supervised works on the RobotCar~\cite{maddern2017oxford} test set up to 40 meters. The models are the same as in Table~\reff{2}.}
\label{table:40m_robotcar}
%\vspace{-0.2cm}
\end{table*}

%% file: tables/60m_robotcar.tex
\begin{table*}
\begin{center}
\begin{tabular}{ll|cccc|cccc}
\toprule
&& \multicolumn{4}{c|}{60m -- \textit{day} -- RobotCar} & \multicolumn{4}{c}{60m -- \textit{night} -- RobotCar} \\
Method & tr.data & absRel & sqRel & RMSE & $\delta_1$   & absRel & sqRel & RMSE & $\delta_1$   \\
\midrule

Monodepth2~\cite{godard2019monodepth2}& \textit{d} & 0.1201 & 0.698 & \textbf{3.215} & 86.38 & 0.3029 & 1.728 & 5.045 & 45.88 \\

[ours] baseline& \textit{d} & 0.1213 & 0.746 & 3.382 & 86.61 & 0.3909 & 3.548 & 8.228 & 22.51 \\

[ours] \pname-AD & \textit{dT(n)} & \textbf{0.1116} & 0.731 & 3.291 & \textbf{88.02} & 0.1231 & 0.903 & 3.812 & \textbf{85.76} \\

[ours] \textbf{\pname-DD}& \textit{dT(n)} & 0.1130 & \textbf{0.661} & 3.234 & 87.13 & \textbf{0.1225} & \textbf{0.824} & \textbf{3.664} & 84.86 \\

\bottomrule
\end{tabular}
\end{center}
\vspace{-0.2cm}
\caption{Evaluation of self-supervised works on the RobotCar~\cite{maddern2017oxford} test set up to 60 meters. The models are the same as in Table~\reff{2}.}
\label{table:60m_robotcar}
%\vspace{-0.2cm}
\end{table*}

%% file: tables/80m_robotcar.tex
\begin{table*}[t!]
\begin{center}
\begin{tabular}{ll|cccc|cccc}
\toprule
&& \multicolumn{4}{c|}{80m -- \textit{day} -- RobotCar} & \multicolumn{4}{c}{80m -- \textit{night} -- RobotCar} \\
Method & tr.data & absRel & sqRel & RMSE & $\delta_1$   & absRel & sqRel & RMSE & $\delta_1$   \\
\midrule

Monodepth2~\cite{godard2019monodepth2}& \textit{d} & 0.1203 & 0.718 & 3.245 & 86.38 & 0.3030 & 1.729 & 5.046 & 45.88 \\

[ours] baseline& \textit{d} & 0.1214 & 0.759 & 3.404 & 86.61 & 0.3909 & 3.548 & 8.228 & 22.51 \\

[ours] \pname-AD& \textit{dT(n)} & \textbf{0.1118} & 0.742 & 3.308 & \textbf{88.02} & 0.1236 & 0.952 & 3.880 & \textbf{85.76} \\

[ours] \textbf{\pname-DD}& \textit{dT(n)} & 0.1131 & \textbf{0.666} & \textbf{3.243} & 87.13 & \textbf{0.1229} & \textbf{0.865} & \textbf{3.713} & 84.86 \\

\bottomrule
\end{tabular}
\end{center}
\vspace{-0.2cm}
\caption{Evaluation of self-supervised works on the RobotCar~\cite{maddern2017oxford} test set up to 80 meters. The models are the same as in Table~\reff{2}.}
\label{table:80m_robotcar}
\vspace{-0.2cm}
\end{table*}